\documentclass[11pt]{article}

\usepackage[preprint]{acl}
\usepackage{enumitem}
\usepackage{times}
\usepackage{latexsym}
\usepackage{makecell}
\usepackage[T1]{fontenc}
\usepackage{booktabs}
\usepackage{amsmath}
\usepackage{tabularx}
\usepackage{booktabs}
\usepackage{multirow}
\usepackage{makecell}
\usepackage{colortbl}
\usepackage{adjustbox}
\usepackage{dblfloatfix}
\usepackage{subcaption}  
\usepackage{adjustbox}
\usepackage{booktabs}
\usepackage{colortbl}
\usepackage{makecell}
\usepackage{multirow}
\usepackage{xurl}  
\usepackage{hyperref}

\usepackage[utf8]{inputenc}
\usepackage{tcolorbox}
\tcbuselibrary{skins, breakable}
\usepackage{multicol}
\usepackage{subcaption}
\usepackage{microtype}
\usepackage{url}
\usepackage{inconsolata}

\usepackage{graphicx}
\usepackage{xcolor}
\usepackage{colortbl}
\usepackage[table]{xcolor} 
\usepackage{multirow}
\usepackage{amssymb} 
\usepackage{soul}
\usepackage{xcolor}
\sethlcolor{yellow!35}
\usepackage{array}
\usepackage{booktabs}
\usepackage{hyperref}
\hypersetup{
  colorlinks=true,
  linkcolor=black,
  urlcolor=blue,
}

\usepackage{tcolorbox}
\tcbuselibrary{skins, breakable}

\usepackage{soul}
\usepackage{tcolorbox}
\sethlcolor{yellow!35}
\sethlcolor{yellow!35}

\definecolor{mygray}{gray}{0.9}

\usepackage[table]{xcolor}

\definecolor{LightGreen}{RGB}{230,245,233}
\definecolor{LightRed}{RGB}{253,232,232}
\definecolor{LightYellow}{RGB}{255,246,204}
\definecolor{LightGray}{RGB}{245,245,245}

\newcommand{\algok}{\cellcolor{LightGreen}\textsc{Correct}}
\newcommand{\algwrong}{\cellcolor{LightRed}\textsc{Incorrect}}

\newcommand{\puNo}{\cellcolor{LightGreen}No}
\newcommand{\puYes}{\cellcolor{LightRed}Yes}

\newcommand{\sevNone}{\cellcolor{LightGreen}None}
\newcommand{\sevMinor}{\cellcolor{LightYellow}Minor}
\newcommand{\sevMajor}{\cellcolor{LightRed}Major}

\title{Idea First, Code Later: Disentangling Problem Solving from Code Generation in Evaluating LLMs for Competitive Programming}

\author{
\textbf{
Sama Hadhoud$^{1}$ \quad
Alaa Elsetohy$^{1}$ \quad
Frederikus Hudi$^{2}$}
\\[0.2em]
\textbf{
Jan Christian Blaise Cruz$^{1}$ \quad
Steven Halim$^{3}$ \quad
Alham Fikri Aji$^{1}$
}\\[0.6em]
$^{1}$MBZUAI \quad
$^{2}$NAIST \quad
$^{3}$National University of Singapore
\\[0.5em]
\texttt{\{sama.hadhoud, alaa.elsetohy, jan.cruz, alham.fikri\}@mbzuai.ac.ae}\\
\texttt{frederikus.hudi.fe7@naist.ac.jp \quad dcssh@nus.edu.sg}
}

\usepackage{tcolorbox}
\tcbuselibrary{skins, breakable}
\newtcolorbox{mainexcerpt}[1][]{
  colback=white,          
  colframe=black!25,      
  boxrule=0.5pt,
  arc=2pt,
  boxsep=2pt,
  left=4pt,
  right=4pt,
  top=4pt,
  bottom=4pt,
  enhanced,
  fontupper=\small,      
  sharp corners,         
}

\newtcolorbox{appexcerpt}[1][]{
  breakable,
  colback=white,
  colframe=black!20,
  boxrule=0.6pt,
  arc=3pt,
  boxsep=4pt,
  left=6pt,
  right=6pt,
  top=6pt,
  bottom=6pt,
  enhanced,
  fontupper=\small,
  title=#1,
  attach boxed title to top left={xshift=4mm,yshift=-3mm},
  boxed title style={colback=black!5, colframe=black!20, boxrule=0.4pt, sharp corners, font=\footnotesize}
}

\begin{document}
\maketitle

\begin{abstract}

Large Language Models (LLMs) increasingly succeed on competitive programming problems, yet existing evaluations conflate algorithmic reasoning with code-level implementation. We argue that competitive programming is fundamentally a problem-solving task and propose centering natural-language editorials in both solution generation and evaluation. Generating an editorial prior to code improves solve rates for some LLMs, with substantially larger gains when using expertly written gold editorials. However, even with gold editorials, models continue to struggle with implementation, while the gap between generated and gold editorials reveals a persistent problem-solving bottleneck in specifying correct and complete algorithms. Beyond pass/fail metrics, we diagnose reasoning errors by comparing model-generated editorials to gold standards using expert annotations and validate an LLM-as-a-judge protocol for scalable evaluation. We introduce a dataset of 83 ICPC-style problems with gold editorials and full test suites, and evaluate 19 LLMs, arguing that future benchmarks should explicitly separate problem solving from implementation.
\end{abstract}

\begin{figure*}[t]
\centering
\includegraphics[width=2\columnwidth]{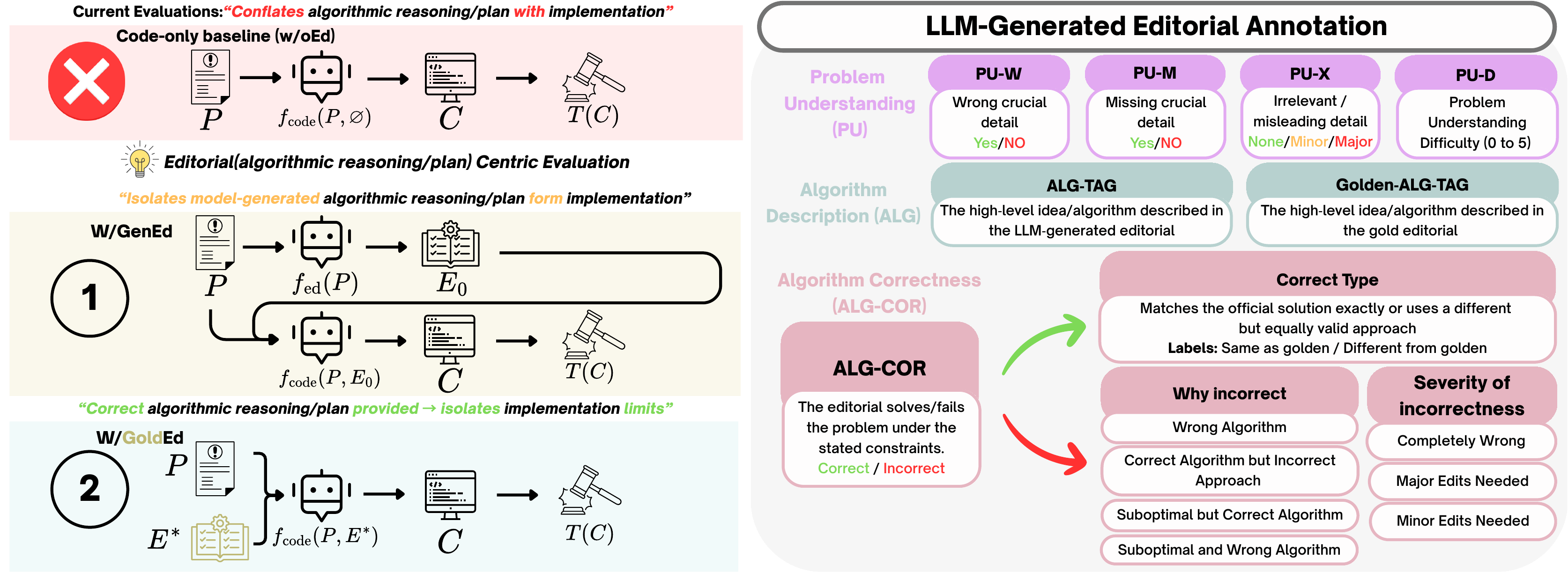}
\caption{
Overview of our evaluation pipeline and editorial annotation scheme.
Left: three settings, \textbf{w/oEd} (problem $\rightarrow$ code, baseline),
\textbf{w/GenEd} (problem $\rightarrow$ generated editorial $\rightarrow$ code),
and \textbf{w/GoldEd} (problem plus gold editorial $\rightarrow$ code).
Right: the LLM-generated editorial annotation rubric used to diagnose reasoning quality, covering \emph{Problem Understanding} (PU-W, PU-M, PU-X, PU-D), \emph{Algorithm Description} (ALG-TAG vs.\ Golden-ALG-TAG), and \emph{Algorithm Correctness} (ALG-COR, correctness type, error type, and severity).
}

    \label{fig:method_overview}
\end{figure*}

\section{Introduction}


Competitive Programming (CP) is framed as a coding contest, but it primarily evaluates \emph{algorithmic problem solving} under strict time and memory limits.
For human contestants, the core deliverable is an \emph{algorithmic plan}: key observations, data structures, invariants, and complexity arguments, while code is a downstream translation into an executable program \citep{halim2020competitive}.
This workflow is reflected by contest \emph{editorials}: natural-language explanations that describe the algorithm and justify correctness, and which serve as the solution reference after a contest.\footnote{See Appendix~\ref{appendix:example-problem} for an example gold editorial.}

LLMs have rapidly improved on CP-style tasks.
Early systems such as AlphaCode demonstrated strong contest performance via large-scale
sampling \cite{Li_2022}, and more recent reasoning-oriented models achieve substantially
higher accuracy on difficult competitive benchmarks
\cite{openai2025competitiveprogramminglargereasoning,deepseekai2025deepseekr1incentivizingreasoningcapability},
with frontier results approaching elite human levels
\cite{openai2025icpc,deepmind2025gemini}.
Despite this progress, most evaluations still treat CP as a single \emph{problem-to-code}
mapping \cite{chen2021codex, hendrycksapps2021, jain2024livecodebench,shi2024languagemodelssolveolympiad, hossain2025llmprosanalyzinglargelanguage, quan2025codeelo}, scoring only the final program.
This end-to-end protocol conflates two distinct capabilities:
\textbf{problem solving}—deriving a correct and efficient algorithm—and
\textbf{implementation}—translating that plan into correct and efficient code.
When a submission fails, standard metrics cannot distinguish between these failure modes.

We revisit CP evaluation by making \textit{editorials an explicit intermediate artifact}.
Figure~\ref{fig:method_overview} introduces an \textbf{editorial-centric evaluation pipeline} with three
conditions:
\textbf{w/oEd} (problem $\rightarrow$ code),
\textbf{w/GenEd} (problem $\rightarrow$ model editorial $\rightarrow$ code), and
\textbf{w/GoldEd} (problem + gold editorial $\rightarrow$ code).
Crucially, \textbf{w/GoldEd approximates performance under perfect problem solving},
isolating implementation limitations, while the gap between w/GenEd and
w/GoldEd directly reflects the reliability of model-generated reasoning.

To support this evaluation, we curate a dataset of 83 ICPC-style problems from seven contests (2017--2025),
each packaged with its original statement, an expert-written gold editorial, and the full official test suite.
We evaluate 19 contemporary LLMs under all three conditions using standard ICPC-style judging.
Beyond pass/fail, we analyze generated editorials directly: we annotate a qualitative subset with an expert
competitive programmer and validate a scalable \emph{LLM-as-a-judge} protocol that labels editorial quality
against gold references (e.g., problem understanding, algorithm description, and algorithmic correctness).

Overall, gold editorials yield large and consistent gains, but performance remains far from
saturated even under gold guidance, highlighting a substantial implementation bottleneck.
Self-generated editorials provide smaller and less reliable gains and can even degrade
performance when reasoning is misleading.
These results suggest that CP benchmarks should move beyond problem-to-code scoring and explicitly evaluate reasoning and implementation as separate, measurable components, with editorials serving as the bridge.

To summarize, our contributions are as follows:
\begin{itemize}[itemsep=3pt, topsep=2pt, parsep=0pt]
    \item We decompose competitive programming evaluation into a \textbf{Problem $\rightarrow$ Editorial $\rightarrow$ Code} pipeline that isolates problem solving from implementation.
    \item We quantify a \textbf{problem-solving gap} by comparing \textbf{w/GenEd} against \textbf{w/GoldEd}, showing that generated editorials often fail to specify correct and efficient algorithms.
    \item We identify an \textbf{implementation gap} by showing that models fail even under \textbf{w/GoldEd}, where the correct algorithm is given.
    \item We provide expert competitive-programmer annotations on editorial quality and validate an \textbf{LLM-as-a-judge} protocol that aligns with expert judgments.
    \item We show that editorials can transfer across models, enabling ``writer--coder'' compositions where a strong planner improves a different model’s implementations.
\end{itemize}

\section{Editorial-Centric Competitive Programming Evaluation}

\subsection{Problem setup}

Let $P$ denote a CP problem statement. Each contest problem in our dataset follows the usual structure: a natural-language statement, explicit time and memory limits, an \emph{Input} section describing the input format, an \emph{Output} section specifying the required output format, and one or more sample input/output pairs. We preserve this layout in Markdown and include one full example in Appendix~\ref{appendix:example-problem}.

We model solution generation using two objects:
an \emph{editorial} $E$, which represents algorithmic reasoning and planning, and a \emph{program} $C$, which represents executable code. Program correctness is evaluated
using a standard ICPC-style compile-and-run judging pipeline on the official contest
test suites (see Appendix~\ref{app:judging} for full details), yielding an outcome
\(
T(C) \in \{\text{PASS},  \text{Compile Error} \text{(CE)}, \text{Wrong Answer}  \text{(WA)}, \newline \text{Time Limit Exceeded} \text{(TLE)}, \text{MLE Memory Limit} \newline  \text{Exceeded}\text{(MLE)}, \text{Runtime Error}\text{(RTE)}\},
\)

We explicitly separate reasoning and coding through two generation operators:
\[
E = f_{\mathrm{ed}}(P), \qquad
C = f_{\mathrm{code}}(P, E),
\]
where $f_{\mathrm{ed}}$ produces an editorial from the problem statement, and $f_{\mathrm{code}}$ produces code conditioned on both the problem and a given editorial. When no editorial is provided, $f_{\mathrm{code}}(P,\varnothing)$ corresponds to direct problem-to-code generation.

\subsection{Editorial-Centric Generation}
\label{sec:Experimental Conditions}

\paragraph{Code-only baseline (w/oEd).}
In the baseline setting, the model is asked to solve the problem directly: \(C = f_{\mathrm{code}}(P, \varnothing).\)
This mirrors standard CP benchmarks and conflates reasoning and implementation into a single step. Any failure may stem from incorrect planning, incorrect coding, or both.

\paragraph{Generated editorial (w/GenEd).}
To isolate model-generated reasoning, we first ask the model to write an editorial and then generate code conditioned on it: \(E_0 = f_{\mathrm{ed}}(P),\; C = f_{\mathrm{code}}(P, E_0).\)
The editorial $E_0$ is generated once and kept fixed. This allows us to distinguish two failure modes: (i) the editorial itself describes an incorrect or incomplete algorithm, or (ii) the editorial is correct, but the model fails to faithfully implement it.

\paragraph{Gold editorial (w/GoldEd).}
To estimate an upper bound given correct reasoning, we provide the model with an expert-written gold editorial $E^{\ast}$: \(C = f_{\mathrm{code}}(P, E^{\ast})\).
Any remaining errors therefore arise from implementation limitations.

\noindent Throughout, we use a \emph{single-shot} setting with a minimal prompting scheme (see Appendix~\ref{appendix:prompts}).


\begin{table*}[t]
\centering

\small
\renewcommand{\arraystretch}{1.10} 
\setlength{\tabcolsep}{3pt}       
\begin{adjustbox}{max width=\textwidth}
\begin{tabular}{lcccccccccccc}
\toprule
\multirow{2}{*}{\textbf{Model}} &
\multicolumn{3}{c}{\textbf{Overall (83 problems)}} &
\multicolumn{3}{c}{\textbf{T1 (26 problems)}} &
\multicolumn{3}{c}{\textbf{T2 (28 problems)}} &
\multicolumn{3}{c}{\textbf{T3 (29 problems)}} \\
\cmidrule(lr){2-4} \cmidrule(lr){5-7} \cmidrule(lr){8-10} \cmidrule(lr){11-13}
& \textbf{w/oEd} & \textbf{w/GenEd (\%/$\Delta$)} & \textbf{w/GoldEd (\%/$\Delta$)}
& \textbf{w/oEd} & \textbf{w/GenEd (\%/$\Delta$)} & \textbf{w/GoldEd (\%/$\Delta$)}
& \textbf{w/oEd} & \textbf{w/GenEd (\%/$\Delta$)} & \textbf{w/GoldEd (\%/$\Delta$)}
& \textbf{w/oEd} & \textbf{w/GenEd (\%/$\Delta$)} & \textbf{w/GoldEd (\%/$\Delta$)} \\
\midrule

\multicolumn{13}{c}{\cellcolor{blue!15}\textbf{Closed Source Models}} \\
\midrule

GPT-5 & \textbf{67.5\%} & \cellcolor{green!10}\textbf{68.7\%} / +1.2\% & \cellcolor{green!60}\textbf{83.1\%} / +15.7\% &
\textbf{92.3\%} & \cellcolor{red!20}\textbf{88.5\%} / -3.8\% & \cellcolor{green!20}\textbf{96.2\%} / +3.8\% &
\textbf{75.0\%} & \cellcolor{gray!10}\textbf{75.0\%} / +0.0\% & \cellcolor{green!60}\textbf{92.9\%} / +17.9\% &
\textbf{37.9\%} & \cellcolor{green!40}\textbf{44.8\%} / \textbf{+6.9\%} & \cellcolor{green!80}\textbf{62.1\%} / \underline{+24.1\%} \\

O3 & \underline{51.8\%} & \cellcolor{red!30}\underline{45.8\%} / -6.0\% & \cellcolor{green!60}63.9\% / +12.0\% & \underline{69.2\%} & \cellcolor{green!40}\underline{76.9\%} / \underline{+7.7\%} & \cellcolor{green!60}80.8\% / +11.5\% & \underline{60.7\%} & \cellcolor{red!40}46.4\% / -14.3\% & \cellcolor{green!60}75.0\% / +14.3\% & \underline{27.6\%} & \cellcolor{red!40}\underline{17.2\%} / -10.3\% & \cellcolor{green!60}37.9\% / +10.3\% \\
Gemini 2.5 Pro & 43.4\% & \cellcolor{green!20}45.8\% / +2.4\% & \cellcolor{green!80}\underline{72.3\%} / \underline{+28.9\%} & 69.2\% & \cellcolor{green!20}73.1\% / +3.8\% & \cellcolor{green!80}\underline{92.3\%} / +23.1\% & 42.9\% & \cellcolor{green!40}\underline{50.0\%} / +7.1\% & \cellcolor{green!80}\underline{78.6\%} / \underline{+35.7\%} & 20.7\% & \cellcolor{red!20}17.2\% / -3.4\% & \cellcolor{green!80}\underline{48.3\%} / \textbf{+27.6\%} \\
Gemini 2.5 Flash & 38.6\% & \cellcolor{red!10}37.3\% / -1.2\% & \cellcolor{green!60}54.2\% / +15.7\% & 65.4\% & \cellcolor{green!40}73.1\% / +7.7\% & \cellcolor{green!60}80.8\% / +15.4\% & 39.3\% & \cellcolor{red!30}32.1\% / -7.1\% & \cellcolor{green!60}53.6\% / +14.3\% & 13.8\% & \cellcolor{red!20}10.3\% / -3.4\% & \cellcolor{green!60}31.0\% / +17.2\% \\

Claude Opus 4 & 21.7\% & \cellcolor{green!40}30.1\% / \textbf{+8.4\%} & \cellcolor{green!80}47.0\% / +25.3\% & 42.3\% & \cellcolor{green!60}57.7\% / \textbf{+15.4\%} & \cellcolor{green!80}76.9\% / \underline{+34.6\%} & 21.4\% & \cellcolor{green!60}32.1\% / \underline{+10.7\%} & \cellcolor{green!80}57.1\% / +35.7\% & 3.4\% & \cellcolor{gray!10}3.4\% / +0.0\% & \cellcolor{green!40}10.3\% / +6.9\% \\
Claude Sonnet 4 & 16.9\% & \cellcolor{green!20}19.3\% / +2.4\% & \cellcolor{green!80}48.2\% / \textbf{+31.3\%} & 34.6\% & \cellcolor{green!20}38.5\% / +3.8\% & \cellcolor{green!80}73.1\% / \textbf{+38.5\%} & 17.9\% & \cellcolor{gray!10}17.9\% / +0.0\% & \cellcolor{green!80}57.1\% / \textbf{+39.3\%} & 0.0\% & \cellcolor{green!20}3.4\% / \underline{+3.4\%} & \cellcolor{green!60}17.2\% / +17.2\% \\

GPT-4.1 & 13.3\% & \cellcolor{green!20}17.1\% / \underline{+3.8\%} & \cellcolor{green!80}33.7\% / +20.5\% & 26.9\% & \cellcolor{green!40}34.6\% / +7.7\% & \cellcolor{green!80}61.5\% / +34.6\% & 3.6\% & \cellcolor{green!60}14.8\% / \textbf{+11.2\%} & \cellcolor{green!80}32.1\% / +28.6\% & 10.3\% & \cellcolor{red!30}3.4\% / -6.9\% & \cellcolor{gray!10}10.3\% / +0.0\% \\
GPT-4o & 7.2\% & \cellcolor{red!20}3.6\% / -3.6\% & \cellcolor{green!40}13.3\% / +6.0\% & 19.2\% & \cellcolor{red!40}7.7\% / -11.5\% & \cellcolor{green!60}38.5\% / +19.2\% & 0.0\% & \cellcolor{green!20}3.6\% / +3.6\% & \cellcolor{green!20}3.6\% / +3.6\% & 3.4\% & \cellcolor{red!20}0.0\% / -3.4\% & \cellcolor{red!20}0.0\% / -3.4\% \\


\midrule
\textit{Closed Source Avg} & \textit{32.5\%} & \cellcolor{green!10}\textit{33.5\% / +0.9\%} & \cellcolor{green!60}\textit{52.0\% / +19.4\%} &
\textit{52.4\%} & \cellcolor{green!20}\textit{56.2\% / +3.8\%} & \cellcolor{green!80}\textit{75.0\% / +22.6\%} &
\textit{32.6\%} & \cellcolor{green!10}\textit{34.0\% / +1.4\%} & \cellcolor{green!80}\textit{56.2\% / +23.7\%} &
\textit{14.7\%} & \cellcolor{red!20}\textit{12.5\% / -2.2\%} & \cellcolor{green!60}\textit{27.2\% / +12.5\%} \\

\midrule
\multicolumn{13}{c}{\cellcolor{orange!15}\textbf{Open Source Models}} \\
\midrule


GPT-OSS-120B & \textbf{41.0\%} & \cellcolor{red!30}\underline{31.3\%} / -9.6\% & \cellcolor{green!60}\textbf{59.0\%} / \textbf{+18.1\%} & \textbf{61.5\%} & \cellcolor{gray!10}\underline{61.5\%} / +0.0\% & \cellcolor{green!60}\textbf{76.9\%} / +15.4\% & \textbf{50.0\%} & \cellcolor{red!40}\underline{32.1\%} / -17.9\% & \cellcolor{green!80}\textbf{75.0\%} / \underline{+25.0\%} & \textbf{13.8\%} & \cellcolor{red!40}\underline{3.4\%} / -10.3\% & \cellcolor{green!60}\textbf{27.6\%} / \underline{+13.8\%} \\
GPT-OSS-20B & \underline{33.7\%} & \cellcolor{red!30}27.7\% / -6.0\% & \cellcolor{green!60}\underline{47.0\%} / +13.3\% & 57.7\% & \cellcolor{red!20}53.8\% / -3.8\% & \cellcolor{green!40}\underline{65.4\%} / +7.7\% & \underline{39.3\%} & \cellcolor{red!40}28.6\% / -10.7\% & \cellcolor{green!60}\underline{53.6\%} / +14.3\% & \underline{6.9\%} & \cellcolor{red!20}3.4\% / -3.4\% & \cellcolor{green!60}\underline{24.1\%} / \textbf{+17.2\%} \\
DeepSeekR1 & 28.9\% & \cellcolor{green!60}\textbf{45.8\%} / \textbf{+16.9\%} & \cellcolor{green!60}43.4\% / +14.5\% & \underline{61.5\%} & \cellcolor{green!80}\textbf{88.5\%} / \textbf{+26.9\%} & \cellcolor{green!20}65.4\% / +3.8\% & 21.4\% & \cellcolor{green!60}\textbf{39.3\%} / \textbf{+17.9\%} & \cellcolor{green!80}53.6\% / \textbf{+32.1\%} & 6.9\% & \cellcolor{green!40}\textbf{13.8\%} / \textbf{+6.9\%} & \cellcolor{green!40}13.8\% / +6.9\% \\

Qwen3-8B & 15.7\% & \cellcolor{red!20}13.3\% / -2.4\% & \cellcolor{green!40}24.1\% / +8.4\% & 34.6\% & \cellcolor{gray!10}34.6\% / +0.0\% & \cellcolor{green!40}42.3\% / +7.7\% & 14.3\% & \cellcolor{red!30}7.1\% / -7.1\% & \cellcolor{green!60}25.0\% / +10.7\% & 0.0\% & \cellcolor{gray!10}0.0\% / +0.0\% & \cellcolor{green!40}6.9\% / +6.9\% \\

DeepSeekV3 & 14.5\% & \cellcolor{red!20}9.6\% / -4.8\% & \cellcolor{green!60}28.9\% / +14.5\% & 34.6\% & \cellcolor{red!30}26.9\% / -7.7\% & \cellcolor{green!80}61.5\% / \underline{+26.9\%} & 10.7\% & \cellcolor{red!30}3.6\% / -7.1\% & \cellcolor{green!60}25.0\% / +14.3\% & 0.0\% & \cellcolor{gray!10}0.0\% / +0.0\% & \cellcolor{green!20}3.4\% / +3.4\% \\

\makecell[l]{Qwen3-Coder-\\480B-A35B} & 13.3\% & \cellcolor{red!20}10.8\% / -2.4\% & \cellcolor{green!60}28.9\% / \underline{+15.7\%} & 23.1\% & \cellcolor{red!20}19.2\% / -3.8\% & \cellcolor{green!80}61.5\% / \textbf{+38.5\%} & 14.3\% & \cellcolor{red!20}10.7\% / -3.6\% & \cellcolor{green!20}17.9\% / +3.6\% & 3.4\% & \cellcolor{gray!10}3.4\% / +0.0\% & \cellcolor{green!40}10.3\% / +6.9\% \\

Kimi-K2 & 13.3\% & \cellcolor{gray!10}13.3\% / +0.0\% & \cellcolor{green!60}26.5\% / +13.3\% & 34.6\% & \cellcolor{gray!10}34.6\% / +0.0\% & \cellcolor{green!60}50.0\% / +15.4\% & 3.6\% & \cellcolor{green!20}7.1\% / \underline{+3.6\%} & \cellcolor{green!80}28.6\% / +25.0\% & 3.4\% & \cellcolor{red!20}0.0\% / -3.4\% & \cellcolor{gray!10}3.4\% / +0.0\% \\

OlympicCoder-7B & 6.0\% & \cellcolor{green!20}8.4\% / \underline{+2.4\%} & \cellcolor{green!20}10.8\% / +4.8\% & 19.2\% & \cellcolor{green!40}26.9\% / \underline{+7.7\%} & \cellcolor{green!40}26.9\% / +7.7\% & 0.0\% & \cellcolor{gray!10}0.0\% / +0.0\% & \cellcolor{green!40}7.1\% / +7.1\% & 0.0\% & \cellcolor{gray!10}0.0\% / +0.0\% & \cellcolor{gray!10}0.0\% / +0.0\% \\

Llama-3.1-405B & 6.0\% & \cellcolor{red!20}2.4\% / -3.6\% & \cellcolor{green!40}15.7\% / +9.6\% & 19.2\% & \cellcolor{red!40}7.7\% / -11.5\% & \cellcolor{green!80}42.3\% / +23.1\% & 0.0\% & \cellcolor{gray!10}0.0\% / +0.0\% & \cellcolor{green!20}3.6\% / +3.6\% & 0.0\% & \cellcolor{gray!10}0.0\% / +0.0\% & \cellcolor{green!20}3.4\% / +3.4\% \\
Llama-3.3-70B & 4.8\% & \cellcolor{green!10}6.0\% / +1.2\% & \cellcolor{green!20}8.4\% / +3.6\% & 11.5\% & \cellcolor{gray!10}11.5\% / +0.0\% & \cellcolor{green!60}23.1\% / +11.5\% & 3.6\% & \cellcolor{green!20}7.1\% / +3.6\% & \cellcolor{red!20}0.0\% / -3.6\% & 0.0\% & \cellcolor{gray!10}0.0\% / +0.0\% & \cellcolor{green!20}3.4\% / +3.4\% \\

Gemma-3-27B & 3.6\% & \cellcolor{red!10}2.4\% / -1.2\% & \cellcolor{green!20}8.4\% / +4.8\% & 7.7\% & \cellcolor{red!20}3.8\% / -3.8\% & \cellcolor{green!60}23.1\% / +15.4\% & 3.6\% & \cellcolor{gray!10}3.6\% / +0.0\% & \cellcolor{gray!10}3.6\% / +0.0\% & 0.0\% & \cellcolor{gray!10}0.0\% / +0.0\% & \cellcolor{gray!10}0.0\% / +0.0\% \\

\midrule
\textit{Open Source Avg} & \textit{16.4\%} & \cellcolor{red!10}\textit{15.6\% / -0.9\%} & \cellcolor{green!60}\textit{27.4\% / +11.0\%} &
\textit{33.2\%} & \cellcolor{green!10}\textit{33.6\% / +0.3\%} & \cellcolor{green!60}\textit{49.0\% / +15.7\%} &
\textit{14.6\%} & \cellcolor{red!10}\textit{12.7\% / -1.9\%} & \cellcolor{green!60}\textit{26.6\% / +12.0\%} &
\textit{3.1\%} & \cellcolor{red!10}\textit{2.2\% / -0.9\%} & \cellcolor{green!40}\textit{8.8\% / +5.6\%} \\

\midrule
\textit{Overall Avg} & \textit{23.2\%} & \cellcolor{red!10}\textit{23.1\% / -0.1\%} & \cellcolor{green!60}\textit{37.7\% / +14.5\%} &
\textit{41.3\%} & \cellcolor{green!10}\textit{43.1\% / +1.8\%} & \cellcolor{green!60}\textit{59.9\% / +18.6\%} &
\textit{22.2\%} & \cellcolor{red!10}\textit{21.6\% / -0.5\%} & \cellcolor{green!60}\textit{39.1\% / +16.9\%} &
\textit{8.0\%} & \cellcolor{red!10}\textit{6.5\% / -1.5\%} & \cellcolor{green!40}\textit{16.5\% / +8.5\%} \\

\bottomrule
\end{tabular}
\end{adjustbox}
\caption{Pass@1 by difficulty tertile (T1 easiest–T3 hardest).
Gold editorials substantially improve performance across all difficulties (up to $\sim$30\%), but hard problems remain challenging, indicating a residual \textbf{implementation bottleneck}. Self-generated editorials yield smaller, model-dependent gains (up to $\sim$15\%) and sometimes hurt performance, highlighting a persistent \textbf{problem-solving gap} in model-generated reasoning.}

\label{tab:model-performance}

\end{table*}
\vspace{-10pt}
\begin{figure*}[t]
\centering
\vspace{-6pt}
\begin{minipage}[t]{0.49\textwidth}
  \centering
  \includegraphics[width=\linewidth,trim=0 15 0 6,clip]{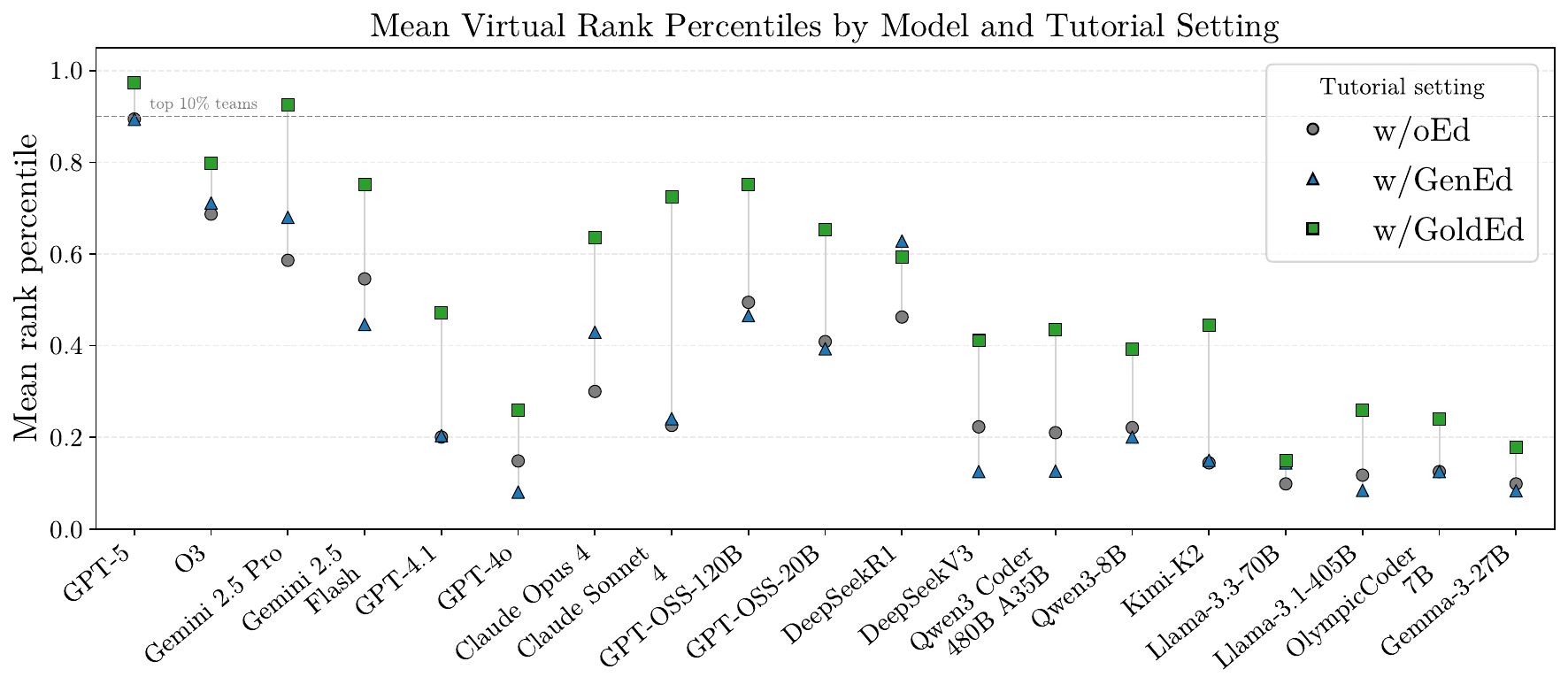}
  \captionof{figure}{Mean virtual rank percentile under w/oEd, w/GenEd, and w/GoldEd (higher is better). Gold editorials yield large and consistent improvements (up to $\sim$0.4), yet even under gold guidance only a small number of models attain high rank percentiles (above $\sim$0.8), with only a handful exceeding $\sim$0.7.}
  \label{fig:ladder-rank}
\end{minipage}
\hfill
\begin{minipage}[t]{0.49\textwidth}
  \centering
  \includegraphics[width=\linewidth,trim=0 15 0 6,clip]{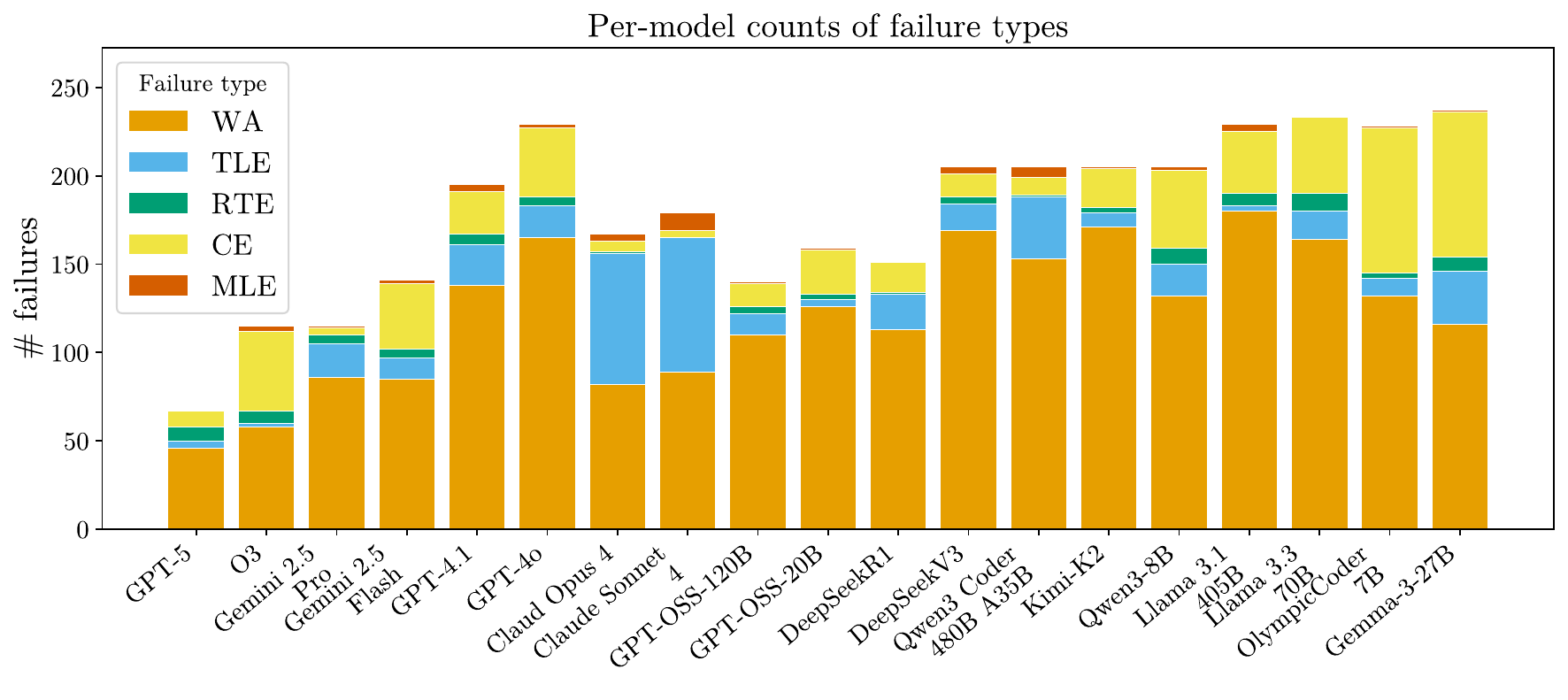}
  \captionof{figure}{Aggregate failure verdict distribution across \emph{all} editorial settings: Wrong Answer (WA), Time Limit Exceeded (TLE), Runtime Error (RTE), Compile Error (CE), and Memory Limit Exceeded (MLE). Remaining failures are dominated by WA, while TLE becomes more salient for some stronger models (notably Claude)}
  \label{fig:failure-types}
\end{minipage}

\vspace{-10pt}
\end{figure*}

\subsection{Contest Dataset and Evaluation Metrics}
\label{sec:dataset-metrics}


We curate 83 problems from seven contests that are not hosted on major public CP platforms (e.g., Codeforces, AtCoder). Some problem statements may be publicly accessible (e.g., contest PDFs or course materials), but we expect contamination risk to be lower than for widely scraped judge platforms.
The problems come from regional ICPC contests and CS3233 (Competitive Programming course) examinations at the National University of Singapore, spanning 2017–2025.
\footnote{See Appendix~\ref{app:data-details} for dataset and release details.}

Each problem package includes the original problem statement, a gold editorial written by the problem setter or tester, and the full official test suite. We rely on complete judge test suites to avoid false positives.


To account for variation in problem difficulty, we group problems using solve rates\footnote{The proportion of teams that solved the problem.} from official scoreboards. Within each contest, problems are ranked by solve rate and partitioned into three contest-relative tertiles: T1 (easiest), T2 (middle), and T3 (hardest). Pooling across contests yields approximately balanced difficulty groups while preserving each contest’s intrinsic difficulty profile.

We report two complementary metrics. \emph{pass@1} measures the fraction of problems for which a model’s first submission passes the full test suite. We also report a \emph{virtual rank percentile} to contextualize performance relative to human teams: for each contest and setting, ignoring time-based penalties, the model is treated as an additional team whose score equals the number of problems solved and is inserted into the official scoreboard (1.0 = top team, 0.5 = median, 0.0 = last place). Averaging these percentiles across contests yields a single measure of human competitiveness.

\subsection{Generated Editorial Evaluation by Experts}
\label{subsec:annotation}

We perform an expert evaluation to understand the reasoning behaviors exhibited in model-generated editorials and how these behaviors relate to the results of the downstream execution. The study is performed on one representative contest (CS3233 2025 Midterm). For each problem, annotators review the gold editorial and the model-generated editorial from the \textbf{w/GenEd} condition.

\paragraph{Annotators}
Annotations were performed by an experienced IOI medalist competitive programmer with ICPC World Final participation; the annotator was blind to the identity of the model.

The pool of qualified annotators is very small; typically restricted to elite competitive programmers such as ICPC World Finalists and producing editorial level annotation under our rubric can take several hours.

\paragraph{Annotation Procedure}
Editorials are evaluated along three dimensions:
\begin{itemize}[nosep, leftmargin=*, labelsep=0.5em]
    \item \textbf{Problem Understanding (PU)}: whether the editorial correctly captures the task, constraints, and corner cases without introducing misleading information.
    \item \textbf{Algorithm Description (ALG)}: Records the algorithmic technique(s) and summaries for both the generated and golden editorials.
    \item \textbf{Algorithm Correctness (ALG-COR)}: whether the described method solves the problem under the stated constraints.
\end{itemize}

Figure~\ref{fig:method_overview} (right) summarizes the annotation rubric; full definitions appear in Appendix~\ref{appendix:annotation-rubric}.

\section{Results}


We evaluate 19 contemporary LLMs spanning three groups: (i) proprietary reasoning-oriented systems, (ii) proprietary general-purpose chat models, and (iii) open-weight models over a wide range of scales, including both frontier and open models.\footnote{Model identifiers and inference details are in Appendix~\ref{app:model-cards}.} We report results in \textbf{C++}, the standard language for ICPC-style contests; \textbf{Python generally performs worse} (Appendix~\ref{appendix:python-vs-cpp}).
Table~\ref{tab:model-performance} reports pass@1 across all problems, stratified by difficulty tertiles (T1 easiest–T3 hardest). We compare three conditions (see Section~\ref{sec:Experimental Conditions}): a code-only baseline (w/oEd), a two-stage setting with a model-generated editorial (w/GenEd), and a setting with a gold editorial (w/GoldEd).


\subsection{Overall Performance and Editorial Effects}

\paragraph{Gold editorials isolate an implementation gap.}
Table~\ref{tab:model-performance} shows a clear asymmetry between generated and gold editorials.
On average, providing a gold editorial yields a large absolute gain, indicating that many baseline failures are problem-solving–limited and can be removed by supplying a correct plan.
However, performance remains far from saturated even with gold guidance, especially on T3 (rising only from 8.0\% to 16.5\% on average), which isolates a substantial residual implementation gap: models often struggle to translate a high-level algorithm into correct and efficient code.

\paragraph{The generated--gold editorial gap reflects a problem-solving limitation.}
In contrast, self-generated editorials yield much smaller and less reliable changes, and can even degrade performance for some models.
A few stronger closed-source models, along with DeepSeek~R1, obtain gains of up to $\sim$15\%, but these improvements are concentrated in T1 and T2 and largely disappear on T3.
The gap between w/GenEd and w/GoldEd therefore directly reflects a problem-solving limitation: many models do not reliably produce editorials that fully specify a correct algorithm under the stated constraints.
When the generated editorial is incomplete, overly complex, or subtly wrong, conditioning on it can lock the model into a flawed plan, explaining why some models benefit from self-editorials while others regress.


\paragraph{Only a few models become strongly human-competitive even with gold editorials.}
Figure~\ref{fig:ladder-rank} places each model as a \emph{virtual team} on official contest scoreboards and reports mean rank percentiles across contests.
w/GoldEd consistently improve rankings (up to $\sim$0.4 absolute percentile points), yet most models remain far from top-team performance, despite receiving full solution guidance, whereas human teams solved the same problems under strict contest time limits.
Only a small number of models exceed the mean rank percentile $\sim$0.8 and only slightly more surpass $\sim$0.7, largely only with gold editorials; w/GenEd instead produces smaller (up to $\sim$0.2) and higher-variance shifts.
Overall, correct plans alone are insufficient for most models to match strong human teams, revealing persistent gaps in implementation fidelity and efficiency under contest-style evaluation. Per-contest ladders are shown in Appendix~\ref{appendix:per-contest-rank}; while absolute percentiles vary across contests, the qualitative editorial effects are consistent.

\paragraph{WA dominates, while TLE rises for stronger models (notably Claude).}
Figure~\ref{fig:failure-types} is WA-heavy for nearly all models, showing that most errors are still correctness-limited—either the model does not reach a correct algorithm or it fails to implement a correct plan robustly. However, TLE is disproportionately common for a subset of stronger models, most clearly the Claude variants, suggesting a shift toward problem-solving failures at the complexity level: the approach is often plausible, but misses the key optimization needed to meet contest limits. The remaining CE/RTE mass is smaller but points to residual implementation fragility. A breakdown per (model, setting) is reported in Appendix~\ref{appendix:failure-dist}.



\subsection{Qualitative analysis of editorial behavior}

While the quantitative results show that editorials affect pass@1, they do not explain \emph{why} or where failures occur. To probe model reasoning more directly, we inspect 22 editorials (11 problems $\times$ 2 models) from a single contest (CS3233 2025 Midterm) using the rubric in Appendix~\ref{appendix:annotation-rubric}. This small but indicative case study focuses on two models, the open-weight reasoner \textbf{DeepSeekR1} and the closed-source frontier model \textbf{GPT-5}. Each editorial is annotated for problem understanding, algorithm description, and correctness, and paired with the execution outcome of its w/GenEd code. Full per-editorial annotations appear in Appendix~\ref{app:cs3233-qual-annotations}.

\paragraph{Severe problem understanding errors are rare; the main risk is hallucination.}

Most editorials accurately capture the task requirements: 21 of 22 are not flagged with any wrong or missing crucial details. Only one case exhibits both wrong and missing crucial details with a \emph{major} level of misleading content. In this instance, the model hallucinates an additional constraint absent from the problem statement (see excerpt below).

\begin{tcolorbox}[title={Example: problem understanding error}]

\small
\textit{Problem statement (excerpt):}
“A chain slide is a continuous sequence of slides such that $Y_{i_j}=X_{i_{j+1}}$ and ends at the button the chain slide started from.”

\medskip

\textit{DeepSeekR1 editorial (excerpt):}
“The selected edges must form an \emph{Eulerian subgraph}, meaning that for every node, the in-degree equals the out-degree.”
\end{tcolorbox}

\emph{Explanation.}
In \textsc{chained maimai slides}, the model editorial hallucinates a global conservation constraint (Eulerian in-degree equals out-degree) that is not implied by the problem statement, thereby ruling out valid solutions composed of multiple disjoint cycles.

That said, minor misleading or imprecise statements do appear, usually reflecting algorithmic assumptions or complexity oversights rather than task misunderstandings. In general, the difficulty of understanding the problem is low, with all problems rated as easy or medium.

\paragraph{Reasoning failures often involve wrong plans, missing invariants, or untightened complexity.}

Among the editorials, 15 are labeled algorithmically correct and 7 Incorrect. The incorrect editorials cluster into a small number of recurring patterns. Five are labeled \emph{Wrong algorithm}, where the high-level idea does not solve the full problem, often due to a subtle but decisive reasoning gap. For example, in \textsc{arts and computing students} (DeepSeek-R1), the editorial treats a constrained \emph{shift} as free \emph{movement}, invalidating the construction.

The remaining two incorrect editorials are labeled ``Suboptimal (Likely TLE or MLE), but correct algorithm'': the idea is sound, but the editorial fails to tighten the complexity, and resulting code is asymptotically too slow, leading to timeouts.

\begin{tcolorbox}[title={Example: missing algorithmic invariant}]

\small
\textit{Gold editorial (excerpt):}
“Let $f_i$ be the maximum path length ending at $i$, and let $b_i$ be the maximum path length starting from $i$.”

\medskip

\textit{GPT-5 editorial (excerpt):}
“For each node, we propagate the accumulated value from the root downwards and record the result at that node.”
\end{tcolorbox}

\emph{Explanation.}
In \textsc{Dependency Flood}, the model preserves the correct problem formulation but tracks only forward propagation and omits the backward aggregation, yielding an incomplete algorithmic specification and incorrect results.

\begin{tcolorbox}[title={Example: suboptimal despite correct idea}]

\small
\textit{Gold editorial (excerpt):}
“A convex polyomino in a board with $m$ columns and $n$ rows has at most $\mathcal{O}(m+n)$ boundary cells.”

\medskip

\textit{DeepSeekR1 editorial (excerpt):}
“For each rotation and translation, we count the overlapping border cells by traversing the grid cell by cell.”
\end{tcolorbox}

\emph{Explanation.} In \textsc{Ficketts Conjecture for Polyominoes}, the editorial correctly enumerates rotations and translations but performs an $\mathcal{O}(RC)$ scan per placement, ignoring the boundary-size bound that reduces the intended solution to $\mathcal{O}(R+C)$, leading to likely timeouts.

\paragraph{Correctness $\neq$ clarity: gold editorials assume CP familiarity, while LLMs spell out steps/definitions.}

The annotations reveal a gap between \emph{correctness} and \emph{clarity}. Gold editorials serve as a reference for algorithmic correctness but are not always the most operationally clear. In \textsc{dependency flood} (DeepSeek-R1), the model editorial is labeled “Same as golden” yet judged easier to grasp than the official explanation, despite implementing the same idea. A similar pattern appears in \textsc{hungry piplups} (GPT-5), where the model’s editorial explains each component step by step, while the gold editorial assumes familiarity with standard CP abstractions (see Appendix~\ref{app:examples}). Consistent with this, model-generated editorials are substantially longer than gold editorials across models (Appendix~\ref{appendix:editorial-length}), suggesting a trade-off between concision and explicit reasoning.

\paragraph{Paradigm tags are only weakly informative: mismatches and false positives both occur.}
Algorithmic tags provide a coarse check of whether a model editorial operates in the same broad paradigm as the gold editorial. On the annotated set, DeepSeek R1 matches the gold tags on 6 of 11 problems, and GPT-5 on 7 of 11. When tags differ, models often omit a key paradigm, for example treating \textsc{hungry piplups} as simulation or greedy reasoning rather than the segment-tree and binary-search approach in the gold editorial. However, tag mismatch does not necessarily imply incorrectness, as in \textsc{easygoing workplace} (both), while tag agreement does not guarantee correctness, as in \textsc{dependency flood} (GPT-5). Overall, tag alignment is only a weak signal of editorial correctness.

\paragraph{Algorithmic correctness is necessary, not sufficient: implementation remains a gap.}
Algorithmic correctness at the editorial level strongly predicts downstream results. Of the 15 editorials labeled algorithmically correct, 12 lead to accepted submissions; the remaining cases fail due to efficiency or robustness issues (two TLEs, one RTE). None of the 7 editorials labeled Incorrect pass. Correct reasoning is therefore necessary but not sufficient: even with a sound plan, models may fail to produce efficient or robust implementations


Overall, the qualitative analysis supports four conclusions: (i) editorial-level algorithmic correctness strongly predicts success; (ii) most reasoning failures arise from incorrect or incomplete plans rather than problem misunderstanding; (iii) implementation efficiency and robustness remain major bottlenecks even with correct editorials; and (iv) while gold editorials provide a reliable correctness reference, model-generated editorials can sometimes be clearer and equally effective.

\begin{figure*}[t]
\centering

\begin{minipage}[t]{0.48\textwidth}
  \centering
  \includegraphics[width=\linewidth,trim=0 5 0 6,clip]{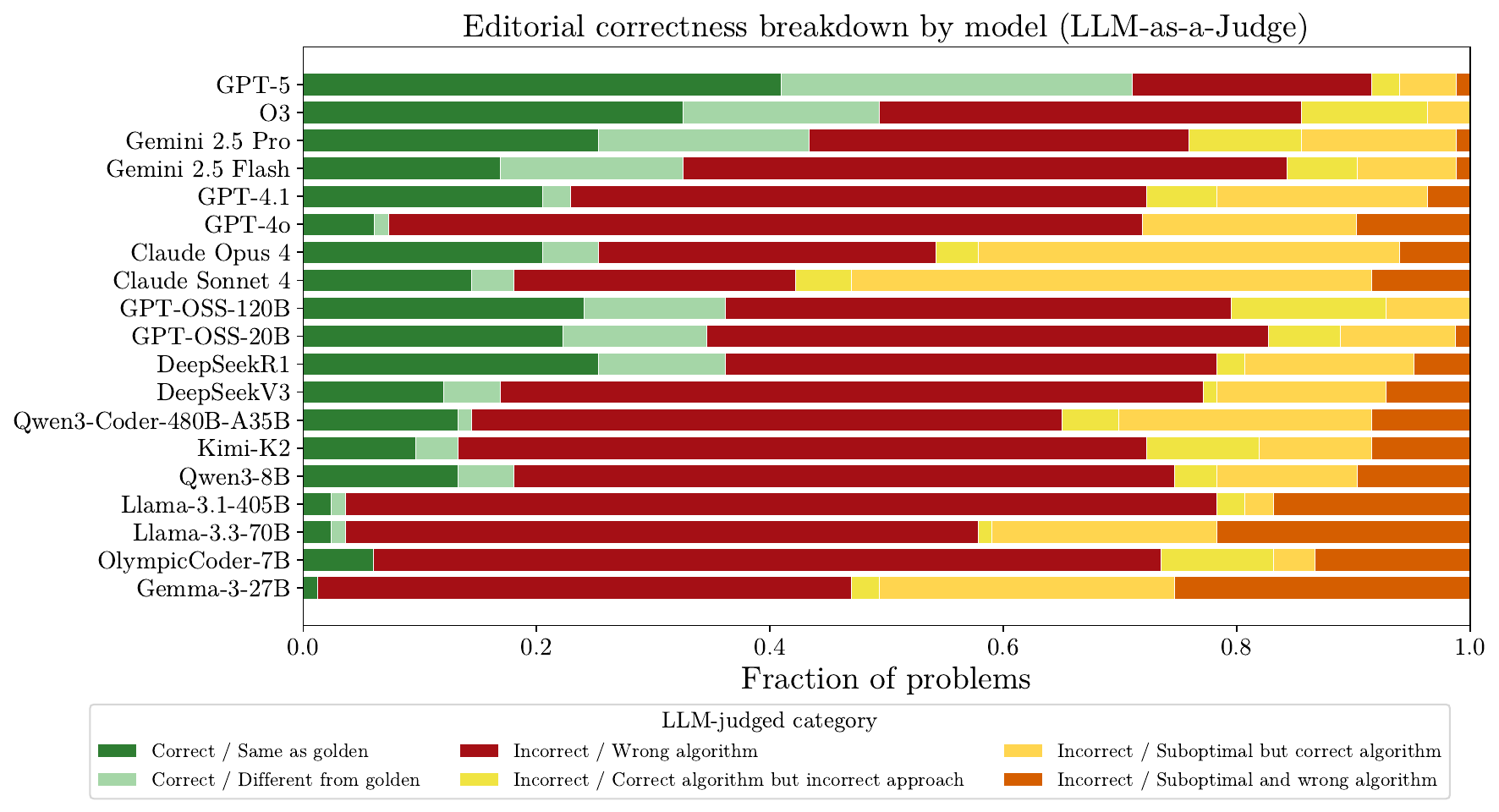}
  \captionof{figure}{Six-way editorial correctness breakdown labeled by the LLM-as-a-judge. Frontier models produce more judge-\textsc{Correct} plans, but \emph{wrong algorithm}—i.e., incorrect problem solving—remains the dominant error across many models.}
  \label{fig:llm-judge-by-model}
\end{minipage}
\hfill
\begin{minipage}[t]{0.48\textwidth}
  \centering
  \includegraphics[width=\linewidth,trim=0 6 0 8,clip]{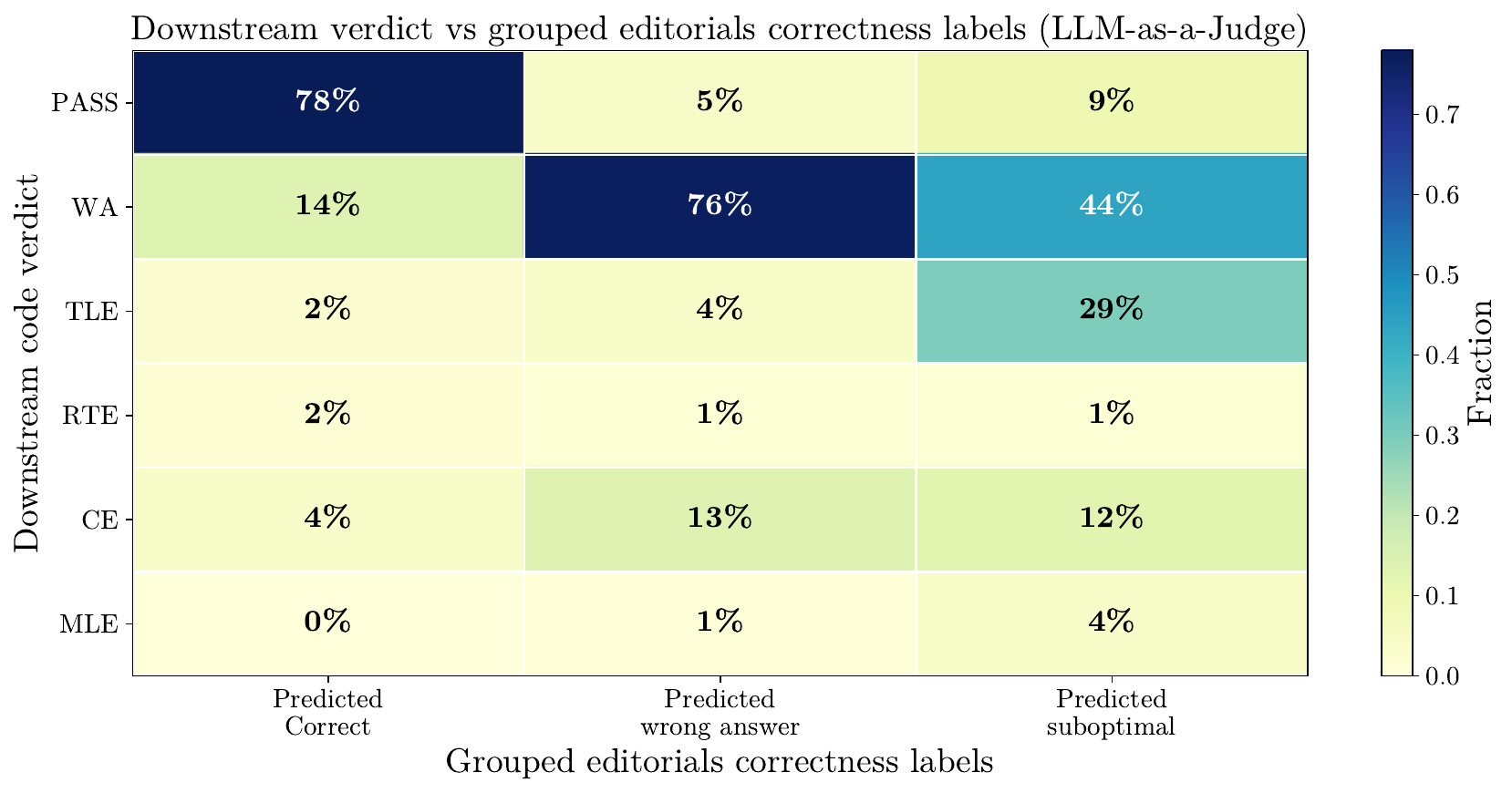}
  \captionof{figure}{Downstream verdict distribution (PASS/WA/TLE/RTE/CE/MLE) conditioned on editorial correctness labels. Editorial correctness labels meaningfully stratify downstream outcomes.}
  \label{fig:llm-judge-vs-verdict}
\end{minipage}

\vspace{-10pt}
\end{figure*}

\subsection{LLM-as-a-Judge Editorial Diagnostics}
\label{subsec:llm-judge}

Expert editorial annotations provide high-fidelity insight into algorithmic reasoning failures, but are expensive to scale.
We therefore adopt an \emph{LLM-as-a-judge} to label \emph{all} generated editorials using the same rubric as the expert annotator (Section~\ref{subsec:annotation}), guided by a gold editorial that provides a consistent reference for correctness and error diagnosis.

\paragraph{Judge setup.}
For each \textbf{w/GenEd} run, the judge is given the problem statement $P$, the gold editorial $E^\ast$, and the model-generated editorial $E$, and outputs structured labels.
We use \texttt{google/gemini-3-pro-preview} \cite{google_deepmind_gemini3pro_model_card} as the judge for reliable long-context rubric following and structured outputs, and because it is not part of the evaluated LLMs (Table~\ref{tab:model-performance}).
The full LLM-as-a-judge prompt and output schema are provided in Appendix~\ref{appendix:llm-judge-prompts}.

\paragraph{Validation against expert labels.}

We validate the judge on an expert-annotated subset of the CS3233 2025 Midterm.
Table~\ref{tab:llm-judge-agreement} reports agreement on the core correctness signal (\textsc{ALG-COR}) and the conditional diagnostics used in our analysis.
We report accuracy and Cohen’s $\kappa$, with conditionally-defined fields evaluated only where applicable under the expert label.
The judge shows strong agreement on overall correctness and moderate agreement on both \emph{correct type} and \emph{why incorrect}, indicating reliable identification of incorrectness and its primary cause.
Agreement on \emph{severity} is weaker; we therefore treat severity as exploratory. See Appendix~\ref{app:additional-agreement} for results on auxiliary rubric fields.

\begin{table}[t]
\centering
\small
\vspace{-3pt}
\setlength{\tabcolsep}{4pt}
\begin{tabular}{lccc}
\toprule
\textbf{Rubric field} & \textbf{$n$} & \textbf{Agr.} & \textbf{$\kappa$} \\
\midrule
ALG-COR (Correct vs.\ Incorrect) & 22 & 0.818 & 0.611 \\
\midrule
Correct type (given expert=Correct) & 15 & 0.733 & 0.474 \\
Why incorrect (given expert=Incorrect) & 7 & 0.714 & 0.500 \\
Severity (given expert=Incorrect) & 7 & 0.429 & 0.152 \\
\bottomrule
\end{tabular}
\caption{
Agreement between expert annotator and Gemini 3 Pro judge on the validation subset.
The LLM judge is reliable for overall correctness and for identifying why an editorial is incorrect; severity is less consistent.
}
\label{tab:llm-judge-agreement}
\end{table}

\paragraph{A persistent problem-solving gap across models.}
Figure~\ref{fig:llm-judge-by-model} decomposes \textbf{w/GenEd} editorials by the six-way \textsc{ALG-COR} taxonomy.
Frontier models produce a substantially larger share of editorials judged \textsc{Correct} (both \emph{same as} and \emph{different from} the gold approach), yet even for these models a large fraction of editorials remain incorrect.
Among incorrect cases, \emph{wrong algorithm} dominates for most models—particularly open-weight systems—indicating that failures under \textbf{w/GenEd} are primarily due to incorrect problem-solving plans rather than purely implementation errors.
Claude variants form a notable exception, allocating comparatively more mass to \emph{suboptimal but correct} editorials, consistent with reasoning that captures the right idea but fails to tighten complexity or efficiency arguments.

\paragraph{Judge labels predict downstream failure modes.}
Figure~\ref{fig:llm-judge-vs-verdict} (right) shows that judge labels translate into distinct execution outcomes for the downstream \textbf{w/GenEd} code.
For readability, we collapse the six \textsc{ALG-COR} categories into three groups:
\emph{Correct} (same/different), \emph{Predicted WA} (wrong algorithm or incorrect approach, suboptimal and wrong)), and \emph{Predicted suboptimal} (suboptimal but correct).
When the editorial is judged \emph{Correct}, the downstream program is accepted most of the time, but the remaining WA/CE/RTE/TLE tail directly exposes an \textbf{implementation gap} even under a sound plan.
When the editorial is \emph{Predicted WA}, failures are overwhelmingly WA (with a small PASS tail), indicating that editorial-level reasoning mistakes largely propagate into incorrect outputs.
When the editorial is \emph{Predicted suboptimal}, TLE is strongly enriched (while WA remains common), providing a sanity check that the judge’s efficiency-related diagnoses align with runtime failures.
See Appendix~\ref{app:llm-judge-extra-diagnostics} for more analysis: problem-understanding errors are relatively rare but consequential, since correct understanding is the bare minimum in competitive programming.
Algorithm-tag alignment is a weak proxy for success compared to fine-grained \textsc{ALG-COR} labels.


\subsection{Cross-model editorial transfer}
\label{subsec:cross-editorial}

Our annotations suggest that model-generated editorials are often easier to follow than the accompanying gold editorials, motivating a transfer question: can an editorial written by one model benefit a different model that only acts as a coder?

So far, models were evaluated in a \emph{matched} configuration, where the same system supplies both $f_{\mathrm{ed}}$ and $f_{\mathrm{code}}$. Here, we decouple these roles by treating models as interchangeable \emph{writers} and \emph{coders}. Given a problem $P$, a writer $m_{\mathrm{w}}$ produces an editorial $E^{(m_{\mathrm{w}})}$, and a coder $m_{\mathrm{c}}$ generates code conditioned on it: \(C^{(m_{\mathrm{w}} \rightarrow m_{\mathrm{c}})} =
f_{\mathrm{code}}^{(m_{\mathrm{c}})}\!\bigl(P, E^{(m_{\mathrm{w}})}\bigr).
\)When $m_{\mathrm{w}} = m_{\mathrm{c}}$, this recovers w/GenEd; replacing $E^{(m_{\mathrm{w}})}$ with a gold editorial yields w/GoldEd.

\begin{figure}[t]
    \centering
    \includegraphics[width=\columnwidth]{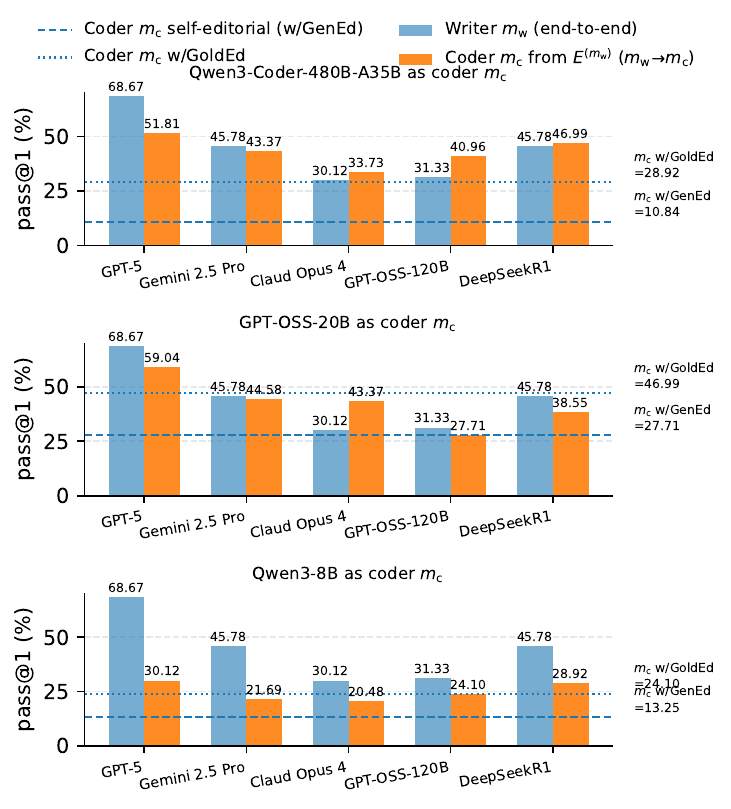}
\caption{ Cross-model editorial transfer.
Each bar reports pass@1 when a fixed coder implements an editorial written by a different model.
Using editorials from stronger writers often improves weaker coders and can occasionally yield
performance competitive with or exceeding the writer’s own end-to-end results.
}

    \label{fig:cross-editorial}
\end{figure}

We evaluate this cross-model setting using three open-weight coders (Qwen3-Coder-480B-A35B, GPT-OSS-20B, Qwen3-8B) and several strong writers (DeepSeek-R1, Claude Opus~4, Gemini~2.5 Pro, GPT-5, GPT-OSS-120B). Figure~\ref{fig:cross-editorial} compares each coder’s self-editorial and gold-editorial baselines to cross-model compositions.

\paragraph{Editorial transfer improves coding.}
Across all coders, every cross-model configuration performs at least as well as the coder’s w/GenEd, and many substantially outperform it. For larger coders, editorials from the strongest writers often recover much of the writer’s own end-to-end performance and can even exceed the coder’s w/GoldEd score; in some cases, a weaker coder implements a stronger model’s plan more effectively than the writer itself. Overall, these results show that reasoning and implementation can be modularized: pairing a strong writer with a competent coder can outperform either model used end-to-end, with editorials serving as a simple, model-agnostic interface.

\section{Related Work}

\paragraph{Competitive Programming Benchmarks}
Programming LLMs are often evaluated end-to-end on unit tests (APPS \cite{hendrycksapps2021}, HumanEval \cite{chen2021codex})
and, more recently, in contest-style settings (AlphaCode/CodeContests \cite{Li_2022}, LiveCodeBench \cite{jain2024livecodebench},
USACO \cite{shi2024languagemodelssolveolympiad}, LLM-ProS \cite{hossain2025llmprosanalyzinglargelanguage}, CodeElo \cite{quan2025codeelo}).
Closest to our goal of disentangling steps, Coding Triangle evaluates editorials, code, and tests \cite{zhang2025codingtriangledoeslarge}, and
\citet{yang-etal-2025-elaboration} studies staged competitive-programming workflows. We focus on editorials as an explicit,
transferable plan separating reasoning from implementation.

\paragraph{Code LLMs and Reasoning LLMs}
Progress spans code-specialized models (e.g., CodeGeeX \cite{zheng2024codegeexpretrainedmodelcode}, StarCoder \cite{li2023starcodersourceyou},
Code Llama \cite{rozière2024codellamaopenfoundation}) and reasoning-oriented systems (OpenAI's O-series \cite{openai2025competitiveprogramminglargereasoning},
DeepSeek-R1 \cite{deepseekai2025deepseekr1incentivizingreasoningcapability}). Our editorial-centric pipeline evaluates plan quality and implementation fidelity
independently and enables writer--coder composition.




\noindent Appendix~\ref{app:extended-related} provides additional related work.

\section{Conclusion}

We revisited competitive-programming evaluation for large language models by treating editorials as explicit artifacts that separate algorithmic reasoning from implementation. Across 83 ICPC-style problems and 19 models, gold editorials consistently yield large gains in pass@1 and virtual rank, while self-generated editorials have smaller and less reliable effects. Even with correct editorials, performance remains far from saturated, with failures dominated by wrong answers and timeouts, underscoring persistent implementation bottlenecks.

We also show that editorials transfer effectively across models: editorials from strong reasoners can boost weaker coders and sometimes outperform the writer’s own end-to-end performance. Together, these results suggest that reasoning and implementation can be modularized, with editorials serving as a model-agnostic interface.

\section{Limitations}

We focus on a constrained but high-quality evaluation setting. All main experiments are conducted in C++, the dominant ICPC contest language; Python results are reported separately in Appendix~\ref{appendix:python-vs-cpp} and exhibit systematically lower pass rates. The dataset is relatively small, spanning seven contests and 83 problems, and may not generalize to other contests, programming languages, or problem formats. However, this scale is a deliberate tradeoff: curating competitive-programming problems with \emph{complete official test suites} and \emph{expert-written gold editorials}, while minimizing contamination risk from widely scraped online judges, substantially constrains dataset size.

Expert annotation further limits coverage. The pool of qualified annotators is very small—typically restricted to elite competitive programmers such as ICPC World Finalists—and producing editorial-level annotation under our rubric can take several hours. As a result, expert annotations are limited to a single contest and a single annotator, and we treat one gold editorial per problem as the reference despite the existence of multiple valid solution strategies.

All evaluations use a single-shot deterministic setup with minimal prompting, without sampling, tool use, or editorial-aware fine-tuning, aside from a small appendix-level exploration of test-time feedback (Appendix~\ref{app:feedback-ablation}). Consequently, our results should be interpreted as evidence for the value of editorial-centric evaluation and modular reasoning–implementation separation, rather than as a definitive ranking of models across all competitive-programming settings.

\section{Ethical Considerations}

This paper studies large language models (LLMs) in competitive programming (CP) via an editorial-centric evaluation
protocol that separates \emph{problem solving} (deriving an algorithm) from \emph{implementation} (writing correct and
efficient code). The work is methodological and diagnostic: we do not deploy models in real-world decision-making.

\paragraph{Academic integrity and misuse.}
Improved CP capability may be misused to cheat in coursework or contests. Our benchmark is built from \emph{past}
contests and course materials and is intended for research and post-hoc benchmarking, not for use in active
competitions. We do not provide tooling designed to bypass contest rules. Because releasing complete test suites can
reduce the future usefulness of these problems for hidden-test assessments, we recommend that educators avoid
reusing included problems as undisclosed evaluation items.

\paragraph{Dataset provenance, permissions, and privacy.}
Our dataset consists of problem statements, gold editorials, and official test suites from seven sources
(Appendix~\ref{app:data-details}). The CS3233 (NUS) portion includes course assessment materials; we obtained
permission from the course instructor to redistribute the problem statements and gold editorials. Other contests were
sourced from publicly accessible task repositories; we preserve attribution and provide provenance metadata for each
problem. The dataset contains no personal data about participants, students, or annotators; we report aggregate
statistics and do not attempt to infer or disclose private information.

\paragraph{Safety of executing model-generated code and evaluation limitations.}
We compile and run untrusted, model-generated programs in a sandboxed ICPC-style judging pipeline with strict time
and memory limits (Appendix~\ref{app:judging}) and do not deploy generated code in production systems. Our benchmark
reflects ICPC-style constraints and is primarily evaluated in C++ (Appendix~\ref{appendix:python-vs-cpp}); conclusions
should be interpreted as CP-style reasoning and implementation performance rather than general-purpose software
engineering ability.

\section{Use of Generative AI}

We used generative AI tools for language-level assistance (e.g., copy-editing, rephrasing, and improving clarity) on
portions of the manuscript. All research ideas, experimental design, dataset construction, implementation of the
evaluation pipeline, quantitative analyses, and conclusions were developed by the authors. We did not use generative
AI to create or alter experimental results, to fabricate citations, or to replace expert validation; all citations
and factual claims were checked by the authors.

\bibliography{custom}

@article{hendrycksapps2021,
  title={Measuring Coding Challenge Competence With APPS},
  author={Dan Hendrycks and Steven Basart and Saurav Kadavath and Mantas Mazeika and Akul Arora and Ethan Guo and Collin Burns and Samir Puranik and Horace He and Dawn Song and Jacob Steinhardt},
  journal={NeurIPS},
  year={2021}
}

@article{chen2021codex,
  title={Evaluating Large Language Models Trained on Code},
  author={Mark Chen and Jerry Tworek and Heewoo Jun and Qiming Yuan and Henrique Ponde de Oliveira Pinto and Jared Kaplan and Harri Edwards and Yuri Burda and Nicholas Joseph and Greg Brockman and Alex Ray and Raul Puri and Gretchen Krueger and Michael Petrov and Heidy Khlaaf and Girish Sastry and Pamela Mishkin and Brooke Chan and Scott Gray and Nick Ryder and Mikhail Pavlov and Alethea Power and Lukasz Kaiser and Mohammad Bavarian and Clemens Winter and Philippe Tillet and Felipe Petroski Such and Dave Cummings and Matthias Plappert and Fotios Chantzis and Elizabeth Barnes and Ariel Herbert-Voss and William Hebgen Guss and Alex Nichol and Alex Paino and Nikolas Tezak and Jie Tang and Igor Babuschkin and Suchir Balaji and Shantanu Jain and William Saunders and Christopher Hesse and Andrew N. Carr and Jan Leike and Josh Achiam and Vedant Misra and Evan Morikawa and Alec Radford and Matthew Knight and Miles Brundage and Mira Murati and Katie Mayer and Peter Welinder and Bob McGrew and Dario Amodei and Sam McCandlish and Ilya Sutskever and Wojciech Zaremba},
  year={2021},
  eprint={2107.03374},
  archivePrefix={arXiv},
  primaryClass={cs.LG}
}

@inproceedings{
jain2024livecodebench,
title={LiveCodeBench: Holistic and Contamination Free Evaluation of Large Language Models for Code},
author={Naman Jain and King Han and Alex Gu and Wen-Ding Li and Fanjia Yan and Tianjun Zhang and Sida Wang and Armando Solar-Lezama and Koushik Sen and Ion Stoica},
booktitle={The Thirteenth International Conference on Learning Representations},
year={2025},
url={https://openreview.net/forum?id=chfJJYC3iL}
}

@article{quan2025codeelo,
  title={CodeElo: Benchmarking Competition-level Code Generation of LLMs with Human-comparable Elo Ratings},
  author={Quan, Shanghaoran and Yang, Jiaxi and Yu, Bowen and Zheng, Bo and Liu, Dayiheng and Yang, An and Ren, Xuancheng and Gao, Bofei and Miao, Yibo and Feng, Yunlong and others},
  journal={arXiv preprint arXiv:2501.01257},
  year={2025}
}

@inproceedings{
shi2024languagemodelssolveolympiad,
title={Can Language Models Solve Olympiad Programming?},
author={Ben Shi and Michael Tang and Karthik R Narasimhan and Shunyu Yao},
booktitle={First Conference on Language Modeling},
year={2024},
url={https://openreview.net/forum?id=kGa4fMtP9l}
}

@INPROCEEDINGS{hossain2025llmprosanalyzinglargelanguage,
  author={Hossain, Md Sifat and Tabassum, Anika and Arefin, Md. Fahim and Shaila Zaman, Tarannum},
  booktitle={2025 IEEE/ACM International Workshop on Large Language Models for Code (LLM4Code)}, 
  title={LLM-ProS: Analyzing Large Language Models’ Performance in Competitive Problem Solving}, 
  year={2025},
  volume={},
  number={},
  pages={80-87},
  doi={10.1109/LLM4Code66737.2025.00015}}

@misc{zheng2024codegeexpretrainedmodelcode,
      title={CodeGeeX: A Pre-Trained Model for Code Generation with Multilingual Benchmarking on HumanEval-X}, 
      author={Qinkai Zheng and Xiao Xia and Xu Zou and Yuxiao Dong and Shan Wang and Yufei Xue and Zihan Wang and Lei Shen and Andi Wang and Yang Li and Teng Su and Zhilin Yang and Jie Tang},
      year={2024},
      eprint={2303.17568},
      archivePrefix={arXiv},
      primaryClass={cs.LG},
      url={https://arxiv.org/abs/2303.17568}, 
}

@misc{li2023starcodersourceyou,
      title={StarCoder: may the source be with you!}, 
      author={Raymond Li and Loubna Ben Allal and Yangtian Zi and Niklas Muennighoff and Denis Kocetkov and Chenghao Mou and Marc Marone and Christopher Akiki and Jia Li and Jenny Chim and Qian Liu and Evgenii Zheltonozhskii and Terry Yue Zhuo and Thomas Wang and Olivier Dehaene and Mishig Davaadorj and Joel Lamy-Poirier and João Monteiro and Oleh Shliazhko and Nicolas Gontier and Nicholas Meade and Armel Zebaze and Ming-Ho Yee and Logesh Kumar Umapathi and Jian Zhu and Benjamin Lipkin and Muhtasham Oblokulov and Zhiruo Wang and Rudra Murthy and Jason Stillerman and Siva Sankalp Patel and Dmitry Abulkhanov and Marco Zocca and Manan Dey and Zhihan Zhang and Nour Fahmy and Urvashi Bhattacharyya and Wenhao Yu and Swayam Singh and Sasha Luccioni and Paulo Villegas and Maxim Kunakov and Fedor Zhdanov and Manuel Romero and Tony Lee and Nadav Timor and Jennifer Ding and Claire Schlesinger and Hailey Schoelkopf and Jan Ebert and Tri Dao and Mayank Mishra and Alex Gu and Jennifer Robinson and Carolyn Jane Anderson and Brendan Dolan-Gavitt and Danish Contractor and Siva Reddy and Daniel Fried and Dzmitry Bahdanau and Yacine Jernite and Carlos Muñoz Ferrandis and Sean Hughes and Thomas Wolf and Arjun Guha and Leandro von Werra and Harm de Vries},
      year={2023},
      eprint={2305.06161},
      archivePrefix={arXiv},
      primaryClass={cs.CL},
      url={https://arxiv.org/abs/2305.06161}, 
}

@misc{rozière2024codellamaopenfoundation,
      title={Code Llama: Open Foundation Models for Code}, 
      author={Baptiste Rozière and Jonas Gehring and Fabian Gloeckle and Sten Sootla and Itai Gat and Xiaoqing Ellen Tan and Yossi Adi and Jingyu Liu and Romain Sauvestre and Tal Remez and Jérémy Rapin and Artyom Kozhevnikov and Ivan Evtimov and Joanna Bitton and Manish Bhatt and Cristian Canton Ferrer and Aaron Grattafiori and Wenhan Xiong and Alexandre Défossez and Jade Copet and Faisal Azhar and Hugo Touvron and Louis Martin and Nicolas Usunier and Thomas Scialom and Gabriel Synnaeve},
      year={2024},
      eprint={2308.12950},
      archivePrefix={arXiv},
      primaryClass={cs.CL},
      url={https://arxiv.org/abs/2308.12950}, 
}

@misc{penedo2025openr1update3,
  author       = {Guilherme Penedo and Lewis Tunstall and Anton Lozhkov and Hynek Kydlicek and
                  Edward Beeching and Loubna Ben Allal and Quentin Gallouédec and
                  Leandro von Werra and Agustín Piqueres Lajarín and Nathan Habib and others},
  title        = {Open R1: Update \#3},
  howpublished = {\url{https://huggingface.co/blog/open-r1/update-3}},
  note         = {Hugging Face Blog},
  year         = {2025},
  month        = March,
  day          = {11}
}

@article{Li_2022,
   title={Competition-level code generation with AlphaCode},
   volume={378},
   ISSN={1095-9203},
   url={http://dx.doi.org/10.1126/science.abq1158},
   DOI={10.1126/science.abq1158},
   number={6624},
   journal={Science},
   publisher={American Association for the Advancement of Science (AAAS)},
   author={Li, Yujia and Choi, David and Chung, Junyoung and Kushman, Nate and Schrittwieser, Julian and Leblond, Rémi and Eccles, Tom and Keeling, James and Gimeno, Felix and Dal Lago, Agustin and Hubert, Thomas and Choy, Peter and de Masson d’Autume, Cyprien and Babuschkin, Igor and Chen, Xinyun and Huang, Po-Sen and Welbl, Johannes and Gowal, Sven and Cherepanov, Alexey and Molloy, James and Mankowitz, Daniel J. and Sutherland Robson, Esme and Kohli, Pushmeet and de Freitas, Nando and Kavukcuoglu, Koray and Vinyals, Oriol},
   year={2022},
   month=dec, pages={1092–1097} }

@misc{openai2025icpc,
  title        = {OpenAI Joins Inaugural AI Tools Experiment at the 2025 ICPC World Finals},
  author       = {{International Collegiate Programming Contest (ICPC)}},
  year         = {2025},
  howpublished = {\url{https://worldfinals.icpc.global/2025/openai.html}},
  month        = sep,
}

@online{deepmind2025gemini,
  author  = {Hanzhao (Maggie) Lin and Heng{-}Tze Cheng},
  title        = {Gemini Achieves Gold-Medal-Level Performance at the International Collegiate Programming Contest World Finals},
  year    = {2025},
  month   = sep,
  url     = {https://deepmind.google/discover/blog/gemini-achieves-gold-level-performance-at-the-international-collegiate-programming-contest-world-finals/},
  note    = {Google DeepMind Blog, 17 Sep 2025}
}

@misc{openai2025competitiveprogramminglargereasoning,
      title={Competitive Programming with Large Reasoning Models}, 
      author={OpenAI and : and Ahmed El-Kishky and Alexander Wei and Andre Saraiva and Borys Minaiev and Daniel Selsam and David Dohan and Francis Song and Hunter Lightman and Ignasi Clavera and Jakub Pachocki and Jerry Tworek and Lorenz Kuhn and Lukasz Kaiser and Mark Chen and Max Schwarzer and Mostafa Rohaninejad and Nat McAleese and o3 contributors and Oleg Mürk and Rhythm Garg and Rui Shu and Szymon Sidor and Vineet Kosaraju and Wenda Zhou},
      year={2025},
      eprint={2502.06807},
      archivePrefix={arXiv},
      primaryClass={cs.LG},
      url={https://arxiv.org/abs/2502.06807}, 
}

@misc{deepseekai2025deepseekr1incentivizingreasoningcapability,
      title={DeepSeek-R1: Incentivizing Reasoning Capability in LLMs via Reinforcement Learning}, 
      author={DeepSeek-AI and Daya Guo and Dejian Yang and Haowei Zhang and Junxiao Song and Ruoyu Zhang and Runxin Xu and Qihao Zhu and Shirong Ma and Peiyi Wang and Xiao Bi and Xiaokang Zhang and Xingkai Yu and Yu Wu and Z. F. Wu and Zhibin Gou and Zhihong Shao and Zhuoshu Li and Ziyi Gao and Aixin Liu and Bing Xue and Bingxuan Wang and Bochao Wu and Bei Feng and Chengda Lu and Chenggang Zhao and Chengqi Deng and Chenyu Zhang and Chong Ruan and Damai Dai and Deli Chen and Dongjie Ji and Erhang Li and Fangyun Lin and Fucong Dai and Fuli Luo and Guangbo Hao and Guanting Chen and Guowei Li and H. Zhang and Han Bao and Hanwei Xu and Haocheng Wang and Honghui Ding and Huajian Xin and Huazuo Gao and Hui Qu and Hui Li and Jianzhong Guo and Jiashi Li and Jiawei Wang and Jingchang Chen and Jingyang Yuan and Junjie Qiu and Junlong Li and J. L. Cai and Jiaqi Ni and Jian Liang and Jin Chen and Kai Dong and Kai Hu and Kaige Gao and Kang Guan and Kexin Huang and Kuai Yu and Lean Wang and Lecong Zhang and Liang Zhao and Litong Wang and Liyue Zhang and Lei Xu and Leyi Xia and Mingchuan Zhang and Minghua Zhang and Minghui Tang and Meng Li and Miaojun Wang and Mingming Li and Ning Tian and Panpan Huang and Peng Zhang and Qiancheng Wang and Qinyu Chen and Qiushi Du and Ruiqi Ge and Ruisong Zhang and Ruizhe Pan and Runji Wang and R. J. Chen and R. L. Jin and Ruyi Chen and Shanghao Lu and Shangyan Zhou and Shanhuang Chen and Shengfeng Ye and Shiyu Wang and Shuiping Yu and Shunfeng Zhou and Shuting Pan and S. S. Li and Shuang Zhou and Shaoqing Wu and Shengfeng Ye and Tao Yun and Tian Pei and Tianyu Sun and T. Wang and Wangding Zeng and Wanjia Zhao and Wen Liu and Wenfeng Liang and Wenjun Gao and Wenqin Yu and Wentao Zhang and W. L. Xiao and Wei An and Xiaodong Liu and Xiaohan Wang and Xiaokang Chen and Xiaotao Nie and Xin Cheng and Xin Liu and Xin Xie and Xingchao Liu and Xinyu Yang and Xinyuan Li and Xuecheng Su and Xuheng Lin and X. Q. Li and Xiangyue Jin and Xiaojin Shen and Xiaosha Chen and Xiaowen Sun and Xiaoxiang Wang and Xinnan Song and Xinyi Zhou and Xianzu Wang and Xinxia Shan and Y. K. Li and Y. Q. Wang and Y. X. Wei and Yang Zhang and Yanhong Xu and Yao Li and Yao Zhao and Yaofeng Sun and Yaohui Wang and Yi Yu and Yichao Zhang and Yifan Shi and Yiliang Xiong and Ying He and Yishi Piao and Yisong Wang and Yixuan Tan and Yiyang Ma and Yiyuan Liu and Yongqiang Guo and Yuan Ou and Yuduan Wang and Yue Gong and Yuheng Zou and Yujia He and Yunfan Xiong and Yuxiang Luo and Yuxiang You and Yuxuan Liu and Yuyang Zhou and Y. X. Zhu and Yanhong Xu and Yanping Huang and Yaohui Li and Yi Zheng and Yuchen Zhu and Yunxian Ma and Ying Tang and Yukun Zha and Yuting Yan and Z. Z. Ren and Zehui Ren and Zhangli Sha and Zhe Fu and Zhean Xu and Zhenda Xie and Zhengyan Zhang and Zhewen Hao and Zhicheng Ma and Zhigang Yan and Zhiyu Wu and Zihui Gu and Zijia Zhu and Zijun Liu and Zilin Li and Ziwei Xie and Ziyang Song and Zizheng Pan and Zhen Huang and Zhipeng Xu and Zhongyu Zhang and Zhen Zhang},
      year={2025},
      eprint={2501.12948},
      archivePrefix={arXiv},
      primaryClass={cs.CL},
      url={https://arxiv.org/abs/2501.12948}, 
}

@misc{comanici2025gemini25pushingfrontier,
      title={Gemini 2.5: Pushing the Frontier with Advanced Reasoning, Multimodality, Long Context, and Next Generation Agentic Capabilities}, 
      author={Gheorghe Comanici and Eric Bieber and Mike Schaekermann and Ice Pasupat and Noveen Sachdeva and Inderjit Dhillon and Marcel Blistein and Ori Ram and Dan Zhang and Evan Rosen and Luke Marris and Sam Petulla and Colin Gaffney and Asaf Aharoni and Nathan Lintz and Tiago Cardal Pais and Henrik Jacobsson and Idan Szpektor and Nan-Jiang Jiang and Krishna Haridasan and Ahmed Omran and Nikunj Saunshi and Dara Bahri and Gaurav Mishra and Eric Chu and Toby Boyd and Brad Hekman and Aaron Parisi and Chaoyi Zhang and Kornraphop Kawintiranon and Tania Bedrax-Weiss and Oliver Wang and Ya Xu and Ollie Purkiss and Uri Mendlovic and Ilaï Deutel and Nam Nguyen and Adam Langley and Flip Korn and Lucia Rossazza and Alexandre Ramé and Sagar Waghmare and Helen Miller and Nathan Byrd and Ashrith Sheshan and Raia Hadsell Sangnie Bhardwaj and Pawel Janus and Tero Rissa and Dan Horgan and Sharon Silver and Ayzaan Wahid and Sergey Brin and Yves Raimond and Klemen Kloboves and Cindy Wang and Nitesh Bharadwaj Gundavarapu and Ilia Shumailov and Bo Wang and Mantas Pajarskas and Joe Heyward and Martin Nikoltchev and Maciej Kula and Hao Zhou and Zachary Garrett and Sushant Kafle and Sercan Arik and Ankita Goel and Mingyao Yang and Jiho Park and Koji Kojima and Parsa Mahmoudieh and Koray Kavukcuoglu and Grace Chen and Doug Fritz and Anton Bulyenov and Sudeshna Roy and Dimitris Paparas and Hadar Shemtov and Bo-Juen Chen and Robin Strudel and David Reitter and Aurko Roy and Andrey Vlasov and Changwan Ryu and Chas Leichner and Haichuan Yang and Zelda Mariet and Denis Vnukov and Tim Sohn and Amy Stuart and Wei Liang and Minmin Chen and Praynaa Rawlani and Christy Koh and JD Co-Reyes and Guangda Lai and Praseem Banzal and Dimitrios Vytiniotis and Jieru Mei and Mu Cai and Mohammed Badawi and Corey Fry and Ale Hartman and Daniel Zheng and Eric Jia and James Keeling and Annie Louis and Ying Chen and Efren Robles and Wei-Chih Hung and Howard Zhou and Nikita Saxena and Sonam Goenka and Olivia Ma and Zach Fisher and Mor Hazan Taege and Emily Graves and David Steiner and Yujia Li and Sarah Nguyen and Rahul Sukthankar and Joe Stanton and Ali Eslami and Gloria Shen and Berkin Akin and Alexey Guseynov and Yiqian Zhou and Jean-Baptiste Alayrac and Armand Joulin and Efrat Farkash and Ashish Thapliyal and Stephen Roller and Noam Shazeer and Todor Davchev and Terry Koo and Hannah Forbes-Pollard and Kartik Audhkhasi and Greg Farquhar and Adi Mayrav Gilady and Maggie Song and John Aslanides and Piermaria Mendolicchio and Alicia Parrish and John Blitzer and Pramod Gupta and Xiaoen Ju and Xiaochen Yang and Puranjay Datta and Andrea Tacchetti and Sanket Vaibhav Mehta and Gregory Dibb and Shubham Gupta and Federico Piccinini and Raia Hadsell and Sujee Rajayogam and Jiepu Jiang and Patrick Griffin and Patrik Sundberg and Jamie Hayes and Alexey Frolov and Tian Xie and Adam Zhang and Kingshuk Dasgupta and Uday Kalra and Lior Shani and Klaus Macherey and Tzu-Kuo Huang and Liam MacDermed and Karthik Duddu and Paulo Zacchello and Zi Yang and Jessica Lo and Kai Hui and Matej Kastelic and Derek Gasaway and Qijun Tan and Summer Yue and Pablo Barrio and John Wieting and Weel Yang and Andrew Nystrom and Solomon Demmessie and Anselm Levskaya and Fabio Viola and Chetan Tekur and Greg Billock and George Necula and Mandar Joshi and Rylan Schaeffer and Swachhand Lokhande and Christina Sorokin and Pradeep Shenoy and Mia Chen and Mark Collier and Hongji Li and Taylor Bos and Nevan Wichers and Sun Jae Lee and Angéline Pouget and Santhosh Thangaraj and Kyriakos Axiotis and Phil Crone and Rachel Sterneck and Nikolai Chinaev and Victoria Krakovna and Oleksandr Ferludin and Ian Gemp and Stephanie Winkler and Dan Goldberg and Ivan Korotkov and Kefan Xiao and Malika Mehrotra and Sandeep Mariserla and Vihari Piratla and Terry Thurk and Khiem Pham and Hongxu Ma and Alexandre Senges and Ravi Kumar and Clemens Meyer and Ellie Talius and Nuo Wang Pierse and Ballie Sandhu and Horia Toma and Kuo Lin and Swaroop Nath and Tom Stone and Dorsa Sadigh and Nikita Gupta and Arthur Guez and Avi Singh and Matt Thomas and Tom Duerig and Yuan Gong and Richard Tanburn and Lydia Lihui Zhang and Phuong Dao and Mohamed Hammad and Sirui Xie and Shruti Rijhwani and Ben Murdoch and Duhyeon Kim and Will Thompson and Heng-Tze Cheng and Daniel Sohn and Pablo Sprechmann and Qiantong Xu and Srinivas Tadepalli and Peter Young and Ye Zhang and Hansa Srinivasan and Miranda Aperghis and Aditya Ayyar and Hen Fitoussi and Ryan Burnell and David Madras and Mike Dusenberry and Xi Xiong and Tayo Oguntebi and Ben Albrecht and Jörg Bornschein and Jovana Mitrović and Mason Dimarco and Bhargav Kanagal Shamanna and Premal Shah and Eren Sezener and Shyam Upadhyay and Dave Lacey and Craig Schiff and Sebastien Baur and Sanjay Ganapathy and Eva Schnider and Mateo Wirth and Connor Schenck and Andrey Simanovsky and Yi-Xuan Tan and Philipp Fränken and Dennis Duan and Bharath Mankalale and Nikhil Dhawan and Kevin Sequeira and Zichuan Wei and Shivanker Goel and Caglar Unlu and Yukun Zhu and Haitian Sun and Ananth Balashankar and Kurt Shuster and Megh Umekar and Mahmoud Alnahlawi and Aäron van den Oord and Kelly Chen and Yuexiang Zhai and Zihang Dai and Kuang-Huei Lee and Eric Doi and Lukas Zilka and Rohith Vallu and Disha Shrivastava and Jason Lee and Hisham Husain and Honglei Zhuang and Vincent Cohen-Addad and Jarred Barber and James Atwood and Adam Sadovsky and Quentin Wellens and Steven Hand and Arunkumar Rajendran and Aybuke Turker and CJ Carey and Yuanzhong Xu and Hagen Soltau and Zefei Li and Xinying Song and Conglong Li and Iurii Kemaev and Sasha Brown and Andrea Burns and Viorica Patraucean and Piotr Stanczyk and Renga Aravamudhan and Mathieu Blondel and Hila Noga and Lorenzo Blanco and Will Song and Michael Isard and Mandar Sharma and Reid Hayes and Dalia El Badawy and Avery Lamp and Itay Laish and Olga Kozlova and Kelvin Chan and Sahil Singla and Srinivas Sunkara and Mayank Upadhyay and Chang Liu and Aijun Bai and Jarek Wilkiewicz and Martin Zlocha and Jeremiah Liu and Zhuowan Li and Haiguang Li and Omer Barak and Ganna Raboshchuk and Jiho Choi and Fangyu Liu and Erik Jue and Mohit Sharma and Andreea Marzoca and Robert Busa-Fekete and Anna Korsun and Andre Elisseeff and Zhe Shen and Sara Mc Carthy and Kay Lamerigts and Anahita Hosseini and Hanzhao Lin and Charlie Chen and Fan Yang and Kushal Chauhan and Mark Omernick and Dawei Jia and Karina Zainullina and Demis Hassabis and Danny Vainstein and Ehsan Amid and Xiang Zhou and Ronny Votel and Eszter Vértes and Xinjian Li and Zongwei Zhou and Angeliki Lazaridou and Brendan McMahan and Arjun Narayanan and Hubert Soyer and Sujoy Basu and Kayi Lee and Bryan Perozzi and Qin Cao and Leonard Berrada and Rahul Arya and Ke Chen and Katrina and Xu and Matthias Lochbrunner and Alex Hofer and Sahand Sharifzadeh and Renjie Wu and Sally Goldman and Pranjal Awasthi and Xuezhi Wang and Yan Wu and Claire Sha and Biao Zhang and Maciej Mikuła and Filippo Graziano and Siobhan Mcloughlin and Irene Giannoumis and Youhei Namiki and Chase Malik and Carey Radebaugh and Jamie Hall and Ramiro Leal-Cavazos and Jianmin Chen and Vikas Sindhwani and David Kao and David Greene and Jordan Griffith and Chris Welty and Ceslee Montgomery and Toshihiro Yoshino and Liangzhe Yuan and Noah Goodman and Assaf Hurwitz Michaely and Kevin Lee and KP Sawhney and Wei Chen and Zheng Zheng and Megan Shum and Nikolay Savinov and Etienne Pot and Alex Pak and Morteza Zadimoghaddam and Sijal Bhatnagar and Yoad Lewenberg and Blair Kutzman and Ji Liu and Lesley Katzen and Jeremy Selier and Josip Djolonga and Dmitry Lepikhin and Kelvin Xu and Jacky Liang and Jiewen Tan and Benoit Schillings and Muge Ersoy and Pete Blois and Bernd Bandemer and Abhimanyu Singh and Sergei Lebedev and Pankaj Joshi and Adam R. Brown and Evan Palmer and Shreya Pathak and Komal Jalan and Fedir Zubach and Shuba Lall and Randall Parker and Alok Gunjan and Sergey Rogulenko and Sumit Sanghai and Zhaoqi Leng and Zoltan Egyed and Shixin Li and Maria Ivanova and Kostas Andriopoulos and Jin Xie and Elan Rosenfeld and Auriel Wright and Ankur Sharma and Xinyang Geng and Yicheng Wang and Sam Kwei and Renke Pan and Yujing Zhang and Gabby Wang and Xi Liu and Chak Yeung and Elizabeth Cole and Aviv Rosenberg and Zhen Yang and Phil Chen and George Polovets and Pranav Nair and Rohun Saxena and Josh Smith and Shuo-yiin Chang and Aroma Mahendru and Svetlana Grant and Anand Iyer and Irene Cai and Jed McGiffin and Jiaming Shen and Alanna Walton and Antonious Girgis and Oliver Woodman and Rosemary Ke and Mike Kwong and Louis Rouillard and Jinmeng Rao and Zhihao Li and Yuntao Xu and Flavien Prost and Chi Zou and Ziwei Ji and Alberto Magni and Tyler Liechty and Dan A. Calian and Deepak Ramachandran and Igor Krivokon and Hui Huang and Terry Chen and Anja Hauth and Anastasija Ilić and Weijuan Xi and Hyeontaek Lim and Vlad-Doru Ion and Pooya Moradi and Metin Toksoz-Exley and Kalesha Bullard and Miltos Allamanis and Xiaomeng Yang and Sophie Wang and Zhi Hong and Anita Gergely and Cheng Li and Bhavishya Mittal and Vitaly Kovalev and Victor Ungureanu and Jane Labanowski and Jan Wassenberg and Nicolas Lacasse and Geoffrey Cideron and Petar Dević and Annie Marsden and Lynn Nguyen and Michael Fink and Yin Zhong and Tatsuya Kiyono and Desi Ivanov and Sally Ma and Max Bain and Kiran Yalasangi and Jennifer She and Anastasia Petrushkina and Mayank Lunayach and Carla Bromberg and Sarah Hodkinson and Vilobh Meshram and Daniel Vlasic and Austin Kyker and Steve Xu and Jeff Stanway and Zuguang Yang and Kai Zhao and Matthew Tung and Seth Odoom and Yasuhisa Fujii and Justin Gilmer and Eunyoung Kim and Felix Halim and Quoc Le and Bernd Bohnet and Seliem El-Sayed and Behnam Neyshabur and Malcolm Reynolds and Dean Reich and Yang Xu and Erica Moreira and Anuj Sharma and Zeyu Liu and Mohammad Javad Hosseini and Naina Raisinghani and Yi Su and Ni Lao and Daniel Formoso and Marco Gelmi and Almog Gueta and Tapomay Dey and Elena Gribovskaya and Domagoj Ćevid and Sidharth Mudgal and Garrett Bingham and Jianling Wang and Anurag Kumar and Alex Cullum and Feng Han and Konstantinos Bousmalis and Diego Cedillo and Grace Chu and Vladimir Magay and Paul Michel and Ester Hlavnova and Daniele Calandriello and Setareh Ariafar and Kaisheng Yao and Vikash Sehwag and Arpi Vezer and Agustin Dal Lago and Zhenkai Zhu and Paul Kishan Rubenstein and Allen Porter and Anirudh Baddepudi and Oriana Riva and Mihai Dorin Istin and Chih-Kuan Yeh and Zhi Li and Andrew Howard and Nilpa Jha and Jeremy Chen and Raoul de Liedekerke and Zafarali Ahmed and Mikel Rodriguez and Tanuj Bhatia and Bangju Wang and Ali Elqursh and David Klinghoffer and Peter Chen and Pushmeet Kohli and Te I and Weiyang Zhang and Zack Nado and Jilin Chen and Maxwell Chen and George Zhang and Aayush Singh and Adam Hillier and Federico Lebron and Yiqing Tao and Ting Liu and Gabriel Dulac-Arnold and Jingwei Zhang and Shashi Narayan and Buhuang Liu and Orhan Firat and Abhishek Bhowmick and Bingyuan Liu and Hao Zhang and Zizhao Zhang and Georges Rotival and Nathan Howard and Anu Sinha and Alexander Grushetsky and Benjamin Beyret and Keerthana Gopalakrishnan and James Zhao and Kyle He and Szabolcs Payrits and Zaid Nabulsi and Zhaoyi Zhang and Weijie Chen and Edward Lee and Nova Fallen and Sreenivas Gollapudi and Aurick Zhou and Filip Pavetić and Thomas Köppe and Shiyu Huang and Rama Pasumarthi and Nick Fernando and Felix Fischer and Daria Ćurko and Yang Gao and James Svensson and Austin Stone and Haroon Qureshi and Abhishek Sinha and Apoorv Kulshreshtha and Martin Matysiak and Jieming Mao and Carl Saroufim and Aleksandra Faust and Qingnan Duan and Gil Fidel and Kaan Katircioglu and Raphaël Lopez Kaufman and Dhruv Shah and Weize Kong and Abhishek Bapna and Gellért Weisz and Emma Dunleavy and Praneet Dutta and Tianqi Liu and Rahma Chaabouni and Carolina Parada and Marcus Wu and Alexandra Belias and Alessandro Bissacco and Stanislav Fort and Li Xiao and Fantine Huot and Chris Knutsen and Yochai Blau and Gang Li and Jennifer Prendki and Juliette Love and Yinlam Chow and Pichi Charoenpanit and Hidetoshi Shimokawa and Vincent Coriou and Karol Gregor and Tomas Izo and Arjun Akula and Mario Pinto and Chris Hahn and Dominik Paulus and Jiaxian Guo and Neha Sharma and Cho-Jui Hsieh and Adaeze Chukwuka and Kazuma Hashimoto and Nathalie Rauschmayr and Ling Wu and Christof Angermueller and Yulong Wang and Sebastian Gerlach and Michael Pliskin and Daniil Mirylenka and Min Ma and Lexi Baugher and Bryan Gale and Shaan Bijwadia and Nemanja Rakićević and David Wood and Jane Park and Chung-Ching Chang and Babi Seal and Chris Tar and Kacper Krasowiak and Yiwen Song and Georgi Stephanov and Gary Wang and Marcello Maggioni and Stein Xudong Lin and Felix Wu and Shachi Paul and Zixuan Jiang and Shubham Agrawal and Bilal Piot and Alex Feng and Cheolmin Kim and Tulsee Doshi and Jonathan Lai and Chuqiao and Xu and Sharad Vikram and Ciprian Chelba and Sebastian Krause and Vincent Zhuang and Jack Rae and Timo Denk and Adrian Collister and Lotte Weerts and Xianghong Luo and Yifeng Lu and Håvard Garnes and Nitish Gupta and Terry Spitz and Avinatan Hassidim and Lihao Liang and Izhak Shafran and Peter Humphreys and Kenny Vassigh and Phil Wallis and Virat Shejwalkar and Nicolas Perez-Nieves and Rachel Hornung and Melissa Tan and Beka Westberg and Andy Ly and Richard Zhang and Brian Farris and Jongbin Park and Alec Kosik and Zeynep Cankara and Andrii Maksai and Yunhan Xu and Albin Cassirer and Sergi Caelles and Abbas Abdolmaleki and Mencher Chiang and Alex Fabrikant and Shravya Shetty and Luheng He and Mai Giménez and Hadi Hashemi and Sheena Panthaplackel and Yana Kulizhskaya and Salil Deshmukh and Daniele Pighin and Robin Alazard and Disha Jindal and Seb Noury and Pradeep Kumar S and Siyang Qin and Xerxes Dotiwalla and Stephen Spencer and Mohammad Babaeizadeh and Blake JianHang Chen and Vaibhav Mehta and Jennie Lees and Andrew Leach and Penporn Koanantakool and Ilia Akolzin and Ramona Comanescu and Junwhan Ahn and Alexey Svyatkovskiy and Basil Mustafa and David D'Ambrosio and Shiva Mohan Reddy Garlapati and Pascal Lamblin and Alekh Agarwal and Shuang Song and Pier Giuseppe Sessa and Pauline Coquinot and John Maggs and Hussain Masoom and Divya Pitta and Yaqing Wang and Patrick Morris-Suzuki and Billy Porter and Johnson Jia and Jeffrey Dudek and Raghavender R and Cosmin Paduraru and Alan Ansell and Tolga Bolukbasi and Tony Lu and Ramya Ganeshan and Zi Wang and Henry Griffiths and Rodrigo Benenson and Yifan He and James Swirhun and George Papamakarios and Aditya Chawla and Kuntal Sengupta and Yan Wang and Vedrana Milutinovic and Igor Mordatch and Zhipeng Jia and Jamie Smith and Will Ng and Shitij Nigam and Matt Young and Eugen Vušak and Blake Hechtman and Sheela Goenka and Avital Zipori and Kareem Ayoub and Ashok Popat and Trilok Acharya and Luo Yu and Dawn Bloxwich and Hugo Song and Paul Roit and Haiqiong Li and Aviel Boag and Nigamaa Nayakanti and Bilva Chandra and Tianli Ding and Aahil Mehta and Cath Hope and Jiageng Zhang and Idan Heimlich Shtacher and Kartikeya Badola and Ryo Nakashima and Andrei Sozanschi and Iulia Comşa and Ante Žužul and Emily Caveness and Julian Odell and Matthew Watson and Dario de Cesare and Phillip Lippe and Derek Lockhart and Siddharth Verma and Huizhong Chen and Sean Sun and Lin Zhuo and Aditya Shah and Prakhar Gupta and Alex Muzio and Ning Niu and Amir Zait and Abhinav Singh and Meenu Gaba and Fan Ye and Prajit Ramachandran and Mohammad Saleh and Raluca Ada Popa and Ayush Dubey and Frederick Liu and Sara Javanmardi and Mark Epstein and Ross Hemsley and Richard Green and Nishant Ranka and Eden Cohen and Chuyuan Kelly Fu and Sanjay Ghemawat and Jed Borovik and James Martens and Anthony Chen and Pranav Shyam and André Susano Pinto and Ming-Hsuan Yang and Alexandru Ţifrea and David Du and Boqing Gong and Ayushi Agarwal and Seungyeon Kim and Christian Frank and Saloni Shah and Xiaodan Song and Zhiwei Deng and Ales Mikhalap and Kleopatra Chatziprimou and Timothy Chung and Toni Creswell and Susan Zhang and Yennie Jun and Carl Lebsack and Will Truong and Slavica Andačić and Itay Yona and Marco Fornoni and Rong Rong and Serge Toropov and Afzal Shama Soudagar and Andrew Audibert and Salah Zaiem and Zaheer Abbas and Andrei Rusu and Sahitya Potluri and Shitao Weng and Anastasios Kementsietsidis and Anton Tsitsulin and Daiyi Peng and Natalie Ha and Sanil Jain and Tejasi Latkar and Simeon Ivanov and Cory McLean and Anirudh GP and Rajesh Venkataraman and Canoee Liu and Dilip Krishnan and Joel D'sa and Roey Yogev and Paul Collins and Benjamin Lee and Lewis Ho and Carl Doersch and Gal Yona and Shawn Gao and Felipe Tiengo Ferreira and Adnan Ozturel and Hannah Muckenhirn and Ce Zheng and Gargi Balasubramaniam and Mudit Bansal and George van den Driessche and Sivan Eiger and Salem Haykal and Vedant Misra and Abhimanyu Goyal and Danilo Martins and Gary Leung and Jonas Valfridsson and Four Flynn and Will Bishop and Chenxi Pang and Yoni Halpern and Honglin Yu and Lawrence Moore and Yuvein and Zhu and Sridhar Thiagarajan and Yoel Drori and Zhisheng Xiao and Lucio Dery and Rolf Jagerman and Jing Lu and Eric Ge and Vaibhav Aggarwal and Arjun Khare and Vinh Tran and Oded Elyada and Ferran Alet and James Rubin and Ian Chou and David Tian and Libin Bai and Lawrence Chan and Lukasz Lew and Karolis Misiunas and Taylan Bilal and Aniket Ray and Sindhu Raghuram and Alex Castro-Ros and Viral Carpenter and CJ Zheng and Michael Kilgore and Josef Broder and Emily Xue and Praveen Kallakuri and Dheeru Dua and Nancy Yuen and Steve Chien and John Schultz and Saurabh Agrawal and Reut Tsarfaty and Jingcao Hu and Ajay Kannan and Dror Marcus and Nisarg Kothari and Baochen Sun and Ben Horn and Matko Bošnjak and Ferjad Naeem and Dean Hirsch and Lewis Chiang and Boya Fang and Jie Han and Qifei Wang and Ben Hora and Antoine He and Mario Lučić and Beer Changpinyo and Anshuman Tripathi and John Youssef and Chester Kwak and Philippe Schlattner and Cat Graves and Rémi Leblond and Wenjun Zeng and Anders Andreassen and Gabriel Rasskin and Yue Song and Eddie Cao and Junhyuk Oh and Matt Hoffman and Wojtek Skut and Yichi Zhang and Jon Stritar and Xingyu Cai and Saarthak Khanna and Kathie Wang and Shriya Sharma and Christian Reisswig and Younghoon Jun and Aman Prasad and Tatiana Sholokhova and Preeti Singh and Adi Gerzi Rosenthal and Anian Ruoss and Françoise Beaufays and Sean Kirmani and Dongkai Chen and Johan Schalkwyk and Jonathan Herzig and Been Kim and Josh Jacob and Damien Vincent and Adrian N Reyes and Ivana Balazevic and Léonard Hussenot and Jon Schneider and Parker Barnes and Luis Castro and Spandana Raj Babbula and Simon Green and Serkan Cabi and Nico Duduta and Danny Driess and Rich Galt and Noam Velan and Junjie Wang and Hongyang Jiao and Matthew Mauger and Du Phan and Miteyan Patel and Vlado Galić and Jerry Chang and Eyal Marcus and Matt Harvey and Julian Salazar and Elahe Dabir and Suraj Satishkumar Sheth and Amol Mandhane and Hanie Sedghi and Jeremiah Willcock and Amir Zandieh and Shruthi Prabhakara and Aida Amini and Antoine Miech and Victor Stone and Massimo Nicosia and Paul Niemczyk and Ying Xiao and Lucy Kim and Sławek Kwasiborski and Vikas Verma and Ada Maksutaj Oflazer and Christoph Hirnschall and Peter Sung and Lu Liu and Richard Everett and Michiel Bakker and Ágoston Weisz and Yufei Wang and Vivek Sampathkumar and Uri Shaham and Bibo Xu and Yasemin Altun and Mingqiu Wang and Takaaki Saeki and Guanjie Chen and Emanuel Taropa and Shanthal Vasanth and Sophia Austin and Lu Huang and Goran Petrovic and Qingyun Dou and Daniel Golovin and Grigory Rozhdestvenskiy and Allie Culp and Will Wu and Motoki Sano and Divya Jain and Julia Proskurnia and Sébastien Cevey and Alejandro Cruzado Ruiz and Piyush Patil and Mahdi Mirzazadeh and Eric Ni and Javier Snaider and Lijie Fan and Alexandre Fréchette and AJ Pierigiovanni and Shariq Iqbal and Kenton Lee and Claudio Fantacci and Jinwei Xing and Lisa Wang and Alex Irpan and David Raposo and Yi Luan and Zhuoyuan Chen and Harish Ganapathy and Kevin Hui and Jiazhong Nie and Isabelle Guyon and Heming Ge and Roopali Vij and Hui Zheng and Dayeong Lee and Alfonso Castaño and Khuslen Baatarsukh and Gabriel Ibagon and Alexandra Chronopoulou and Nicholas FitzGerald and Shashank Viswanadha and Safeen Huda and Rivka Moroshko and Georgi Stoyanov and Prateek Kolhar and Alain Vaucher and Ishaan Watts and Adhi Kuncoro and Henryk Michalewski and Satish Kambala and Bat-Orgil Batsaikhan and Alek Andreev and Irina Jurenka and Maigo Le and Qihang Chen and Wael Al Jishi and Sarah Chakera and Zhe Chen and Aditya Kini and Vikas Yadav and Aditya Siddhant and Ilia Labzovsky and Balaji Lakshminarayanan and Carrie Grimes Bostock and Pankil Botadra and Ankesh Anand and Colton Bishop and Sam Conway-Rahman and Mohit Agarwal and Yani Donchev and Achintya Singhal and Félix de Chaumont Quitry and Natalia Ponomareva and Nishant Agrawal and Bin Ni and Kalpesh Krishna and Masha Samsikova and John Karro and Yilun Du and Tamara von Glehn and Caden Lu and Christopher A. Choquette-Choo and Zhen Qin and Tingnan Zhang and Sicheng Li and Divya Tyam and Swaroop Mishra and Wing Lowe and Colin Ji and Weiyi Wang and Manaal Faruqui and Ambrose Slone and Valentin Dalibard and Arunachalam Narayanaswamy and John Lambert and Pierre-Antoine Manzagol and Dan Karliner and Andrew Bolt and Ivan Lobov and Aditya Kusupati and Chang Ye and Xuan Yang and Heiga Zen and Nelson George and Mukul Bhutani and Olivier Lacombe and Robert Riachi and Gagan Bansal and Rachel Soh and Yue Gao and Yang Yu and Adams Yu and Emily Nottage and Tania Rojas-Esponda and James Noraky and Manish Gupta and Ragha Kotikalapudi and Jichuan Chang and Sanja Deur and Dan Graur and Alex Mossin and Erin Farnese and Ricardo Figueira and Alexandre Moufarek and Austin Huang and Patrik Zochbauer and Ben Ingram and Tongzhou Chen and Zelin Wu and Adrià Puigdomènech and Leland Rechis and Da Yu and Sri Gayatri Sundara Padmanabhan and Rui Zhu and Chu-ling Ko and Andrea Banino and Samira Daruki and Aarush Selvan and Dhruva Bhaswar and Daniel Hernandez Diaz and Chen Su and Salvatore Scellato and Jennifer Brennan and Woohyun Han and Grace Chung and Priyanka Agrawal and Urvashi Khandelwal and Khe Chai Sim and Morgane Lustman and Sam Ritter and Kelvin Guu and Jiawei Xia and Prateek Jain and Emma Wang and Tyrone Hill and Mirko Rossini and Marija Kostelac and Tautvydas Misiunas and Amit Sabne and Kyuyeun Kim and Ahmet Iscen and Congchao Wang and José Leal and Ashwin Sreevatsa and Utku Evci and Manfred Warmuth and Saket Joshi and Daniel Suo and James Lottes and Garrett Honke and Brendan Jou and Stefani Karp and Jieru Hu and Himanshu Sahni and Adrien Ali Taïga and William Kong and Samrat Ghosh and Renshen Wang and Jay Pavagadhi and Natalie Axelsson and Nikolai Grigorev and Patrick Siegler and Rebecca Lin and Guohui Wang and Emilio Parisotto and Sharath Maddineni and Krishan Subudhi and Eyal Ben-David and Elena Pochernina and Orgad Keller and Thi Avrahami and Zhe Yuan and Pulkit Mehta and Jialu Liu and Sherry Yang and Wendy Kan and Katherine Lee and Tom Funkhouser and Derek Cheng and Hongzhi Shi and Archit Sharma and Joe Kelley and Matan Eyal and Yury Malkov and Corentin Tallec and Yuval Bahat and Shen Yan and Xintian and Wu and David Lindner and Chengda Wu and Avi Caciularu and Xiyang Luo and Rodolphe Jenatton and Tim Zaman and Yingying Bi and Ilya Kornakov and Ganesh Mallya and Daisuke Ikeda and Itay Karo and Anima Singh and Colin Evans and Praneeth Netrapalli and Vincent Nallatamby and Isaac Tian and Yannis Assael and Vikas Raunak and Victor Carbune and Ioana Bica and Lior Madmoni and Dee Cattle and Snchit Grover and Krishna Somandepalli and Sid Lall and Amelio Vázquez-Reina and Riccardo Patana and Jiaqi Mu and Pranav Talluri and Maggie Tran and Rajeev Aggarwal and RJ Skerry-Ryan and Jun Xu and Mike Burrows and Xiaoyue Pan and Edouard Yvinec and Di Lu and Zhiying Zhang and Duc Dung Nguyen and Hairong Mu and Gabriel Barcik and Helen Ran and Lauren Beltrone and Krzysztof Choromanski and Dia Kharrat and Samuel Albanie and Sean Purser-haskell and David Bieber and Carrie Zhang and Jing Wang and Tom Hudson and Zhiyuan Zhang and Han Fu and Johannes Mauerer and Mohammad Hossein Bateni and AJ Maschinot and Bing Wang and Muye Zhu and Arjun Pillai and Tobias Weyand and Shuang Liu and Oscar Akerlund and Fred Bertsch and Vittal Premachandran and Alicia Jin and Vincent Roulet and Peter de Boursac and Shubham Mittal and Ndaba Ndebele and Georgi Karadzhov and Sahra Ghalebikesabi and Ricky Liang and Allen Wu and Yale Cong and Nimesh Ghelani and Sumeet Singh and Bahar Fatemi and Warren and Chen and Charles Kwong and Alexey Kolganov and Steve Li and Richard Song and Chenkai Kuang and Sobhan Miryoosefi and Dale Webster and James Wendt and Arkadiusz Socala and Guolong Su and Artur Mendonça and Abhinav Gupta and Xiaowei Li and Tomy Tsai and Qiong and Hu and Kai Kang and Angie Chen and Sertan Girgin and Yongqin Xian and Andrew Lee and Nolan Ramsden and Leslie Baker and Madeleine Clare Elish and Varvara Krayvanova and Rishabh Joshi and Jiri Simsa and Yao-Yuan Yang and Piotr Ambroszczyk and Dipankar Ghosh and Arjun Kar and Yuan Shangguan and Yumeya Yamamori and Yaroslav Akulov and Andy Brock and Haotian Tang and Siddharth Vashishtha and Rich Munoz and Andreas Steiner and Kalyan Andra and Daniel Eppens and Qixuan Feng and Hayato Kobayashi and Sasha Goldshtein and Mona El Mahdy and Xin Wang and Jilei and Wang and Richard Killam and Tom Kwiatkowski and Kavya Kopparapu and Serena Zhan and Chao Jia and Alexei Bendebury and Sheryl Luo and Adrià Recasens and Timothy Knight and Jing Chen and Mohak Patel and YaGuang Li and Ben Withbroe and Dean Weesner and Kush Bhatia and Jie Ren and Danielle Eisenbud and Ebrahim Songhori and Yanhua Sun and Travis Choma and Tasos Kementsietsidis and Lucas Manning and Brian Roark and Wael Farhan and Jie Feng and Susheel Tatineni and James Cobon-Kerr and Yunjie Li and Lisa Anne Hendricks and Isaac Noble and Chris Breaux and Nate Kushman and Liqian Peng and Fuzhao Xue and Taylor Tobin and Jamie Rogers and Josh Lipschultz and Chris Alberti and Alexey Vlaskin and Mostafa Dehghani and Roshan Sharma and Tris Warkentin and Chen-Yu Lee and Benigno Uria and Da-Cheng Juan and Angad Chandorkar and Hila Sheftel and Ruibo Liu and Elnaz Davoodi and Borja De Balle Pigem and Kedar Dhamdhere and David Ross and Jonathan Hoech and Mahdis Mahdieh and Li Liu and Qiujia Li and Liam McCafferty and Chenxi Liu and Markus Mircea and Yunting Song and Omkar Savant and Alaa Saade and Colin Cherry and Vincent Hellendoorn and Siddharth Goyal and Paul Pucciarelli and David Vilar Torres and Zohar Yahav and Hyo Lee and Lars Lowe Sjoesund and Christo Kirov and Bo Chang and Deepanway Ghoshal and Lu Li and Gilles Baechler and Sébastien Pereira and Tara Sainath and Anudhyan Boral and Dominik Grewe and Afief Halumi and Nguyet Minh Phu and Tianxiao Shen and Marco Tulio Ribeiro and Dhriti Varma and Alex Kaskasoli and Vlad Feinberg and Navneet Potti and Jarrod Kahn and Matheus Wisniewski and Shakir Mohamed and Arnar Mar Hrafnkelsson and Bobak Shahriari and Jean-Baptiste Lespiau and Lisa Patel and Legg Yeung and Tom Paine and Lantao Mei and Alex Ramirez and Rakesh Shivanna and Li Zhong and Josh Woodward and Guilherme Tubone and Samira Khan and Heng Chen and Elizabeth Nielsen and Catalin Ionescu and Utsav Prabhu and Mingcen Gao and Qingze Wang and Sean Augenstein and Neesha Subramaniam and Jason Chang and Fotis Iliopoulos and Jiaming Luo and Myriam Khan and Weicheng Kuo and Denis Teplyashin and Florence Perot and Logan Kilpatrick and Amir Globerson and Hongkun Yu and Anfal Siddiqui and Nick Sukhanov and Arun Kandoor and Umang Gupta and Marco Andreetto and Moran Ambar and Donnie Kim and Paweł Wesołowski and Sarah Perrin and Ben Limonchik and Wei Fan and Jim Stephan and Ian Stewart-Binks and Ryan Kappedal and Tong He and Sarah Cogan and Romina Datta and Tong Zhou and Jiayu Ye and Leandro Kieliger and Ana Ramalho and Kyle Kastner and Fabian Mentzer and Wei-Jen Ko and Arun Suggala and Tianhao Zhou and Shiraz Butt and Hana Strejček and Lior Belenki and Subhashini Venugopalan and Mingyang Ling and Evgenii Eltyshev and Yunxiao Deng and Geza Kovacs and Mukund Raghavachari and Hanjun Dai and Tal Schuster and Steven Schwarcz and Richard Nguyen and Arthur Nguyen and Gavin Buttimore and Shrestha Basu Mallick and Sudeep Gandhe and Seth Benjamin and Michal Jastrzebski and Le Yan and Sugato Basu and Chris Apps and Isabel Edkins and James Allingham and Immanuel Odisho and Tomas Kocisky and Jewel Zhao and Linting Xue and Apoorv Reddy and Chrysovalantis Anastasiou and Aviel Atias and Sam Redmond and Kieran Milan and Nicolas Heess and Herman Schmit and Allan Dafoe and Daniel Andor and Tynan Gangwani and Anca Dragan and Sheng Zhang and Ashyana Kachra and Gang Wu and Siyang Xue and Kevin Aydin and Siqi Liu and Yuxiang Zhou and Mahan Malihi and Austin Wu and Siddharth Gopal and Candice Schumann and Peter Stys and Alek Wang and Mirek Olšák and Dangyi Liu and Christian Schallhart and Yiran Mao and Demetra Brady and Hao Xu and Tomas Mery and Chawin Sitawarin and Siva Velusamy and Tom Cobley and Alex Zhai and Christian Walder and Nitzan Katz and Ganesh Jawahar and Chinmay Kulkarni and Antoine Yang and Adam Paszke and Yinan Wang and Bogdan Damoc and Zalán Borsos and Ray Smith and Jinning Li and Mansi Gupta and Andrei Kapishnikov and Sushant Prakash and Florian Luisier and Rishabh Agarwal and Will Grathwohl and Kuangyuan Chen and Kehang Han and Nikhil Mehta and Andrew Over and Shekoofeh Azizi and Lei Meng and Niccolò Dal Santo and Kelvin Zheng and Jane Shapiro and Igor Petrovski and Jeffrey Hui and Amin Ghafouri and Jasper Snoek and James Qin and Mandy Jordan and Caitlin Sikora and Jonathan Malmaud and Yuheng Kuang and Aga Świetlik and Ruoxin Sang and Chongyang Shi and Leon Li and Andrew Rosenberg and Shubin Zhao and Andy Crawford and Jan-Thorsten Peter and Yun Lei and Xavier Garcia and Long Le and Todd Wang and Julien Amelot and Dave Orr and Praneeth Kacham and Dana Alon and Gladys Tyen and Abhinav Arora and James Lyon and Alex Kurakin and Mimi Ly and Theo Guidroz and Zhipeng Yan and Rina Panigrahy and Pingmei Xu and Thais Kagohara and Yong Cheng and Eric Noland and Jinhyuk Lee and Jonathan Lee and Cathy Yip and Maria Wang and Efrat Nehoran and Alexander Bykovsky and Zhihao Shan and Ankit Bhagatwala and Chaochao Yan and Jie Tan and Guillermo Garrido and Dan Ethier and Nate Hurley and Grace Vesom and Xu Chen and Siyuan Qiao and Abhishek Nayyar and Julian Walker and Paramjit Sandhu and Mihaela Rosca and Danny Swisher and Mikhail Dektiarev and Josh Dillon and George-Cristian Muraru and Manuel Tragut and Artiom Myaskovsky and David Reid and Marko Velic and Owen Xiao and Jasmine George and Mark Brand and Jing Li and Wenhao Yu and Shane Gu and Xiang Deng and François-Xavier Aubet and Soheil Hassas Yeganeh and Fred Alcober and Celine Smith and Trevor Cohn and Kay McKinney and Michael Tschannen and Ramesh Sampath and Gowoon Cheon and Liangchen Luo and Luyang Liu and Jordi Orbay and Hui Peng and Gabriela Botea and Xiaofan Zhang and Charles Yoon and Cesar Magalhaes and Paweł Stradomski and Ian Mackinnon and Steven Hemingray and Kumaran Venkatesan and Rhys May and Jaeyoun Kim and Alex Druinsky and Jingchen Ye and Zheng Xu and Terry Huang and Jad Al Abdallah and Adil Dostmohamed and Rachana Fellinger and Tsendsuren Munkhdalai and Akanksha Maurya and Peter Garst and Yin Zhang and Maxim Krikun and Simon Bucher and Aditya Srikanth Veerubhotla and Yaxin Liu and Sheng Li and Nishesh Gupta and Jakub Adamek and Hanwen Chen and Bernett Orlando and Aleksandr Zaks and Joost van Amersfoort and Josh Camp and Hui Wan and HyunJeong Choe and Zhichun Wu and Kate Olszewska and Weiren Yu and Archita Vadali and Martin Scholz and Daniel De Freitas and Jason Lin and Amy Hua and Xin Liu and Frank Ding and Yichao Zhou and Boone Severson and Katerina Tsihlas and Samuel Yang and Tammo Spalink and Varun Yerram and Helena Pankov and Rory Blevins and Ben Vargas and Sarthak Jauhari and Matt Miecnikowski and Ming Zhang and Sandeep Kumar and Clement Farabet and Charline Le Lan and Sebastian Flennerhag and Yonatan Bitton and Ada Ma and Arthur Bražinskas and Eli Collins and Niharika Ahuja and Sneha Kudugunta and Anna Bortsova and Minh Giang and Wanzheng Zhu and Ed Chi and Scott Lundberg and Alexey Stern and Subha Puttagunta and Jing Xiong and Xiao Wu and Yash Pande and Amit Jhindal and Daniel Murphy and Jon Clark and Marc Brockschmidt and Maxine Deines and Kevin R. McKee and Dan Bahir and Jiajun Shen and Minh Truong and Daniel McDuff and Andrea Gesmundo and Edouard Rosseel and Bowen Liang and Ken Caluwaerts and Jessica Hamrick and Joseph Kready and Mary Cassin and Rishikesh Ingale and Li Lao and Scott Pollom and Yifan Ding and Wei He and Lizzetth Bellot and Joana Iljazi and Ramya Sree Boppana and Shan Han and Tara Thompson and Amr Khalifa and Anna Bulanova and Blagoj Mitrevski and Bo Pang and Emma Cooney and Tian Shi and Rey Coaguila and Tamar Yakar and Marc'aurelio Ranzato and Nikola Momchev and Chris Rawles and Zachary Charles and Young Maeng and Yuan Zhang and Rishabh Bansal and Xiaokai Zhao and Brian Albert and Yuan Yuan and Sudheendra Vijayanarasimhan and Roy Hirsch and Vinay Ramasesh and Kiran Vodrahalli and Xingyu Wang and Arushi Gupta and DJ Strouse and Jianmo Ni and Roma Patel and Gabe Taubman and Zhouyuan Huo and Dero Gharibian and Marianne Monteiro and Hoi Lam and Shobha Vasudevan and Aditi Chaudhary and Isabela Albuquerque and Kilol Gupta and Sebastian Riedel and Chaitra Hegde and Avraham Ruderman and András György and Marcus Wainwright and Ashwin Chaugule and Burcu Karagol Ayan and Tomer Levinboim and Sam Shleifer and Yogesh Kalley and Vahab Mirrokni and Abhishek Rao and Prabakar Radhakrishnan and Jay Hartford and Jialin Wu and Zhenhai Zhu and Francesco Bertolini and Hao Xiong and Nicolas Serrano and Hamish Tomlinson and Myle Ott and Yifan Chang and Mark Graham and Jian Li and Marco Liang and Xiangzhu Long and Sebastian Borgeaud and Yanif Ahmad and Alex Grills and Diana Mincu and Martin Izzard and Yuan Liu and Jinyu Xie and Louis O'Bryan and Sameera Ponda and Simon Tong and Michelle Liu and Dan Malkin and Khalid Salama and Yuankai Chen and Rohan Anil and Anand Rao and Rigel Swavely and Misha Bilenko and Nina Anderson and Tat Tan and Jing Xie and Xing Wu and Lijun Yu and Oriol Vinyals and Andrey Ryabtsev and Rumen Dangovski and Kate Baumli and Daniel Keysers and Christian Wright and Zoe Ashwood and Betty Chan and Artem Shtefan and Yaohui Guo and Ankur Bapna and Radu Soricut and Steven Pecht and Sabela Ramos and Rui Wang and Jiahao Cai and Trieu Trinh and Paul Barham and Linda Friso and Eli Stickgold and Xiangzhuo Ding and Siamak Shakeri and Diego Ardila and Eleftheria Briakou and Phil Culliton and Adam Raveret and Jingyu Cui and David Saxton and Subhrajit Roy and Javad Azizi and Pengcheng Yin and Lucia Loher and Andrew Bunner and Min Choi and Faruk Ahmed and Eric Li and Yin Li and Shengyang Dai and Michael Elabd and Sriram Ganapathy and Shivani Agrawal and Yiqing Hua and Paige Kunkle and Sujeevan Rajayogam and Arun Ahuja and Arthur Conmy and Alex Vasiloff and Parker Beak and Christopher Yew and Jayaram Mudigonda and Bartek Wydrowski and Jon Blanton and Zhengdong Wang and Yann Dauphin and Zhuo Xu and Martin Polacek and Xi Chen and Hexiang Hu and Pauline Sho and Markus Kunesch and Mehdi Hafezi Manshadi and Eliza Rutherford and Bo Li and Sissie Hsiao and Iain Barr and Alex Tudor and Matija Kecman and Arsha Nagrani and Vladimir Pchelin and Martin Sundermeyer and Aishwarya P S and Abhijit Karmarkar and Yi Gao and Grishma Chole and Olivier Bachem and Isabel Gao and Arturo BC and Matt Dibb and Mauro Verzetti and Felix Hernandez-Campos and Yana Lunts and Matthew Johnson and Julia Di Trapani and Raphael Koster and Idan Brusilovsky and Binbin Xiong and Megha Mohabey and Han Ke and Joe Zou and Tea Sabolić and Víctor Campos and John Palowitch and Alex Morris and Linhai Qiu and Pranavaraj Ponnuramu and Fangtao Li and Vivek Sharma and Kiranbir Sodhia and Kaan Tekelioglu and Aleksandr Chuklin and Madhavi Yenugula and Erika Gemzer and Theofilos Strinopoulos and Sam El-Husseini and Huiyu Wang and Yan Zhong and Edouard Leurent and Paul Natsev and Weijun Wang and Dre Mahaarachchi and Tao Zhu and Songyou Peng and Sami Alabed and Cheng-Chun Lee and Anthony Brohan and Arthur Szlam and GS Oh and Anton Kovsharov and Jenny Lee and Renee Wong and Megan Barnes and Gregory Thornton and Felix Gimeno and Omer Levy and Martin Sevenich and Melvin Johnson and Jonathan Mallinson and Robert Dadashi and Ziyue Wang and Qingchun Ren and Preethi Lahoti and Arka Dhar and Josh Feldman and Dan Zheng and Thatcher Ulrich and Liviu Panait and Michiel Blokzijl and Cip Baetu and Josip Matak and Jitendra Harlalka and Maulik Shah and Tal Marian and Daniel von Dincklage and Cosmo Du and Ruy Ley-Wild and Bethanie Brownfield and Max Schumacher and Yury Stuken and Shadi Noghabi and Sonal Gupta and Xiaoqi Ren and Eric Malmi and Felix Weissenberger and Blanca Huergo and Maria Bauza and Thomas Lampe and Arthur Douillard and Mojtaba Seyedhosseini and Roy Frostig and Zoubin Ghahramani and Kelvin Nguyen and Kashyap Krishnakumar and Chengxi Ye and Rahul Gupta and Alireza Nazari and Robert Geirhos and Pete Shaw and Ahmed Eleryan and Dima Damen and Jennimaria Palomaki and Ted Xiao and Qiyin Wu and Quan Yuan and Phoenix Meadowlark and Matthew Bilotti and Raymond Lin and Mukund Sridhar and Yannick Schroecker and Da-Woon Chung and Jincheng Luo and Trevor Strohman and Tianlin Liu and Anne Zheng and Jesse Emond and Wei Wang and Andrew Lampinen and Toshiyuki Fukuzawa and Folawiyo Campbell-Ajala and Monica Roy and James Lee-Thorp and Lily Wang and Iftekhar Naim and Tony and Nguy\~ên and Guy Bensky and Aditya Gupta and Dominika Rogozińska and Justin Fu and Thanumalayan Sankaranarayana Pillai and Petar Veličković and Shahar Drath and Philipp Neubeck and Vaibhav Tulsyan and Arseniy Klimovskiy and Don Metzler and Sage Stevens and Angel Yeh and Junwei Yuan and Tianhe Yu and Kelvin Zhang and Alec Go and Vincent Tsang and Ying Xu and Andy Wan and Isaac Galatzer-Levy and Sam Sobell and Abodunrinwa Toki and Elizabeth Salesky and Wenlei Zhou and Diego Antognini and Sholto Douglas and Shimu Wu and Adam Lelkes and Frank Kim and Paul Cavallaro and Ana Salazar and Yuchi Liu and James Besley and Tiziana Refice and Yiling Jia and Zhang Li and Michal Sokolik and Arvind Kannan and Jon Simon and Jo Chick and Avia Aharon and Meet Gandhi and Mayank Daswani and Keyvan Amiri and Vighnesh Birodkar and Abe Ittycheriah and Peter Grabowski and Oscar Chang and Charles Sutton and Zhixin and Lai and Umesh Telang and Susie Sargsyan and Tao Jiang and Raphael Hoffmann and Nicole Brichtova and Matteo Hessel and Jonathan Halcrow and Sammy Jerome and Geoff Brown and Alex Tomala and Elena Buchatskaya and Dian Yu and Sachit Menon and Pol Moreno and Yuguo Liao and Vicky Zayats and Luming Tang and SQ Mah and Ashish Shenoy and Alex Siegman and Majid Hadian and Okwan Kwon and Tao Tu and Nima Khajehnouri and Ryan Foley and Parisa Haghani and Zhongru Wu and Vaishakh Keshava and Khyatti Gupta and Tony Bruguier and Rui Yao and Danny Karmon and Luisa Zintgraf and Zhicheng Wang and Enrique Piqueras and Junehyuk Jung and Jenny Brennan and Diego Machado and Marissa Giustina and MH Tessler and Kamyu Lee and Qiao Zhang and Joss Moore and Kaspar Daugaard and Alexander Frömmgen and Jennifer Beattie and Fred Zhang and Daniel Kasenberg and Ty Geri and Danfeng Qin and Gaurav Singh Tomar and Tom Ouyang and Tianli Yu and Luowei Zhou and Rajiv Mathews and Andy Davis and Yaoyiran Li and Jai Gupta and Damion Yates and Linda Deng and Elizabeth Kemp and Ga-Young Joung and Sergei Vassilvitskii and Mandy Guo and Pallavi LV and Dave Dopson and Sami Lachgar and Lara McConnaughey and Himadri Choudhury and Dragos Dena and Aaron Cohen and Joshua Ainslie and Sergey Levi and Parthasarathy Gopavarapu and Polina Zablotskaia and Hugo Vallet and Sanaz Bahargam and Xiaodan Tang and Nenad Tomasev and Ethan Dyer and Daniel Balle and Hongrae Lee and William Bono and Jorge Gonzalez Mendez and Vadim Zubov and Shentao Yang and Ivor Rendulic and Yanyan Zheng and Andrew Hogue and Golan Pundak and Ralph Leith and Avishkar Bhoopchand and Michael Han and Mislav Žanić and Tom Schaul and Manolis Delakis and Tejas Iyer and Guanyu Wang and Harman Singh and Abdelrahman Abdelhamed and Tara Thomas and Siddhartha Brahma and Hilal Dib and Naveen Kumar and Wenxuan Zhou and Liang Bai and Pushkar Mishra and Jiao Sun and Valentin Anklin and Roykrong Sukkerd and Lauren Agubuzu and Anton Briukhov and Anmol Gulati and Maximilian Sieb and Fabio Pardo and Sara Nasso and Junquan Chen and Kexin Zhu and Tiberiu Sosea and Alex Goldin and Keith Rush and Spurthi Amba Hombaiah and Andreas Noever and Allan Zhou and Sam Haves and Mary Phuong and Jake Ades and Yi-ting Chen and Lin Yang and Joseph Pagadora and Stan Bileschi and Victor Cotruta and Rachel Saputro and Arijit Pramanik and Sean Ammirati and Dan Garrette and Kevin Villela and Tim Blyth and Canfer Akbulut and Neha Jha and Alban Rrustemi and Arissa Wongpanich and Chirag Nagpal and Yonghui Wu and Morgane Rivière and Sergey Kishchenko and Pranesh Srinivasan and Alice Chen and Animesh Sinha and Trang Pham and Bill Jia and Tom Hennigan and Anton Bakalov and Nithya Attaluri and Drew Garmon and Daniel Rodriguez and Dawid Wegner and Wenhao Jia and Evan Senter and Noah Fiedel and Denis Petek and Yuchuan Liu and Cassidy Hardin and Harshal Tushar Lehri and Joao Carreira and Sara Smoot and Marcel Prasetya and Nami Akazawa and Anca Stefanoiu and Chia-Hua Ho and Anelia Angelova and Kate Lin and Min Kim and Charles Chen and Marcin Sieniek and Alice Li and Tongfei Guo and Sorin Baltateanu and Pouya Tafti and Michael Wunder and Nadav Olmert and Divyansh Shukla and Jingwei Shen and Neel Kovelamudi and Balaji Venkatraman and Seth Neel and Romal Thoppilan and Jerome Connor and Frederik Benzing and Axel Stjerngren and Golnaz Ghiasi and Alex Polozov and Joshua Howland and Theophane Weber and Justin Chiu and Ganesh Poomal Girirajan and Andreas Terzis and Pidong Wang and Fangda Li and Yoav Ben Shalom and Dinesh Tewari and Matthew Denton and Roee Aharoni and Norbert Kalb and Heri Zhao and Junlin Zhang and Angelos Filos and Matthew Rahtz and Lalit Jain and Connie Fan and Vitor Rodrigues and Ruth Wang and Richard Shin and Jacob Austin and Roman Ring and Mariella Sanchez-Vargas and Mehadi Hassen and Ido Kessler and Uri Alon and Gufeng Zhang and Wenhu Chen and Yenai Ma and Xiance Si and Le Hou and Azalia Mirhoseini and Marc Wilson and Geoff Bacon and Becca Roelofs and Lei Shu and Gautam Vasudevan and Jonas Adler and Artur Dwornik and Tayfun Terzi and Matt Lawlor and Harry Askham and Mike Bernico and Xuanyi Dong and Chris Hidey and Kevin Kilgour and Gaël Liu and Surya Bhupatiraju and Luke Leonhard and Siqi Zuo and Partha Talukdar and Qing Wei and Aliaksei Severyn and Vít Listík and Jong Lee and Aditya Tripathi and SK Park and Yossi Matias and Hao Liu and Alex Ruiz and Rajesh Jayaram and Jackson Tolins and Pierre Marcenac and Yiming Wang and Bryan Seybold and Henry Prior and Deepak Sharma and Jack Weber and Mikhail Sirotenko and Yunhsuan Sung and Dayou Du and Ellie Pavlick and Stefan Zinke and Markus Freitag and Max Dylla and Montse Gonzalez Arenas and Natan Potikha and Omer Goldman and Connie Tao and Rachita Chhaparia and Maria Voitovich and Pawan Dogra and Andrija Ražnatović and Zak Tsai and Chong You and Oleaser Johnson and George Tucker and Chenjie Gu and Jae Yoo and Maryam Majzoubi and Valentin Gabeur and Bahram Raad and Rocky Rhodes and Kashyap Kolipaka and Heidi Howard and Geta Sampemane and Benny Li and Chulayuth Asawaroengchai and Duy Nguyen and Chiyuan Zhang and Timothee Cour and Xinxin Yu and Zhao Fu and Joe Jiang and Po-Sen Huang and Gabriela Surita and Iñaki Iturrate and Yael Karov and Michael Collins and Martin Baeuml and Fabian Fuchs and Shilpa Shetty and Swaroop Ramaswamy and Sayna Ebrahimi and Qiuchen Guo and Jeremy Shar and Gabe Barth-Maron and Sravanti Addepalli and Bryan Richter and Chin-Yi Cheng and Eugénie Rives and Fei Zheng and Johannes Griesser and Nishanth Dikkala and Yoel Zeldes and Ilkin Safarli and Dipanjan Das and Himanshu Srivastava and Sadh MNM Khan and Xin Li and Aditya Pandey and Larisa Markeeva and Dan Belov and Qiqi Yan and Mikołaj Rybiński and Tao Chen and Megha Nawhal and Michael Quinn and Vineetha Govindaraj and Sarah York and Reed Roberts and Roopal Garg and Namrata Godbole and Jake Abernethy and Anil Das and Lam Nguyen Thiet and Jonathan Tompson and John Nham and Neera Vats and Ben Caine and Wesley Helmholz and Francesco Pongetti and Yeongil Ko and James An and Clara Huiyi Hu and Yu-Cheng Ling and Julia Pawar and Robert Leland and Keisuke Kinoshita and Waleed Khawaja and Marco Selvi and Eugene Ie and Danila Sinopalnikov and Lev Proleev and Nilesh Tripuraneni and Michele Bevilacqua and Seungji Lee and Clayton Sanford and Dan Suh and Dustin Tran and Jeff Dean and Simon Baumgartner and Jens Heitkaemper and Sagar Gubbi and Kristina Toutanova and Yichong Xu and Chandu Thekkath and Keran Rong and Palak Jain and Annie Xie and Yan Virin and Yang Li and Lubo Litchev and Richard Powell and Tarun Bharti and Adam Kraft and Nan Hua and Marissa Ikonomidis and Ayal Hitron and Sanjiv Kumar and Loic Matthey and Sophie Bridgers and Lauren Lax and Ishaan Malhi and Ondrej Skopek and Ashish Gupta and Jiawei Cao and Mitchelle Rasquinha and Siim Põder and Wojciech Stokowiec and Nicholas Roth and Guowang Li and Michaël Sander and Joshua Kessinger and Vihan Jain and Edward Loper and Wonpyo Park and Michal Yarom and Liqun Cheng and Guru Guruganesh and Kanishka Rao and Yan Li and Catarina Barros and Mikhail Sushkov and Chun-Sung Ferng and Rohin Shah and Ophir Aharoni and Ravin Kumar and Tim McConnell and Peiran Li and Chen Wang and Fernando Pereira and Craig Swanson and Fayaz Jamil and Yan Xiong and Anitha Vijayakumar and Prakash Shroff and Kedar Soparkar and Jindong Gu and Livio Baldini Soares and Eric Wang and Kushal Majmundar and Aurora Wei and Kai Bailey and Nora Kassner and Chizu Kawamoto and Goran Žužić and Victor Gomes and Abhirut Gupta and Michael Guzman and Ishita Dasgupta and Xinyi Bai and Zhufeng Pan and Francesco Piccinno and Hadas Natalie Vogel and Octavio Ponce and Adrian Hutter and Paul Chang and Pan-Pan Jiang and Ionel Gog and Vlad Ionescu and James Manyika and Fabian Pedregosa and Harry Ragan and Zach Behrman and Ryan Mullins and Coline Devin and Aroonalok Pyne and Swapnil Gawde and Martin Chadwick and Yiming Gu and Sasan Tavakkol and Andy Twigg and Naman Goyal and Ndidi Elue and Anna Goldie and Srinivasan Venkatachary and Hongliang Fei and Ziqiang Feng and Marvin Ritter and Isabel Leal and Sudeep Dasari and Pei Sun and Alif Raditya Rochman and Brendan O'Donoghue and Yuchen Liu and Jim Sproch and Kai Chen and Natalie Clay and Slav Petrov and Sailesh Sidhwani and Ioana Mihailescu and Alex Panagopoulos and AJ Piergiovanni and Yunfei Bai and George Powell and Deep Karkhanis and Trevor Yacovone and Petr Mitrichev and Joe Kovac and Dave Uthus and Amir Yazdanbakhsh and David Amos and Steven Zheng and Bing Zhang and Jin Miao and Bhuvana Ramabhadran and Soroush Radpour and Shantanu Thakoor and Josh Newlan and Oran Lang and Orion Jankowski and Shikhar Bharadwaj and Jean-Michel Sarr and Shereen Ashraf and Sneha Mondal and Jun Yan and Ankit Singh Rawat and Sarmishta Velury and Greg Kochanski and Tom Eccles and Franz Och and Abhanshu Sharma and Ethan Mahintorabi and Alex Gurney and Carrie Muir and Vered Cohen and Saksham Thakur and Adam Bloniarz and Asier Mujika and Alexander Pritzel and Paul Caron and Altaf Rahman and Fiona Lang and Yasumasa Onoe and Petar Sirkovic and Jay Hoover and Ying Jian and Pablo Duque and Arun Narayanan and David Soergel and Alex Haig and Loren Maggiore and Shyamal Buch and Josef Dean and Ilya Figotin and Igor Karpov and Shaleen Gupta and Denny Zhou and Muhuan Huang and Ashwin Vaswani and Christopher Semturs and Kaushik Shivakumar and Yu Watanabe and Vinodh Kumar Rajendran and Eva Lu and Yanhan Hou and Wenting Ye and Shikhar Vashishth and Nana Nti and Vytenis Sakenas and Darren Ni and Doug DeCarlo and Michael Bendersky and Sumit Bagri and Nacho Cano and Elijah Peake and Simon Tokumine and Varun Godbole and Carlos Guía and Tanya Lando and Vittorio Selo and Seher Ellis and Danny Tarlow and Daniel Gillick and Alessandro Epasto and Siddhartha Reddy Jonnalagadda and Meng Wei and Meiyan Xie and Ankur Taly and Michela Paganini and Mukund Sundararajan and Daniel Toyama and Ting Yu and Dessie Petrova and Aneesh Pappu and Rohan Agrawal and Senaka Buthpitiya and Justin Frye and Thomas Buschmann and Remi Crocker and Marco Tagliasacchi and Mengchao Wang and Da Huang and Sagi Perel and Brian Wieder and Hideto Kazawa and Weiyue Wang and Jeremy Cole and Himanshu Gupta and Ben Golan and Seojin Bang and Nitish Kulkarni and Ken Franko and Casper Liu and Doug Reid and Sid Dalmia and Jay Whang and Kevin Cen and Prasha Sundaram and Johan Ferret and Berivan Isik and Lucian Ionita and Guan Sun and Anna Shekhawat and Muqthar Mohammad and Philip Pham and Ronny Huang and Karthik Raman and Xingyi Zhou and Ross Mcilroy and Austin Myers and Sheng Peng and Jacob Scott and Paul Covington and Sofia Erell and Pratik Joshi and João Gabriel Oliveira and Natasha Noy and Tajwar Nasir and Jake Walker and Vera Axelrod and Tim Dozat and Pu Han and Chun-Te Chu and Eugene Weinstein and Anand Shukla and Shreyas Chandrakaladharan and Petra Poklukar and Bonnie Li and Ye Jin and Prem Eruvbetine and Steven Hansen and Avigail Dabush and Alon Jacovi and Samrat Phatale and Chen Zhu and Steven Baker and Mo Shomrat and Yang Xiao and Jean Pouget-Abadie and Mingyang Zhang and Fanny Wei and Yang Song and Helen King and Yiling Huang and Yun Zhu and Ruoxi Sun and Juliana Vicente Franco and Chu-Cheng Lin and Sho Arora and Hui and Li and Vivian Xia and Luke Vilnis and Mariano Schain and Kaiz Alarakyia and Laurel Prince and Aaron Phillips and Caleb Habtegebriel and Luyao Xu and Huan Gui and Santiago Ontanon and Lora Aroyo and Karan Gill and Peggy Lu and Yash Katariya and Dhruv Madeka and Shankar Krishnan and Shubha Srinivas Raghvendra and James Freedman and Yi Tay and Gaurav Menghani and Peter Choy and Nishita Shetty and Dan Abolafia and Doron Kukliansky and Edward Chou and Jared Lichtarge and Ken Burke and Ben Coleman and Dee Guo and Larry Jin and Indro Bhattacharya and Victoria Langston and Yiming Li and Suyog Kotecha and Alex Yakubovich and Xinyun Chen and Petre Petrov and Tolly Powell and Yanzhang He and Corbin Quick and Kanav Garg and Dawsen Hwang and Yang Lu and Srinadh Bhojanapalli and Kristian Kjems and Ramin Mehran and Aaron Archer and Hado van Hasselt and Ashwin Balakrishna and JK Kearns and Meiqi Guo and Jason Riesa and Mikita Sazanovich and Xu Gao and Chris Sauer and Chengrun Yang and XiangHai Sheng and Thomas Jimma and Wouter Van Gansbeke and Vitaly Nikolaev and Wei Wei and Katie Millican and Ruizhe Zhao and Justin Snyder and Levent Bolelli and Maura O'Brien and Shawn Xu and Fei Xia and Wentao Yuan and Arvind Neelakantan and David Barker and Sachin Yadav and Hannah Kirkwood and Farooq Ahmad and Joel Wee and Jordan Grimstad and Boyu Wang and Matthew Wiethoff and Shane Settle and Miaosen Wang and Charles Blundell and Jingjing Chen and Chris Duvarney and Grace Hu and Olaf Ronneberger and Alex Lee and Yuanzhen Li and Abhishek Chakladar and Alena Butryna and Georgios Evangelopoulos and Guillaume Desjardins and Jonni Kanerva and Henry Wang and Averi Nowak and Nick Li and Alyssa Loo and Art Khurshudov and Laurent El Shafey and Nagabhushan Baddi and Karel Lenc and Yasaman Razeghi and Tom Lieber and Amer Sinha and Xiao Ma and Yao Su and James Huang and Asahi Ushio and Hanna Klimczak-Plucińska and Kareem Mohamed and JD Chen and Simon Osindero and Stav Ginzburg and Lampros Lamprou and Vasilisa Bashlovkina and Duc-Hieu Tran and Ali Khodaei and Ankit Anand and Yixian Di and Ramy Eskander and Manish Reddy Vuyyuru and Jasmine Liu and Aishwarya Kamath and Roman Goldenberg and Mathias Bellaiche and Juliette Pluto and Bill Rosgen and Hassan Mansoor and William Wong and Suhas Ganesh and Eric Bailey and Scott Baird and Dan Deutsch and Jinoo Baek and Xuhui Jia and Chansoo Lee and Abe Friesen and Nathaniel Braun and Kate Lee and Amayika Panda and Steven M. Hernandez and Duncan Williams and Jianqiao Liu and Ethan Liang and Arnaud Autef and Emily Pitler and Deepali Jain and Phoebe Kirk and Oskar Bunyan and Jaume Sanchez Elias and Tongxin Yin and Machel Reid and Aedan Pope and Nikita Putikhin and Bidisha Samanta and Sergio Guadarrama and Dahun Kim and Simon Rowe and Marcella Valentine and Geng Yan and Alex Salcianu and David Silver and Gan Song and Richa Singh and Shuai Ye and Hannah DeBalsi and Majd Al Merey and Eran Ofek and Albert Webson and Shibl Mourad and Ashwin Kakarla and Silvio Lattanzi and Nick Roy and Evgeny Sluzhaev and Christina Butterfield and Alessio Tonioni and Nathan Waters and Sudhindra Kopalle and Jason Chase and James Cohan and Girish Ramchandra Rao and Robert Berry and Michael Voznesensky and Shuguang Hu and Kristen Chiafullo and Sharat Chikkerur and George Scrivener and Ivy Zheng and Jeremy Wiesner and Wolfgang Macherey and Timothy Lillicrap and Fei Liu and Brian Walker and David Welling and Elinor Davies and Yangsibo Huang and Lijie Ren and Nir Shabat and Alessandro Agostini and Mariko Iinuma and Dustin Zelle and Rohit Sathyanarayana and Andrea D'olimpio and Morgan Redshaw and Matt Ginsberg and Ashwin Murthy and Mark Geller and Tatiana Matejovicova and Ayan Chakrabarti and Ryan Julian and Christine Chan and Qiong Hu and Daniel Jarrett and Manu Agarwal and Jeshwanth Challagundla and Tao Li and Sandeep Tata and Wen Ding and Maya Meng and Zhuyun Dai and Giulia Vezzani and Shefali Garg and Jannis Bulian and Mary Jasarevic and Honglong Cai and Harish Rajamani and Adam Santoro and Florian Hartmann and Chen Liang and Bartek Perz and Apoorv Jindal and Fan Bu and Sungyong Seo and Ryan Poplin and Adrian Goedeckemeyer and Badih Ghazi and Nikhil Khadke and Leon Liu and Kevin Mather and Mingda Zhang and Ali Shah and Alex Chen and Jinliang Wei and Keshav Shivam and Yuan Cao and Donghyun Cho and Angelo Scorza Scarpati and Michael Moffitt and Clara Barbu and Ivan Jurin and Ming-Wei Chang and Hongbin Liu and Hao Zheng and Shachi Dave and Christine Kaeser-Chen and Xiaobin Yu and Alvin Abdagic and Lucas Gonzalez and Yanping Huang and Peilin Zhong and Cordelia Schmid and Bryce Petrini and Alex Wertheim and Jifan Zhu and Hoang Nguyen and Kaiyang Ji and Yanqi Zhou and Tao Zhou and Fangxiaoyu Feng and Regev Cohen and David Rim and Shubham Milind Phal and Petko Georgiev and Ariel Brand and Yue Ma and Wei Li and Somit Gupta and Chao Wang and Pavel Dubov and Jean Tarbouriech and Kingshuk Majumder and Huijian Li and Norman Rink and Apurv Suman and Yang Guo and Yinghao Sun and Arun Nair and Xiaowei Xu and Mohamed Elhawaty and Rodrigo Cabrera and Guangxing Han and Julian Eisenschlos and Junwen Bai and Yuqi Li and Yamini Bansal and Thibault Sellam and Mina Khan and Hung Nguyen and Justin Mao-Jones and Nikos Parotsidis and Jake Marcus and Cindy Fan and Roland Zimmermann and Yony Kochinski and Laura Graesser and Feryal Behbahani and Alvaro Caceres and Michael Riley and Patrick Kane and Sandra Lefdal and Rob Willoughby and Paul Vicol and Lun Wang and Shujian Zhang and Ashleah Gill and Yu Liang and Gautam Prasad and Soroosh Mariooryad and Mehran Kazemi and Zifeng Wang and Kritika Muralidharan and Paul Voigtlaender and Jeffrey Zhao and Huanjie Zhou and Nina D'Souza and Aditi Mavalankar and Séb Arnold and Nick Young and Obaid Sarvana and Chace Lee and Milad Nasr and Tingting Zou and Seokhwan Kim and Lukas Haas and Kaushal Patel and Neslihan Bulut and David Parkinson and Courtney Biles and Dmitry Kalashnikov and Chi Ming To and Aviral Kumar and Jessica Austin and Alex Greve and Lei Zhang and Megha Goel and Yeqing Li and Sergey Yaroshenko and Max Chang and Abhishek Jindal and Geoff Clark and Hagai Taitelbaum and Dale Johnson and Ofir Roval and Jeongwoo Ko and Anhad Mohananey and Christian Schuler and Shenil Dodhia and Ruichao Li and Kazuki Osawa and Claire Cui and Peng Xu and Rushin Shah and Tao Huang and Ela Gruzewska and Nathan Clement and Mudit Verma and Olcan Sercinoglu and Hai Qian and Viral Shah and Masa Yamaguchi and Abhinit Modi and Takahiro Kosakai and Thomas Strohmann and Junhao Zeng and Beliz Gunel and Jun Qian and Austin Tarango and Krzysztof Jastrzębski and Robert David and Jyn Shan and Parker Schuh and Kunal Lad and Willi Gierke and Mukundan Madhavan and Xinyi Chen and Mark Kurzeja and Rebeca Santamaria-Fernandez and Dawn Chen and Alexandra Cordell and Yuri Chervonyi and Frankie Garcia and Nithish Kannen and Vincent Perot and Nan Ding and Shlomi Cohen-Ganor and Victor Lavrenko and Junru Wu and Georgie Evans and Cicero Nogueira dos Santos and Madhavi Sewak and Ashley Brown and Andrew Hard and Joan Puigcerver and Zeyu Zheng and Yizhong Liang and Evgeny Gladchenko and Reeve Ingle and Uri First and Pierre Sermanet and Charlotte Magister and Mihajlo Velimirović and Sashank Reddi and Susanna Ricco and Eirikur Agustsson and Hartwig Adam and Nir Levine and David Gaddy and Dan Holtmann-Rice and Xuanhui Wang and Ashutosh Sathe and Abhijit Guha Roy and Blaž Bratanič and Alen Carin and Harsh Mehta and Silvano Bonacina and Nicola De Cao and Mara Finkelstein and Verena Rieser and Xinyi Wu and Florent Altché and Dylan Scandinaro and Li Li and Nino Vieillard and Nikhil Sethi and Garrett Tanzer and Zhi Xing and Shibo Wang and Parul Bhatia and Gui Citovsky and Thomas Anthony and Sharon Lin and Tianze Shi and Shoshana Jakobovits and Gena Gibson and Raj Apte and Lisa Lee and Mingqing Chen and Arunkumar Byravan and Petros Maniatis and Kellie Webster and Andrew Dai and Pu-Chin Chen and Jiaqi Pan and Asya Fadeeva and Zach Gleicher and Thang Luong and Niket Kumar Bhumihar},
      year={2025},
      eprint={2507.06261},
      archivePrefix={arXiv},
      primaryClass={cs.CL},
      url={https://arxiv.org/abs/2507.06261}, 
}

@misc{openai2025gpt5,
  author       = {{OpenAI}},
  title        = {Introducing GPT-5},
  howpublished = {\url{https://openai.com/index/introducing-gpt-5/}},
  year         = {2025},
  month        = {aug},
}

@misc{openai2024gpt4ocard,
      title={GPT-4o System Card}, 
      author={OpenAI and : and Aaron Hurst and Adam Lerer and Adam P. Goucher and Adam Perelman and Aditya Ramesh and Aidan Clark and AJ Ostrow and Akila Welihinda and Alan Hayes and Alec Radford and Aleksander Mądry and Alex Baker-Whitcomb and Alex Beutel and Alex Borzunov and Alex Carney and Alex Chow and Alex Kirillov and Alex Nichol and Alex Paino and Alex Renzin and Alex Tachard Passos and Alexander Kirillov and Alexi Christakis and Alexis Conneau and Ali Kamali and Allan Jabri and Allison Moyer and Allison Tam and Amadou Crookes and Amin Tootoochian and Amin Tootoonchian and Ananya Kumar and Andrea Vallone and Andrej Karpathy and Andrew Braunstein and Andrew Cann and Andrew Codispoti and Andrew Galu and Andrew Kondrich and Andrew Tulloch and Andrey Mishchenko and Angela Baek and Angela Jiang and Antoine Pelisse and Antonia Woodford and Anuj Gosalia and Arka Dhar and Ashley Pantuliano and Avi Nayak and Avital Oliver and Barret Zoph and Behrooz Ghorbani and Ben Leimberger and Ben Rossen and Ben Sokolowsky and Ben Wang and Benjamin Zweig and Beth Hoover and Blake Samic and Bob McGrew and Bobby Spero and Bogo Giertler and Bowen Cheng and Brad Lightcap and Brandon Walkin and Brendan Quinn and Brian Guarraci and Brian Hsu and Bright Kellogg and Brydon Eastman and Camillo Lugaresi and Carroll Wainwright and Cary Bassin and Cary Hudson and Casey Chu and Chad Nelson and Chak Li and Chan Jun Shern and Channing Conger and Charlotte Barette and Chelsea Voss and Chen Ding and Cheng Lu and Chong Zhang and Chris Beaumont and Chris Hallacy and Chris Koch and Christian Gibson and Christina Kim and Christine Choi and Christine McLeavey and Christopher Hesse and Claudia Fischer and Clemens Winter and Coley Czarnecki and Colin Jarvis and Colin Wei and Constantin Koumouzelis and Dane Sherburn and Daniel Kappler and Daniel Levin and Daniel Levy and David Carr and David Farhi and David Mely and David Robinson and David Sasaki and Denny Jin and Dev Valladares and Dimitris Tsipras and Doug Li and Duc Phong Nguyen and Duncan Findlay and Edede Oiwoh and Edmund Wong and Ehsan Asdar and Elizabeth Proehl and Elizabeth Yang and Eric Antonow and Eric Kramer and Eric Peterson and Eric Sigler and Eric Wallace and Eugene Brevdo and Evan Mays and Farzad Khorasani and Felipe Petroski Such and Filippo Raso and Francis Zhang and Fred von Lohmann and Freddie Sulit and Gabriel Goh and Gene Oden and Geoff Salmon and Giulio Starace and Greg Brockman and Hadi Salman and Haiming Bao and Haitang Hu and Hannah Wong and Haoyu Wang and Heather Schmidt and Heather Whitney and Heewoo Jun and Hendrik Kirchner and Henrique Ponde de Oliveira Pinto and Hongyu Ren and Huiwen Chang and Hyung Won Chung and Ian Kivlichan and Ian O'Connell and Ian O'Connell and Ian Osband and Ian Silber and Ian Sohl and Ibrahim Okuyucu and Ikai Lan and Ilya Kostrikov and Ilya Sutskever and Ingmar Kanitscheider and Ishaan Gulrajani and Jacob Coxon and Jacob Menick and Jakub Pachocki and James Aung and James Betker and James Crooks and James Lennon and Jamie Kiros and Jan Leike and Jane Park and Jason Kwon and Jason Phang and Jason Teplitz and Jason Wei and Jason Wolfe and Jay Chen and Jeff Harris and Jenia Varavva and Jessica Gan Lee and Jessica Shieh and Ji Lin and Jiahui Yu and Jiayi Weng and Jie Tang and Jieqi Yu and Joanne Jang and Joaquin Quinonero Candela and Joe Beutler and Joe Landers and Joel Parish and Johannes Heidecke and John Schulman and Jonathan Lachman and Jonathan McKay and Jonathan Uesato and Jonathan Ward and Jong Wook Kim and Joost Huizinga and Jordan Sitkin and Jos Kraaijeveld and Josh Gross and Josh Kaplan and Josh Snyder and Joshua Achiam and Joy Jiao and Joyce Lee and Juntang Zhuang and Justyn Harriman and Kai Fricke and Kai Hayashi and Karan Singhal and Katy Shi and Kavin Karthik and Kayla Wood and Kendra Rimbach and Kenny Hsu and Kenny Nguyen and Keren Gu-Lemberg and Kevin Button and Kevin Liu and Kiel Howe and Krithika Muthukumar and Kyle Luther and Lama Ahmad and Larry Kai and Lauren Itow and Lauren Workman and Leher Pathak and Leo Chen and Li Jing and Lia Guy and Liam Fedus and Liang Zhou and Lien Mamitsuka and Lilian Weng and Lindsay McCallum and Lindsey Held and Long Ouyang and Louis Feuvrier and Lu Zhang and Lukas Kondraciuk and Lukasz Kaiser and Luke Hewitt and Luke Metz and Lyric Doshi and Mada Aflak and Maddie Simens and Madelaine Boyd and Madeleine Thompson and Marat Dukhan and Mark Chen and Mark Gray and Mark Hudnall and Marvin Zhang and Marwan Aljubeh and Mateusz Litwin and Matthew Zeng and Max Johnson and Maya Shetty and Mayank Gupta and Meghan Shah and Mehmet Yatbaz and Meng Jia Yang and Mengchao Zhong and Mia Glaese and Mianna Chen and Michael Janner and Michael Lampe and Michael Petrov and Michael Wu and Michele Wang and Michelle Fradin and Michelle Pokrass and Miguel Castro and Miguel Oom Temudo de Castro and Mikhail Pavlov and Miles Brundage and Miles Wang and Minal Khan and Mira Murati and Mo Bavarian and Molly Lin and Murat Yesildal and Nacho Soto and Natalia Gimelshein and Natalie Cone and Natalie Staudacher and Natalie Summers and Natan LaFontaine and Neil Chowdhury and Nick Ryder and Nick Stathas and Nick Turley and Nik Tezak and Niko Felix and Nithanth Kudige and Nitish Keskar and Noah Deutsch and Noel Bundick and Nora Puckett and Ofir Nachum and Ola Okelola and Oleg Boiko and Oleg Murk and Oliver Jaffe and Olivia Watkins and Olivier Godement and Owen Campbell-Moore and Patrick Chao and Paul McMillan and Pavel Belov and Peng Su and Peter Bak and Peter Bakkum and Peter Deng and Peter Dolan and Peter Hoeschele and Peter Welinder and Phil Tillet and Philip Pronin and Philippe Tillet and Prafulla Dhariwal and Qiming Yuan and Rachel Dias and Rachel Lim and Rahul Arora and Rajan Troll and Randall Lin and Rapha Gontijo Lopes and Raul Puri and Reah Miyara and Reimar Leike and Renaud Gaubert and Reza Zamani and Ricky Wang and Rob Donnelly and Rob Honsby and Rocky Smith and Rohan Sahai and Rohit Ramchandani and Romain Huet and Rory Carmichael and Rowan Zellers and Roy Chen and Ruby Chen and Ruslan Nigmatullin and Ryan Cheu and Saachi Jain and Sam Altman and Sam Schoenholz and Sam Toizer and Samuel Miserendino and Sandhini Agarwal and Sara Culver and Scott Ethersmith and Scott Gray and Sean Grove and Sean Metzger and Shamez Hermani and Shantanu Jain and Shengjia Zhao and Sherwin Wu and Shino Jomoto and Shirong Wu and Shuaiqi and Xia and Sonia Phene and Spencer Papay and Srinivas Narayanan and Steve Coffey and Steve Lee and Stewart Hall and Suchir Balaji and Tal Broda and Tal Stramer and Tao Xu and Tarun Gogineni and Taya Christianson and Ted Sanders and Tejal Patwardhan and Thomas Cunninghman and Thomas Degry and Thomas Dimson and Thomas Raoux and Thomas Shadwell and Tianhao Zheng and Todd Underwood and Todor Markov and Toki Sherbakov and Tom Rubin and Tom Stasi and Tomer Kaftan and Tristan Heywood and Troy Peterson and Tyce Walters and Tyna Eloundou and Valerie Qi and Veit Moeller and Vinnie Monaco and Vishal Kuo and Vlad Fomenko and Wayne Chang and Weiyi Zheng and Wenda Zhou and Wesam Manassra and Will Sheu and Wojciech Zaremba and Yash Patil and Yilei Qian and Yongjik Kim and Youlong Cheng and Yu Zhang and Yuchen He and Yuchen Zhang and Yujia Jin and Yunxing Dai and Yury Malkov},
      year={2024},
      eprint={2410.21276},
      archivePrefix={arXiv},
      primaryClass={cs.CL},
      url={https://arxiv.org/abs/2410.21276}, 
}

@misc{openai2025gpt41,
  author       = {{OpenAI}},
  title        = {Introducing GPT-4.1 in the API},
  howpublished = {\url{https://openai.com/index/gpt-4-1/}},
  year         = {2025},
  month        = may,
}

@techreport{anthropic2025claude4systemcard,
  author       = {Anthropic},
  title        = {Claude 4 System Card: Claude Opus 4 \& Claude Sonnet 4},
  institution  = {Anthropic},
  number       = {4263b940cabb546aa0e3283f35b686f4f3b2ff47},
  year         = {2025},
  month        = {may},
  url          = {https://www-cdn.anthropic.com/4263b940cabb546aa0e3283f35b686f4f3b2ff47.pdf},
}

@misc{deepseekai2025deepseekv3technicalreport,
      title={DeepSeek-V3 Technical Report}, 
      author={DeepSeek-AI and Aixin Liu and Bei Feng and Bing Xue and Bingxuan Wang and Bochao Wu and Chengda Lu and Chenggang Zhao and Chengqi Deng and Chenyu Zhang and Chong Ruan and Damai Dai and Daya Guo and Dejian Yang and Deli Chen and Dongjie Ji and Erhang Li and Fangyun Lin and Fucong Dai and Fuli Luo and Guangbo Hao and Guanting Chen and Guowei Li and H. Zhang and Han Bao and Hanwei Xu and Haocheng Wang and Haowei Zhang and Honghui Ding and Huajian Xin and Huazuo Gao and Hui Li and Hui Qu and J. L. Cai and Jian Liang and Jianzhong Guo and Jiaqi Ni and Jiashi Li and Jiawei Wang and Jin Chen and Jingchang Chen and Jingyang Yuan and Junjie Qiu and Junlong Li and Junxiao Song and Kai Dong and Kai Hu and Kaige Gao and Kang Guan and Kexin Huang and Kuai Yu and Lean Wang and Lecong Zhang and Lei Xu and Leyi Xia and Liang Zhao and Litong Wang and Liyue Zhang and Meng Li and Miaojun Wang and Mingchuan Zhang and Minghua Zhang and Minghui Tang and Mingming Li and Ning Tian and Panpan Huang and Peiyi Wang and Peng Zhang and Qiancheng Wang and Qihao Zhu and Qinyu Chen and Qiushi Du and R. J. Chen and R. L. Jin and Ruiqi Ge and Ruisong Zhang and Ruizhe Pan and Runji Wang and Runxin Xu and Ruoyu Zhang and Ruyi Chen and S. S. Li and Shanghao Lu and Shangyan Zhou and Shanhuang Chen and Shaoqing Wu and Shengfeng Ye and Shengfeng Ye and Shirong Ma and Shiyu Wang and Shuang Zhou and Shuiping Yu and Shunfeng Zhou and Shuting Pan and T. Wang and Tao Yun and Tian Pei and Tianyu Sun and W. L. Xiao and Wangding Zeng and Wanjia Zhao and Wei An and Wen Liu and Wenfeng Liang and Wenjun Gao and Wenqin Yu and Wentao Zhang and X. Q. Li and Xiangyue Jin and Xianzu Wang and Xiao Bi and Xiaodong Liu and Xiaohan Wang and Xiaojin Shen and Xiaokang Chen and Xiaokang Zhang and Xiaosha Chen and Xiaotao Nie and Xiaowen Sun and Xiaoxiang Wang and Xin Cheng and Xin Liu and Xin Xie and Xingchao Liu and Xingkai Yu and Xinnan Song and Xinxia Shan and Xinyi Zhou and Xinyu Yang and Xinyuan Li and Xuecheng Su and Xuheng Lin and Y. K. Li and Y. Q. Wang and Y. X. Wei and Y. X. Zhu and Yang Zhang and Yanhong Xu and Yanhong Xu and Yanping Huang and Yao Li and Yao Zhao and Yaofeng Sun and Yaohui Li and Yaohui Wang and Yi Yu and Yi Zheng and Yichao Zhang and Yifan Shi and Yiliang Xiong and Ying He and Ying Tang and Yishi Piao and Yisong Wang and Yixuan Tan and Yiyang Ma and Yiyuan Liu and Yongqiang Guo and Yu Wu and Yuan Ou and Yuchen Zhu and Yuduan Wang and Yue Gong and Yuheng Zou and Yujia He and Yukun Zha and Yunfan Xiong and Yunxian Ma and Yuting Yan and Yuxiang Luo and Yuxiang You and Yuxuan Liu and Yuyang Zhou and Z. F. Wu and Z. Z. Ren and Zehui Ren and Zhangli Sha and Zhe Fu and Zhean Xu and Zhen Huang and Zhen Zhang and Zhenda Xie and Zhengyan Zhang and Zhewen Hao and Zhibin Gou and Zhicheng Ma and Zhigang Yan and Zhihong Shao and Zhipeng Xu and Zhiyu Wu and Zhongyu Zhang and Zhuoshu Li and Zihui Gu and Zijia Zhu and Zijun Liu and Zilin Li and Ziwei Xie and Ziyang Song and Ziyi Gao and Zizheng Pan},
      year={2025},
      eprint={2412.19437},
      archivePrefix={arXiv},
      primaryClass={cs.CL},
      url={https://arxiv.org/abs/2412.19437}, 
}

@misc{kimiteam2025kimik2openagentic,
      title={Kimi K2: Open Agentic Intelligence}, 
      author={Kimi Team and Yifan Bai and Yiping Bao and Guanduo Chen and Jiahao Chen and Ningxin Chen and Ruijue Chen and Yanru Chen and Yuankun Chen and Yutian Chen and Zhuofu Chen and Jialei Cui and Hao Ding and Mengnan Dong and Angang Du and Chenzhuang Du and Dikang Du and Yulun Du and Yu Fan and Yichen Feng and Kelin Fu and Bofei Gao and Hongcheng Gao and Peizhong Gao and Tong Gao and Xinran Gu and Longyu Guan and Haiqing Guo and Jianhang Guo and Hao Hu and Xiaoru Hao and Tianhong He and Weiran He and Wenyang He and Chao Hong and Yangyang Hu and Zhenxing Hu and Weixiao Huang and Zhiqi Huang and Zihao Huang and Tao Jiang and Zhejun Jiang and Xinyi Jin and Yongsheng Kang and Guokun Lai and Cheng Li and Fang Li and Haoyang Li and Ming Li and Wentao Li and Yanhao Li and Yiwei Li and Zhaowei Li and Zheming Li and Hongzhan Lin and Xiaohan Lin and Zongyu Lin and Chengyin Liu and Chenyu Liu and Hongzhang Liu and Jingyuan Liu and Junqi Liu and Liang Liu and Shaowei Liu and T. Y. Liu and Tianwei Liu and Weizhou Liu and Yangyang Liu and Yibo Liu and Yiping Liu and Yue Liu and Zhengying Liu and Enzhe Lu and Lijun Lu and Shengling Ma and Xinyu Ma and Yingwei Ma and Shaoguang Mao and Jie Mei and Xin Men and Yibo Miao and Siyuan Pan and Yebo Peng and Ruoyu Qin and Bowen Qu and Zeyu Shang and Lidong Shi and Shengyuan Shi and Feifan Song and Jianlin Su and Zhengyuan Su and Xinjie Sun and Flood Sung and Heyi Tang and Jiawen Tao and Qifeng Teng and Chensi Wang and Dinglu Wang and Feng Wang and Haiming Wang and Jianzhou Wang and Jiaxing Wang and Jinhong Wang and Shengjie Wang and Shuyi Wang and Yao Wang and Yejie Wang and Yiqin Wang and Yuxin Wang and Yuzhi Wang and Zhaoji Wang and Zhengtao Wang and Zhexu Wang and Chu Wei and Qianqian Wei and Wenhao Wu and Xingzhe Wu and Yuxin Wu and Chenjun Xiao and Xiaotong Xie and Weimin Xiong and Boyu Xu and Jing Xu and Jinjing Xu and L. H. Xu and Lin Xu and Suting Xu and Weixin Xu and Xinran Xu and Yangchuan Xu and Ziyao Xu and Junjie Yan and Yuzi Yan and Xiaofei Yang and Ying Yang and Zhen Yang and Zhilin Yang and Zonghan Yang and Haotian Yao and Xingcheng Yao and Wenjie Ye and Zhuorui Ye and Bohong Yin and Longhui Yu and Enming Yuan and Hongbang Yuan and Mengjie Yuan and Haobing Zhan and Dehao Zhang and Hao Zhang and Wanlu Zhang and Xiaobin Zhang and Yangkun Zhang and Yizhi Zhang and Yongting Zhang and Yu Zhang and Yutao Zhang and Yutong Zhang and Zheng Zhang and Haotian Zhao and Yikai Zhao and Huabin Zheng and Shaojie Zheng and Jianren Zhou and Xinyu Zhou and Zaida Zhou and Zhen Zhu and Weiyu Zhuang and Xinxing Zu},
      year={2025},
      eprint={2507.20534},
      archivePrefix={arXiv},
      primaryClass={cs.LG},
      url={https://arxiv.org/abs/2507.20534}, 
}

@misc{yang2025qwen3technicalreport,
      title={Qwen3 Technical Report}, 
      author={An Yang and Anfeng Li and Baosong Yang and Beichen Zhang and Binyuan Hui and Bo Zheng and Bowen Yu and Chang Gao and Chengen Huang and Chenxu Lv and Chujie Zheng and Dayiheng Liu and Fan Zhou and Fei Huang and Feng Hu and Hao Ge and Haoran Wei and Huan Lin and Jialong Tang and Jian Yang and Jianhong Tu and Jianwei Zhang and Jianxin Yang and Jiaxi Yang and Jing Zhou and Jingren Zhou and Junyang Lin and Kai Dang and Keqin Bao and Kexin Yang and Le Yu and Lianghao Deng and Mei Li and Mingfeng Xue and Mingze Li and Pei Zhang and Peng Wang and Qin Zhu and Rui Men and Ruize Gao and Shixuan Liu and Shuang Luo and Tianhao Li and Tianyi Tang and Wenbiao Yin and Xingzhang Ren and Xinyu Wang and Xinyu Zhang and Xuancheng Ren and Yang Fan and Yang Su and Yichang Zhang and Yinger Zhang and Yu Wan and Yuqiong Liu and Zekun Wang and Zeyu Cui and Zhenru Zhang and Zhipeng Zhou and Zihan Qiu},
      year={2025},
      eprint={2505.09388},
      archivePrefix={arXiv},
      primaryClass={cs.CL},
      url={https://arxiv.org/abs/2505.09388}, 
}

@misc{gemmateam2025gemma3technicalreport,
      title={Gemma 3 Technical Report}, 
      author={Gemma Team and Aishwarya Kamath and Johan Ferret and Shreya Pathak and Nino Vieillard and Ramona Merhej and Sarah Perrin and Tatiana Matejovicova and Alexandre Ramé and Morgane Rivière and Louis Rouillard and Thomas Mesnard and Geoffrey Cideron and Jean-bastien Grill and Sabela Ramos and Edouard Yvinec and Michelle Casbon and Etienne Pot and Ivo Penchev and Gaël Liu and Francesco Visin and Kathleen Kenealy and Lucas Beyer and Xiaohai Zhai and Anton Tsitsulin and Robert Busa-Fekete and Alex Feng and Noveen Sachdeva and Benjamin Coleman and Yi Gao and Basil Mustafa and Iain Barr and Emilio Parisotto and David Tian and Matan Eyal and Colin Cherry and Jan-Thorsten Peter and Danila Sinopalnikov and Surya Bhupatiraju and Rishabh Agarwal and Mehran Kazemi and Dan Malkin and Ravin Kumar and David Vilar and Idan Brusilovsky and Jiaming Luo and Andreas Steiner and Abe Friesen and Abhanshu Sharma and Abheesht Sharma and Adi Mayrav Gilady and Adrian Goedeckemeyer and Alaa Saade and Alex Feng and Alexander Kolesnikov and Alexei Bendebury and Alvin Abdagic and Amit Vadi and András György and André Susano Pinto and Anil Das and Ankur Bapna and Antoine Miech and Antoine Yang and Antonia Paterson and Ashish Shenoy and Ayan Chakrabarti and Bilal Piot and Bo Wu and Bobak Shahriari and Bryce Petrini and Charlie Chen and Charline Le Lan and Christopher A. Choquette-Choo and CJ Carey and Cormac Brick and Daniel Deutsch and Danielle Eisenbud and Dee Cattle and Derek Cheng and Dimitris Paparas and Divyashree Shivakumar Sreepathihalli and Doug Reid and Dustin Tran and Dustin Zelle and Eric Noland and Erwin Huizenga and Eugene Kharitonov and Frederick Liu and Gagik Amirkhanyan and Glenn Cameron and Hadi Hashemi and Hanna Klimczak-Plucińska and Harman Singh and Harsh Mehta and Harshal Tushar Lehri and Hussein Hazimeh and Ian Ballantyne and Idan Szpektor and Ivan Nardini and Jean Pouget-Abadie and Jetha Chan and Joe Stanton and John Wieting and Jonathan Lai and Jordi Orbay and Joseph Fernandez and Josh Newlan and Ju-yeong Ji and Jyotinder Singh and Kat Black and Kathy Yu and Kevin Hui and Kiran Vodrahalli and Klaus Greff and Linhai Qiu and Marcella Valentine and Marina Coelho and Marvin Ritter and Matt Hoffman and Matthew Watson and Mayank Chaturvedi and Michael Moynihan and Min Ma and Nabila Babar and Natasha Noy and Nathan Byrd and Nick Roy and Nikola Momchev and Nilay Chauhan and Noveen Sachdeva and Oskar Bunyan and Pankil Botarda and Paul Caron and Paul Kishan Rubenstein and Phil Culliton and Philipp Schmid and Pier Giuseppe Sessa and Pingmei Xu and Piotr Stanczyk and Pouya Tafti and Rakesh Shivanna and Renjie Wu and Renke Pan and Reza Rokni and Rob Willoughby and Rohith Vallu and Ryan Mullins and Sammy Jerome and Sara Smoot and Sertan Girgin and Shariq Iqbal and Shashir Reddy and Shruti Sheth and Siim Põder and Sijal Bhatnagar and Sindhu Raghuram Panyam and Sivan Eiger and Susan Zhang and Tianqi Liu and Trevor Yacovone and Tyler Liechty and Uday Kalra and Utku Evci and Vedant Misra and Vincent Roseberry and Vlad Feinberg and Vlad Kolesnikov and Woohyun Han and Woosuk Kwon and Xi Chen and Yinlam Chow and Yuvein Zhu and Zichuan Wei and Zoltan Egyed and Victor Cotruta and Minh Giang and Phoebe Kirk and Anand Rao and Kat Black and Nabila Babar and Jessica Lo and Erica Moreira and Luiz Gustavo Martins and Omar Sanseviero and Lucas Gonzalez and Zach Gleicher and Tris Warkentin and Vahab Mirrokni and Evan Senter and Eli Collins and Joelle Barral and Zoubin Ghahramani and Raia Hadsell and Yossi Matias and D. Sculley and Slav Petrov and Noah Fiedel and Noam Shazeer and Oriol Vinyals and Jeff Dean and Demis Hassabis and Koray Kavukcuoglu and Clement Farabet and Elena Buchatskaya and Jean-Baptiste Alayrac and Rohan Anil and Dmitry and Lepikhin and Sebastian Borgeaud and Olivier Bachem and Armand Joulin and Alek Andreev and Cassidy Hardin and Robert Dadashi and Léonard Hussenot},
      year={2025},
      eprint={2503.19786},
      archivePrefix={arXiv},
      primaryClass={cs.CL},
      url={https://arxiv.org/abs/2503.19786}, 
}

@misc{grattafiori2024llama3herdmodels,
      title={The Llama 3 Herd of Models}, 
      author={Aaron Grattafiori and Abhimanyu Dubey and Abhinav Jauhri and Abhinav Pandey and Abhishek Kadian and Ahmad Al-Dahle and Aiesha Letman and Akhil Mathur and Alan Schelten and Alex Vaughan and Amy Yang and Angela Fan and Anirudh Goyal and Anthony Hartshorn and Aobo Yang and Archi Mitra and Archie Sravankumar and Artem Korenev and Arthur Hinsvark and Arun Rao and Aston Zhang and Aurelien Rodriguez and Austen Gregerson and Ava Spataru and Baptiste Roziere and Bethany Biron and Binh Tang and Bobbie Chern and Charlotte Caucheteux and Chaya Nayak and Chloe Bi and Chris Marra and Chris McConnell and Christian Keller and Christophe Touret and Chunyang Wu and Corinne Wong and Cristian Canton Ferrer and Cyrus Nikolaidis and Damien Allonsius and Daniel Song and Danielle Pintz and Danny Livshits and Danny Wyatt and David Esiobu and Dhruv Choudhary and Dhruv Mahajan and Diego Garcia-Olano and Diego Perino and Dieuwke Hupkes and Egor Lakomkin and Ehab AlBadawy and Elina Lobanova and Emily Dinan and Eric Michael Smith and Filip Radenovic and Francisco Guzmán and Frank Zhang and Gabriel Synnaeve and Gabrielle Lee and Georgia Lewis Anderson and Govind Thattai and Graeme Nail and Gregoire Mialon and Guan Pang and Guillem Cucurell and Hailey Nguyen and Hannah Korevaar and Hu Xu and Hugo Touvron and Iliyan Zarov and Imanol Arrieta Ibarra and Isabel Kloumann and Ishan Misra and Ivan Evtimov and Jack Zhang and Jade Copet and Jaewon Lee and Jan Geffert and Jana Vranes and Jason Park and Jay Mahadeokar and Jeet Shah and Jelmer van der Linde and Jennifer Billock and Jenny Hong and Jenya Lee and Jeremy Fu and Jianfeng Chi and Jianyu Huang and Jiawen Liu and Jie Wang and Jiecao Yu and Joanna Bitton and Joe Spisak and Jongsoo Park and Joseph Rocca and Joshua Johnstun and Joshua Saxe and Junteng Jia and Kalyan Vasuden Alwala and Karthik Prasad and Kartikeya Upasani and Kate Plawiak and Ke Li and Kenneth Heafield and Kevin Stone and Khalid El-Arini and Krithika Iyer and Kshitiz Malik and Kuenley Chiu and Kunal Bhalla and Kushal Lakhotia and Lauren Rantala-Yeary and Laurens van der Maaten and Lawrence Chen and Liang Tan and Liz Jenkins and Louis Martin and Lovish Madaan and Lubo Malo and Lukas Blecher and Lukas Landzaat and Luke de Oliveira and Madeline Muzzi and Mahesh Pasupuleti and Mannat Singh and Manohar Paluri and Marcin Kardas and Maria Tsimpoukelli and Mathew Oldham and Mathieu Rita and Maya Pavlova and Melanie Kambadur and Mike Lewis and Min Si and Mitesh Kumar Singh and Mona Hassan and Naman Goyal and Narjes Torabi and Nikolay Bashlykov and Nikolay Bogoychev and Niladri Chatterji and Ning Zhang and Olivier Duchenne and Onur Çelebi and Patrick Alrassy and Pengchuan Zhang and Pengwei Li and Petar Vasic and Peter Weng and Prajjwal Bhargava and Pratik Dubal and Praveen Krishnan and Punit Singh Koura and Puxin Xu and Qing He and Qingxiao Dong and Ragavan Srinivasan and Raj Ganapathy and Ramon Calderer and Ricardo Silveira Cabral and Robert Stojnic and Roberta Raileanu and Rohan Maheswari and Rohit Girdhar and Rohit Patel and Romain Sauvestre and Ronnie Polidoro and Roshan Sumbaly and Ross Taylor and Ruan Silva and Rui Hou and Rui Wang and Saghar Hosseini and Sahana Chennabasappa and Sanjay Singh and Sean Bell and Seohyun Sonia Kim and Sergey Edunov and Shaoliang Nie and Sharan Narang and Sharath Raparthy and Sheng Shen and Shengye Wan and Shruti Bhosale and Shun Zhang and Simon Vandenhende and Soumya Batra and Spencer Whitman and Sten Sootla and Stephane Collot and Suchin Gururangan and Sydney Borodinsky and Tamar Herman and Tara Fowler and Tarek Sheasha and Thomas Georgiou and Thomas Scialom and Tobias Speckbacher and Todor Mihaylov and Tong Xiao and Ujjwal Karn and Vedanuj Goswami and Vibhor Gupta and Vignesh Ramanathan and Viktor Kerkez and Vincent Gonguet and Virginie Do and Vish Vogeti and Vítor Albiero and Vladan Petrovic and Weiwei Chu and Wenhan Xiong and Wenyin Fu and Whitney Meers and Xavier Martinet and Xiaodong Wang and Xiaofang Wang and Xiaoqing Ellen Tan and Xide Xia and Xinfeng Xie and Xuchao Jia and Xuewei Wang and Yaelle Goldschlag and Yashesh Gaur and Yasmine Babaei and Yi Wen and Yiwen Song and Yuchen Zhang and Yue Li and Yuning Mao and Zacharie Delpierre Coudert and Zheng Yan and Zhengxing Chen and Zoe Papakipos and Aaditya Singh and Aayushi Srivastava and Abha Jain and Adam Kelsey and Adam Shajnfeld and Adithya Gangidi and Adolfo Victoria and Ahuva Goldstand and Ajay Menon and Ajay Sharma and Alex Boesenberg and Alexei Baevski and Allie Feinstein and Amanda Kallet and Amit Sangani and Amos Teo and Anam Yunus and Andrei Lupu and Andres Alvarado and Andrew Caples and Andrew Gu and Andrew Ho and Andrew Poulton and Andrew Ryan and Ankit Ramchandani and Annie Dong and Annie Franco and Anuj Goyal and Aparajita Saraf and Arkabandhu Chowdhury and Ashley Gabriel and Ashwin Bharambe and Assaf Eisenman and Azadeh Yazdan and Beau James and Ben Maurer and Benjamin Leonhardi and Bernie Huang and Beth Loyd and Beto De Paola and Bhargavi Paranjape and Bing Liu and Bo Wu and Boyu Ni and Braden Hancock and Bram Wasti and Brandon Spence and Brani Stojkovic and Brian Gamido and Britt Montalvo and Carl Parker and Carly Burton and Catalina Mejia and Ce Liu and Changhan Wang and Changkyu Kim and Chao Zhou and Chester Hu and Ching-Hsiang Chu and Chris Cai and Chris Tindal and Christoph Feichtenhofer and Cynthia Gao and Damon Civin and Dana Beaty and Daniel Kreymer and Daniel Li and David Adkins and David Xu and Davide Testuggine and Delia David and Devi Parikh and Diana Liskovich and Didem Foss and Dingkang Wang and Duc Le and Dustin Holland and Edward Dowling and Eissa Jamil and Elaine Montgomery and Eleonora Presani and Emily Hahn and Emily Wood and Eric-Tuan Le and Erik Brinkman and Esteban Arcaute and Evan Dunbar and Evan Smothers and Fei Sun and Felix Kreuk and Feng Tian and Filippos Kokkinos and Firat Ozgenel and Francesco Caggioni and Frank Kanayet and Frank Seide and Gabriela Medina Florez and Gabriella Schwarz and Gada Badeer and Georgia Swee and Gil Halpern and Grant Herman and Grigory Sizov and Guangyi and Zhang and Guna Lakshminarayanan and Hakan Inan and Hamid Shojanazeri and Han Zou and Hannah Wang and Hanwen Zha and Haroun Habeeb and Harrison Rudolph and Helen Suk and Henry Aspegren and Hunter Goldman and Hongyuan Zhan and Ibrahim Damlaj and Igor Molybog and Igor Tufanov and Ilias Leontiadis and Irina-Elena Veliche and Itai Gat and Jake Weissman and James Geboski and James Kohli and Janice Lam and Japhet Asher and Jean-Baptiste Gaya and Jeff Marcus and Jeff Tang and Jennifer Chan and Jenny Zhen and Jeremy Reizenstein and Jeremy Teboul and Jessica Zhong and Jian Jin and Jingyi Yang and Joe Cummings and Jon Carvill and Jon Shepard and Jonathan McPhie and Jonathan Torres and Josh Ginsburg and Junjie Wang and Kai Wu and Kam Hou U and Karan Saxena and Kartikay Khandelwal and Katayoun Zand and Kathy Matosich and Kaushik Veeraraghavan and Kelly Michelena and Keqian Li and Kiran Jagadeesh and Kun Huang and Kunal Chawla and Kyle Huang and Lailin Chen and Lakshya Garg and Lavender A and Leandro Silva and Lee Bell and Lei Zhang and Liangpeng Guo and Licheng Yu and Liron Moshkovich and Luca Wehrstedt and Madian Khabsa and Manav Avalani and Manish Bhatt and Martynas Mankus and Matan Hasson and Matthew Lennie and Matthias Reso and Maxim Groshev and Maxim Naumov and Maya Lathi and Meghan Keneally and Miao Liu and Michael L. Seltzer and Michal Valko and Michelle Restrepo and Mihir Patel and Mik Vyatskov and Mikayel Samvelyan and Mike Clark and Mike Macey and Mike Wang and Miquel Jubert Hermoso and Mo Metanat and Mohammad Rastegari and Munish Bansal and Nandhini Santhanam and Natascha Parks and Natasha White and Navyata Bawa and Nayan Singhal and Nick Egebo and Nicolas Usunier and Nikhil Mehta and Nikolay Pavlovich Laptev and Ning Dong and Norman Cheng and Oleg Chernoguz and Olivia Hart and Omkar Salpekar and Ozlem Kalinli and Parkin Kent and Parth Parekh and Paul Saab and Pavan Balaji and Pedro Rittner and Philip Bontrager and Pierre Roux and Piotr Dollar and Polina Zvyagina and Prashant Ratanchandani and Pritish Yuvraj and Qian Liang and Rachad Alao and Rachel Rodriguez and Rafi Ayub and Raghotham Murthy and Raghu Nayani and Rahul Mitra and Rangaprabhu Parthasarathy and Raymond Li and Rebekkah Hogan and Robin Battey and Rocky Wang and Russ Howes and Ruty Rinott and Sachin Mehta and Sachin Siby and Sai Jayesh Bondu and Samyak Datta and Sara Chugh and Sara Hunt and Sargun Dhillon and Sasha Sidorov and Satadru Pan and Saurabh Mahajan and Saurabh Verma and Seiji Yamamoto and Sharadh Ramaswamy and Shaun Lindsay and Shaun Lindsay and Sheng Feng and Shenghao Lin and Shengxin Cindy Zha and Shishir Patil and Shiva Shankar and Shuqiang Zhang and Shuqiang Zhang and Sinong Wang and Sneha Agarwal and Soji Sajuyigbe and Soumith Chintala and Stephanie Max and Stephen Chen and Steve Kehoe and Steve Satterfield and Sudarshan Govindaprasad and Sumit Gupta and Summer Deng and Sungmin Cho and Sunny Virk and Suraj Subramanian and Sy Choudhury and Sydney Goldman and Tal Remez and Tamar Glaser and Tamara Best and Thilo Koehler and Thomas Robinson and Tianhe Li and Tianjun Zhang and Tim Matthews and Timothy Chou and Tzook Shaked and Varun Vontimitta and Victoria Ajayi and Victoria Montanez and Vijai Mohan and Vinay Satish Kumar and Vishal Mangla and Vlad Ionescu and Vlad Poenaru and Vlad Tiberiu Mihailescu and Vladimir Ivanov and Wei Li and Wenchen Wang and Wenwen Jiang and Wes Bouaziz and Will Constable and Xiaocheng Tang and Xiaojian Wu and Xiaolan Wang and Xilun Wu and Xinbo Gao and Yaniv Kleinman and Yanjun Chen and Ye Hu and Ye Jia and Ye Qi and Yenda Li and Yilin Zhang and Ying Zhang and Yossi Adi and Youngjin Nam and Yu and Wang and Yu Zhao and Yuchen Hao and Yundi Qian and Yunlu Li and Yuzi He and Zach Rait and Zachary DeVito and Zef Rosnbrick and Zhaoduo Wen and Zhenyu Yang and Zhiwei Zhao and Zhiyu Ma},
      year={2024},
      eprint={2407.21783},
      archivePrefix={arXiv},
      primaryClass={cs.AI},
      url={https://arxiv.org/abs/2407.21783}, 
}

@inproceedings{
liu2023is,
title={Is Your Code Generated by Chat{GPT} Really Correct? Rigorous Evaluation of Large Language Models for Code Generation},
author={Jiawei Liu and Chunqiu Steven Xia and Yuyao Wang and LINGMING ZHANG},
booktitle={Thirty-seventh Conference on Neural Information Processing Systems},
year={2023},
url={https://openreview.net/forum?id=1qvx610Cu7}
}

@inproceedings{wang-etal-2025-codecontests,
    title = "{C}ode{C}ontests+: High-Quality Test Case Generation for Competitive Programming",
    author = "Wang, Zihan  and
      Liu, Siyao  and
      Sun, Yang  and
      Ding, Ming  and
      Li, Hongyan",
    editor = "Christodoulopoulos, Christos  and
      Chakraborty, Tanmoy  and
      Rose, Carolyn  and
      Peng, Violet",
    booktitle = "Findings of the Association for Computational Linguistics: EMNLP 2025",
    month = nov,
    year = "2025",
    address = "Suzhou, China",
    publisher = "Association for Computational Linguistics",
    url = "https://aclanthology.org/2025.findings-emnlp.299/",
    pages = "5576--5600",
    ISBN = "979-8-89176-335-7"
}

@misc{zhang2025codingtriangledoeslarge,
      title={Coding Triangle: How Does Large Language Model Understand Code?}, 
      author={Taolin Zhang and Zihan Ma and Maosong Cao and Junnan Liu and Songyang Zhang and Kai Chen},
      year={2025},
      eprint={2507.06138},
      archivePrefix={arXiv},
      primaryClass={cs.CL},
      url={https://arxiv.org/abs/2507.06138}, 
}

@inproceedings{yang-etal-2025-elaboration,
    title = "{ELABORATION}: A Comprehensive Benchmark on Human-{LLM} Competitive Programming",
    author = "Yang, Xinwei  and
      Liu, Zhaofeng  and
      Huang, Chen  and
      Zhang, Jiashuai  and
      Zhang, Tong  and
      Zhang, Yifan  and
      Lei, Wenqiang",
    editor = "Che, Wanxiang  and
      Nabende, Joyce  and
      Shutova, Ekaterina  and
      Pilehvar, Mohammad Taher",
    booktitle = "Proceedings of the 63rd Annual Meeting of the Association for Computational Linguistics (Volume 1: Long Papers)",
    month = jul,
    year = "2025",
    address = "Vienna, Austria",
    publisher = "Association for Computational Linguistics",
    url = "https://aclanthology.org/2025.acl-long.4/",
    doi = "10.18653/v1/2025.acl-long.4",
    pages = "59--104",
    ISBN = "979-8-89176-251-0"
}

@inproceedings{islam-etal-2024-mapcoder,
    title = "{M}ap{C}oder: Multi-Agent Code Generation for Competitive Problem Solving",
    author = "Islam, Md. Ashraful  and
      Ali, Mohammed Eunus  and
      Parvez, Md Rizwan",
    editor = "Ku, Lun-Wei  and
      Martins, Andre  and
      Srikumar, Vivek",
    booktitle = "Proceedings of the 62nd Annual Meeting of the Association for Computational Linguistics (Volume 1: Long Papers)",
    month = aug,
    year = "2024",
    address = "Bangkok, Thailand",
    publisher = "Association for Computational Linguistics",
    url = "https://aclanthology.org/2024.acl-long.269/",
    doi = "10.18653/v1/2024.acl-long.269",
    pages = "4912--4944"
}

@inproceedings{
chen2023codet,
title={CodeT:  Code Generation with Generated Tests},
author={Bei Chen and Fengji Zhang and Anh Nguyen and Daoguang Zan and Zeqi Lin and Jian-Guang Lou and Weizhu Chen},
booktitle={The Eleventh International Conference on Learning Representations },
year={2023},
url={https://openreview.net/forum?id=ktrw68Cmu9c}
}

@techreport{google_deepmind_gemini3pro_model_card,
  author      = {{Google DeepMind}},
  title       = {{Gemini 3 Pro Model Card}},
  institution = {Google DeepMind},
  year        = {2025},
  type        = {Model Card},
  url         = {https://storage.googleapis.com/deepmind-media/Model-Cards/Gemini-3-Pro-Model-Card.pdf},
}

@book{halim2020competitive,
  title={Competitive Programming 4 - Book 2: The Lower Bound of Programming Contests in the 2020s},
  author={Halim, S. and Halim, F. and Effendy, S.},
  number={bk. 2},
  isbn={9781716745515},
  url={https://books.google.com.eg/books?id=CBnDzQEACAAJ},
  year={2020},
}

\appendix

\section{Example Problem and Editorial}
\label{appendix:example-problem}

This appendix shows an example problem and its corresponding gold editorial from our dataset. Figure~\ref{fig:app-example-problem} shows a representative competitive-programming problem and illustrates the problem format used throughout our dataset. Figure~\ref{fig:app-gold-editorial} shows an example gold editorial, illustrating the level of algorithmic detail expected from expert-written editorials.


\begin{figure*}
\begin{tcolorbox}[ title={Example Problem Statement}]

\small
\textbf{Contest:} CS3233 Midterm Contest 2025

\textbf{Problem title:} Arts and Computing Students

\textbf{Time limit:} 1 second

\textbf{Memory limit:} 1024 MB
\normalsize

\vspace{8pt}
\hrule
\vspace{8pt}
\small
\textbf{Problem Statement.}

To encourage interdisciplinary collaboration, the School of Computing has decided to hold a dinner event for students in School of Computing and the Faculty of Arts and Social Sciences. The event will be held in a large hall with a long table that can seat $2N$ students. The $2N$ seats are arranged on a single line, with $N$ pairs of seats. Seats belonging to the same pair are adjacent in the line. Each seat will either contain a student from the School of Computing, a student from the Faculty of Arts and Social Sciences, or be empty. Two students sitting next to each other within the same pair of seats are called ``buddies''. There are at most $N$ pairs of buddies by definition.

The Dean of the School of Computing suspects that School of Computing students will start Leetcoding on their laptops if they do not have a student from the Faculty of Arts and Social Sciences as a buddy. To prevent this, the Dean wants to rearrange the students such that each student from the School of Computing has a buddy from the Faculty of Arts and Social Sciences.

The Dean can perform the following operation: he can choose a student sitting next to an empty seat and move that student to the empty seat. These two seats need not belong to the same pair.

Given the initial arrangement of students, determine whether it is possible to rearrange the students such that each student from the School of Computing has a buddy from the Faculty of Arts and Social Sciences.

\vspace{6pt}
\textbf{Input.}

The first line contains an integer $N$ $(1 \le N \le 2\times 10^{5})$, the number of pairs of seats.

The second line contains a string of length $2N$ consisting of the characters \texttt{C}, \texttt{A}, and \texttt{.}. The $i$-th character is \texttt{C} if the student in the $i$-th seat is from the School of Computing, \texttt{A} if the student is from the Faculty of Arts and Social Sciences, and \texttt{.} if the seat is empty.

\vspace{6pt}

\textbf{Output.}

If it is not possible to rearrange the students such that each \texttt{C} has an adjacent \texttt{A} in its pair, output:
\begin{verbatim}
NO
\end{verbatim}

Otherwise, output:
\begin{verbatim}
YES
<final-arrangement-string>
\end{verbatim}
The second line must be a string of length $2N$ over \texttt{\{C,A,.\}} representing a final valid arrangement after allowed moves. Any valid final arrangement may be output.

\vspace{6pt}
\textbf{Sample Input \#1}
\begin{verbatim}
3
C.AAAC
\end{verbatim}

\textbf{Sample Output \#1}
\begin{verbatim}
YES
CAA.AC
\end{verbatim}

\textbf{Sample Input \#2}
\begin{verbatim}
4
AAA.CCAC
\end{verbatim}

\textbf{Sample Output \#2}
\begin{verbatim}
NO
\end{verbatim}

\textbf{Sample Input \#3}
\begin{verbatim}
4
CA.A.C.A
\end{verbatim}

\textbf{Sample Output \#3}
\begin{verbatim}
YES
CA..ACA.
\end{verbatim}
\end{tcolorbox}
\caption{Example competitive-programming problem from our dataset (\textsc{Arts and Computing Students}).}
\label{fig:app-example-problem}
\end{figure*}

\begin{figure*}
\begin{tcolorbox}[ title={Example Gold Editorial: \emph{Arts and Computing Students}}]

\small
There are two solutions to this problem:

\begin{itemize}
    \item \textbf{Dynamic Programming}: Observe that our ultimate goal is to pair up the students as much as possible, as we risk running out of space (there are only $N$ pairs of seats) if we do not pair them up sufficiently. We can let $\texttt{dp}[i]$ be the minimum number of pairs needed to pair up students $1, 2, \ldots, i$ such that the constraint is satisfied. To compute $\texttt{dp}[i]$, it suffices to condition on whether student $i$ is paired with student $i - 1$:

    $$
    \texttt{dp}[i] = \min \left\{
    \begin{array}{ll}
    \texttt{dp}[i - 2] + 1 & \text{if student } i \text{ and } i-1 \text{ forms a valid pair} \\
    \texttt{dp}[i - 1] + 1 & \text{if student } i \text{ can sit alone (i.e. arts student)}
    \end{array}
    \right.
    $$

    A solution exists if $\texttt{dp}[\#\text{ students}] \leq N$, and we can backtrack to find the construction.

    \item \textbf{Greedy}: Notice that if we have a computing student at the beginning / end of the array or we have two consecutive computing students, then their seating arrangements are fixed. (For example, if we have ACCA as a substring, we know that the pairs must be AC and CA.) We first split the string whenever we see \textbf{ACCA}. For the remaining chunks without two consecutive computing students, we can greedily seat them using $\left\lfloor \frac{\# \text{students}}{2} \right\rfloor$ pairs (if we have more computing students than arts students, then this is clearly impossible). It is easy to see that this greedy strategy minimizes the number of pairs needed to seat all students.
\end{itemize}

\vspace{4pt}
\noindent\textbf{P.S.} Many teams attempted to code a (wrong) greedy solution right away. It’s a better idea to (informally) justify the correctness of your greedy approach or try coming up with counter-examples before coding. When in doubt, dynamic programming is the safe choice to go with, as long as it’s fast enough!
\normalsize

\end{tcolorbox}
\caption{Gold editorial for \textsc{Arts and Computing Students}.}
\label{fig:app-gold-editorial}
\end{figure*}

\section{Dataset Details and Metadata}
\label{app:data-details}

\paragraph{Provenance and contest coverage.}
Our dataset contains 83 ICPC-style problems drawn from seven contests spanning 2017--2025:
three CS3233 midterm contests hosted on NUS (course instance) and four regional ICPC contests
distributed via public task repositories.
Table~\ref{tab:dataset-contests} summarizes contest-level statistics.

\paragraph{Copyright and permissions.}
The CS3233 portion of the dataset consists of course assessment materials from the
National University of Singapore; we requested copyright permission from the course
instructor to include and redistribute these materials (problem statements, gold editorials)
as part of our dataset release.
The CS3233 gold editorials are private course materials that were not publicly released
prior to this work.

\paragraph{Release plan.}
We will release the dataset upon publication, including problem statements, gold editorials,
and the full official test suites / judging harness.

\begin{table*}
\centering
\small
\setlength{\tabcolsep}{4pt}
\begin{tabular}{lcccc}
\toprule
\textbf{Contest} & \textbf{Year} & \textbf{Source} & \textbf{\#Teams} & \textbf{\#Problems} \\
\midrule
CS3233 Midterm Contest & 2023 & NUS  & 25 & 11 \\
CS3233 Midterm Contest & 2024 & NUS  & 15 & 12 \\
CS3233 Midterm Contest & 2025 & NUS  & 16 & 11 \\
ICPC Asia Pacific Championship & 2024 & GitHub & 65 & 13 \\
ICPC Asia Jakarta Regional & 2017 & GitHub & 80 & 12 \\
ICPC Asia Jakarta Regional & 2018 & GitHub & 75 & 12 \\
ICPC Asia Jakarta Regional & 2019 & GitHub & 80 & 12 \\
\midrule
\textbf{Total} & -- & -- & -- & \textbf{83} \\
\bottomrule
\end{tabular}
\caption{
Contest-level composition of the dataset.
}
\label{tab:dataset-contests}
\end{table*}

\section{Judging Protocol Details}
\label{app:judging}

We evaluated all generated programs using a standard ICPC-style compile-and-run judging
pipeline in the official contest test suites.

For each submission, we first attempt to compile the program (or perform a syntax check
for interpreted languages). Compilation failures are labeled \textsc{CE} (Compile Error).
Otherwise, the resulting executable is run in each test case under the specified time
and memory limits. Executions that exceed the time or memory budget are labeled
\textsc{TLE} (Time Limit Exceeded) or \textsc{MLE} (Memory Limit Exceeded), respectively,
and abnormal termination (e.g., segmentation faults or runtime exceptions) is labeled
\textsc{RTE} (Runtime Error).

If the execution is completed successfully, the program’s output is compared against the
reference output provided by the contest judges; any mismatch yields \textsc{WA}
(Wrong Answer). The first failure verdict encountered during the testing is reported as the
result $T(C)$. A submission is labeled \textsc{PASS} only if it compiles successfully
and produces correct output for all test cases within the prescribed resource limits.

Submissions that do not yield a usable program (e.g., explicit refusals, incomplete code,
or outputs in the wrong programming language) are conservatively treated as \textsc{CE}
or \textsc{RTE}, consistent with standard contest judging practice. These rare failure
modes are analyzed separately in the Appendix~\ref{appendix:failure-dist}.

\section{Editorial Annotation Rubric}
\label{appendix:annotation-rubric}

Figure~\ref{fig:annotation-rubric} provides the full rubric used for annotating model-generated editorials. The rubric is organized into three components: Problem Understanding (PU), Algorithm Description (ALG), and Algorithm Correctness (ALG-COR). Each component includes detailed fields and rating options used by annotators.
\begin{figure*}

\begin{tcolorbox}[ title={Editorial Annotation Rubric}]
\small

\textbf{1.Problem Understanding (PU)}

\medskip

\textbf{Purpose:} Verify that the editorial accurately captures every essential detail of the problem statement, without misinterpreting constraints, omitting crucial subtleties, or introducing misleading additions.

\medskip

\textbf{PU-W — Wrong crucial detail}
Does the editorial assert something that changes the problem’s meaning?  
Options: \textit{Yes / No}.  
If \textit{Yes}, specify whether the misinformation is \textit{explicit} or \textit{implicit}.

\textbf{PU-M — Missing crucial detail}
Does the editorial omit a constraint or subtlety that affects the meaning of the problem?  
Options: \textit{Yes / No}.  
If \textit{Yes}, note whether the missing information is \textit{explicit} or \textit{implicit} in the statement.

\textbf{PU-X — Irrelevant or misleading detail}
Extra statements that do not affect correctness but muddy understanding.  
Options: \textit{None / Minor / Major}.

\textbf{PU-D — Problem Understanding Difficulty}
Annotator's assessment of how difficult the problem is to understand.  
Options: scale \textit{0–5}, where 0 = very clear and 5 = extremely difficult.

\bigskip

\textbf{2.Algorithm Description (ALG)}

\medskip

\textbf{Purpose:} Assess whether the editorial presents a coherent and appropriate high-level algorithmic idea.

\medskip

\textbf{ALG-TAG}
Algorithmic paradigm(s) described in the model-generated editorial.  
Annotators select one or more from:  
\textit{DP, Greedy, DFS/BFS, Dijkstra, Segment Tree, Binary Lifting, FFT, Flow, Geometry, Math/Number Theory, Other}.

\textbf{ALG-FREE}
A concise ($\approx40$ words) free-text summary of the core idea in the model-generated editorial.

\textbf{Golden-ALG-TAG}
Algorithmic paradigm(s) described in the gold-standard editorial, using the same tag set as above.

\textbf{Golden-ALG-FREE}
A concise ($\approx40$ words) summary of the corresponding idea in the gold editorial.

\bigskip

\textbf{3.Algorithm Correctness (ALG-COR)}

\medskip
\medskip
\textbf{Purpose:} Determine whether the editorial’s algorithm—\textit{as described}—solves the problem correctly and efficiently under the stated constraints.

\medskip

\textbf{ALG-COR}
Options: \textit{Correct / Incorrect}.



\medskip
\paragraph*{If Correct}

\textbf{Correct Type}
Options:  
\begin{itemize}
    \item \textbf{Same as golden} — matches the official solution exactly.  
          Example: identifies the correct DP recurrence just as in the gold editorial.
    \item \textbf{Different from golden} — uses a different but equally valid algorithm.  
          Example: employs a greedy method instead of DP, but still satisfies all constraints.
\end{itemize}
\medskip
\paragraph*{If Incorrect}

Annotators diagnose why the approach fails.

\textbf{Why Incorrect}
Options:
\begin{enumerate}
    \item \textbf{Wrong algorithm (WA)}  
          The high-level idea is unsuitable; it cannot solve the full problem.  
          Example: tries to solve a graph problem using DP on sequences.

    \item \textbf{Correct algorithm but incorrect approach (WA)}  
          The algorithmic idea is sound, but the reasoning includes a mistake that breaks correctness.  
          Example: binary search is intended but the midpoint update never converges.

    \item \textbf{Suboptimal (Likely TLE or MLE), but correct algorithm}  
          The method returns correct answers but violates resource limits in worst-case scenarios.  
          Example: uses an \(O(n^2)\) technique where \(O(n \log n)\) is required.

    \item \textbf{Suboptimal (Likely TLE or MLE), and wrong algorithm}  
          The algorithm is both inefficient and structurally unsuitable.  
          Example: brute force or an unnecessary DP when a linear greedy method is required.
\end{enumerate}

\textbf{Severity of Incorrectness}
Options:
\begin{itemize}
    \item \textbf{Completely wrong}  
          No part of the solution can be salvaged; the idea is unrelated to a valid strategy.
    \item \textbf{Major edits needed}  
          The structure almost works but requires a significant reworking of key components.
    \item \textbf{Minor edits needed}  
          Small adjustments (e.g., off-by-one fix, missing boundary case, or initialization) restore correctness.
\end{itemize}
\end{tcolorbox}
\caption{
Editorial annotation rubric used for both expert evaluation.
The rubric decomposes reasoning quality into problem understanding (PU),
algorithm description (ALG), and algorithm correctness (ALG-COR).
}
\label{fig:annotation-rubric}

\end{figure*}

\section{Prompt templates}
\label{appendix:prompts}

Figures~\ref{fig:prompt-codegen} and~\ref{fig:prompt-editorial} show the exact prompts
used for code generation and editorial generation in our experiments. The wording below is preserved verbatim; only formatting has been changed
to improve readability.

\begin{figure*}

\begin{tcolorbox}[ title={Code-generation prompts (w/oEd and w/Ed)}]

\small
\textbf{Without editorial (w/oEd):}

\small
\medskip

\textbf{System message.}

You are an expert competitive programmer: write clean, efficient, and correct code
that solves the problem statement within its constraints.

\medskip
\textbf{User message.}

Here is the problem statement with time limit: \texttt{<time limit>} and memory limit: \texttt{<memory limit>}:
 
\texttt{<problem statement>}

Please generate an optimized and correct solution in
\texttt{<programming language>}.
\medskip

\normalsize
\textbf{With editorial (w/GenEd and w/GoldEd).}

\small
\medskip
\textbf{System message.}

You are an expert competitive programmer: write clean, efficient, and correct code
that solves the problem statement within its constraints and strictly follows the
editorial instructions.

\medskip
\textbf{User message.}

Here is the problem statement with time limit: \texttt{<time limit>} and memory limit: \texttt{<memory limit>}:

\texttt{<problem statement>}

Editorial guidelines (follow these exactly):

\texttt{<editorial text>}

Please generate an optimized and correct solution in
\texttt{<programming language>}.

\end{tcolorbox}
\caption{
Code-generation prompts used in our evaluation.
The top prompt corresponds to direct problem-to-code generation (\textbf{w/oEd}),
while the bottom prompt conditions code generation on an editorial
(\textbf{w/GenEd} and \textbf{w/GoldEd}).
}
\label{fig:prompt-codegen}
\end{figure*}
\begin{figure*}
    
\begin{tcolorbox}[ title={Editorial-generation prompt}]

\small

\textbf{System message.}

You are an expert competitive programmer. Your task is to write a clear and detailed
editorial for the given problem. Begin by restating the problem in your own words.
Then, walk through the key ideas and algorithmic approach needed to solve it.
Explain each step carefully — think step by step, and highlight any non-trivial
reasoning. Avoid writing actual code. If helpful, include brief pseudocode to
illustrate the logic. Focus on understanding and explanation. Your goal is to teach
the reader how to approach and solve the problem, not just provide an answer.
Make sure the solution respects the given time and memory limits. Finally, explain
why your solution is correct and how it handles all possible cases.

\medskip
\textbf{User message.}

Here is a competitive programming problem with time limit: \texttt{<time limit>} and memory limit: \texttt{<memory limit>}:

\texttt{<problem statement>}

Please generate a detailed editorial.

\end{tcolorbox}
\caption{
Editorial-generation prompt used to elicit model-written editorials.
Models are instructed to explain the algorithmic solution in natural language
without producing code.
}
\label{fig:prompt-editorial}

\end{figure*}

\section{Model Cards and Inference Setup}
\label{app:model-cards}

We evaluate a total of 19 language models spanning proprietary and open-weight systems.
Unless otherwise noted, all models are evaluated in a strictly single-shot regime:
each model produces at most one editorial (when applicable) and one code submission
per problem and setting. We do not perform test-time sampling, majority voting,
or iterative refinement; reported pass@1 results therefore reflect a single shot
completion per model, problem, and condition.

Proprietary models are queried via their official APIs, using provider-recommended
or default inference settings wherever possible. Open-weight models are served either
locally (via \texttt{vLLM} or HuggingFace Transformers) or through hosted providers
(OpenRouter, DeepInfra). To ensure comparability across heterogeneous backends,
we favor deterministic decoding (e.g., greedy decoding or zero temperature) unless
the model authors explicitly recommend otherwise.

For some proprietary APIs, exact maximum token limits are not publicly specified or
are managed dynamically by the provider; in these cases, the corresponding entries
are marked as ``--'' in Table~\ref{tab:model-suite}. Throughout the table, token limits
are reported in thousands (k = 1{,}000) for compactness.

Table~\ref{tab:model-suite} summarizes the exact model identifiers, serving backends,
public HuggingFace endpoints (when applicable), citations, token limits, and inference
configurations used in our experiments.
\begin{table*}
\centering
\small
\setlength{\tabcolsep}{3pt}
\begin{tabular}{>{\centering\arraybackslash}p{2.3cm}
                >{\centering\arraybackslash}p{2cm}
                >{\centering\arraybackslash}p{3.7cm}
                >{\centering\arraybackslash}p{2.2cm}
                >{\centering\arraybackslash}p{1.3cm}
                >{\centering\arraybackslash}p{3.0cm}}
\toprule
Short name(s) & Backend & HuggingFace endpoint(s) & Citation(s) & Max tokens & Configurations \\
\midrule
\multicolumn{6}{c}{Closed-source models} \\
\midrule
GPT-5 (2025-08-07), O3 (2025-04-16) &
OpenAI API &
-- &
\cite{openai2025gpt5,openai2025competitiveprogramminglargereasoning} &
-- &
GPT-5 uses high-effort reasoning, O3 uses medium effort; other parameters at provider defaults. \\

\addlinespace[0.3em]

Gemini-2.5-Pro, Gemini-2.5-Flash &
Google Gemini API &
-- &
\cite{comanici2025gemini25pushingfrontier} &
-- &
Deterministic decoding with \texttt{temperature = 0}; \texttt{thinking\_budget = -1}. \\

\addlinespace[0.3em]

GPT-4.1 (2025-04-14), GPT-4o (2024-08-06) &
OpenAI API &
-- &
\cite{openai2025gpt41,openai2024gpt4ocard} &
-- &
Provider default decoding and safety settings. \\

\addlinespace[0.3em]

Claude-Opus-4, Claude-Sonnet-4 &
Anthropic API &
-- &
\cite{anthropic2025claude4systemcard} &
32k &
\texttt{thinking\_budget = 22k}; other parameters at provider defaults. \\

\midrule
\multicolumn{6}{c}{Open-weight models} \\
\midrule
GPT-OSS-120B, GPT-OSS-20B &
OpenRouter (120B); local vLLM (20B) &
\href{https://huggingface.co/gpt-oss/gpt-oss-120b}{gpt-oss/gpt-oss-120b}; \href{https://huggingface.co/gpt-oss/gpt-oss-20b}{gpt-oss/gpt-oss-20b} &
-- &
65k &
\texttt{temperature = 0.0}; \texttt{reasoning\_effort = "medium"}. \\

\addlinespace[0.3em]

DeepSeek-R1, DeepSeek-V3 &
DeepSeek official API &
\href{https://huggingface.co/deepseek-ai/DeepSeek-R1}{deepseek-ai/DeepSeek-R1}; \href{https://huggingface.co/deepseek-ai/DeepSeek-V3}{deepseek-ai/DeepSeek-V3} &
\cite{deepseekai2025deepseekr1incentivizingreasoningcapability,deepseekai2025deepseekv3technicalreport} &
-- &
Recommended reasoning (R1) and base (V3) endpoints with provider defaults. \\

\addlinespace[0.3em]

Qwen3-Coder-480B-A35B; 
Kimi-K2; Llama-3.1-405B; Llama-3.3-70B; Gemma-3-27B-it &
DeepInfra API &
\href{https://huggingface.co/Qwen/Qwen3-Coder-480B-A35B-Instruct}{Qwen/Qwen3-Coder-480B-A35B-Instruct}; \href{https://huggingface.co/moonshotai/Kimi-K2-Instruct}{moonshotai/Kimi-K2-Instruct}; \href{https://huggingface.co/meta-llama/Meta-Llama-3.1-405B-Instruct}{meta-llama/Meta-Llama-3.1-405B-Instruct}; \href{https://huggingface.co/meta-llama/Llama-3.3-70B-Instruct}{meta-llama/Llama-3.3-70B-Instruct}; \href{https://huggingface.co/google/gemma-3-27b-it}{google/gemma-3-27b-it} &
\cite{yang2025qwen3technicalreport,kimiteam2025kimik2openagentic,grattafiori2024llama3herdmodels,gemmateam2025gemma3technicalreport} &
65k &
Shared config; \texttt{temperature = 0.0}; \texttt{do\_sample = False} (greedy). \\

\addlinespace[0.3em]

Qwen3-8B &
Local (Transformers) &
\href{https://huggingface.co/Qwen/Qwen3-8B}{Qwen/Qwen3-8B} &
\cite{yang2025qwen3technicalreport} &
32k &
Transformers defaults; deterministic decoding. \\

\addlinespace[0.3em]
\addlinespace[0.3em]

OlympicCoder-7B &
Local &
\href{https://huggingface.co/open-r1/OlympicCoder-7B}{open-r1/OlympicCoder-7B} &
\cite{penedo2025openr1update3} &
32k &
Authors' recommended settings: \texttt{do\_sample = True}, \texttt{top\_k = 50}, \texttt{top\_p = 0.95}, \texttt{temperature = 0.7}. \\
\bottomrule
\end{tabular}
\caption{
Model cards and inference configuration for all systems in
Table~\ref{tab:model-performance}. Token limits are reported
in thousands (k = 1{,}000). All models are evaluated with a single
completion per problem and editorial setting.
}
\label{tab:model-suite}
\end{table*}

\section{Python vs. C++ Performance}
\label{appendix:python-vs-cpp}

Although competitive programming is overwhelmingly conducted in \textbf{C++} due to its
performance and memory guarantees, we additionally evaluate a subset of models in
\textbf{Python} to assess language sensitivity.

\begin{figure}[t]
  \centering
  \includegraphics[width=\linewidth]{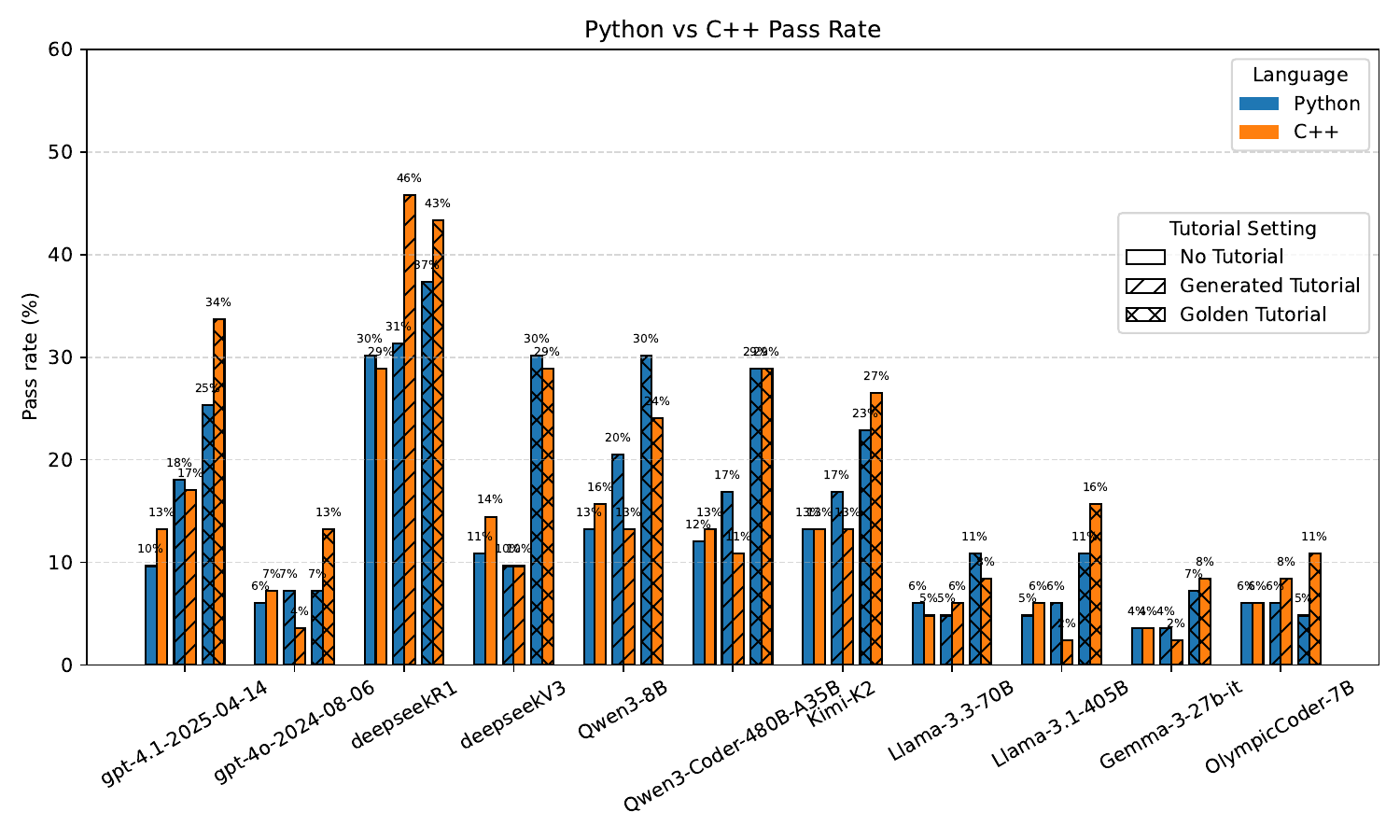}
  \caption{Pass@1 comparison between Python and C++ across editorial settings.}
  \label{fig:python-vs-cpp}
\end{figure}

Figure~\ref{fig:python-vs-cpp} compares pass@1 for Python and C++ across all 83 problems
under the three editorial settings (w/oEd, w/GenEd, w/GoldEd). Across nearly all models and
settings, \textbf{C++ consistently outperforms Python}.

This pattern is expected in ICPC-style problems, which frequently rely on tight constant
factors, low-level data structures, and explicit memory control. Python implementations
often incur TLE or excessive overhead even when the underlying algorithm is correct.
Accordingly, we treat C++ as the primary evaluation language and report Python results
only as supplementary analysis.

\section{Per-Contest Virtual Rank Percentiles}
\label{appendix:per-contest-rank}

\begin{figure*}[t]
\centering
\begin{subfigure}[t]{0.49\textwidth}
  \centering
  \includegraphics[width=\linewidth]{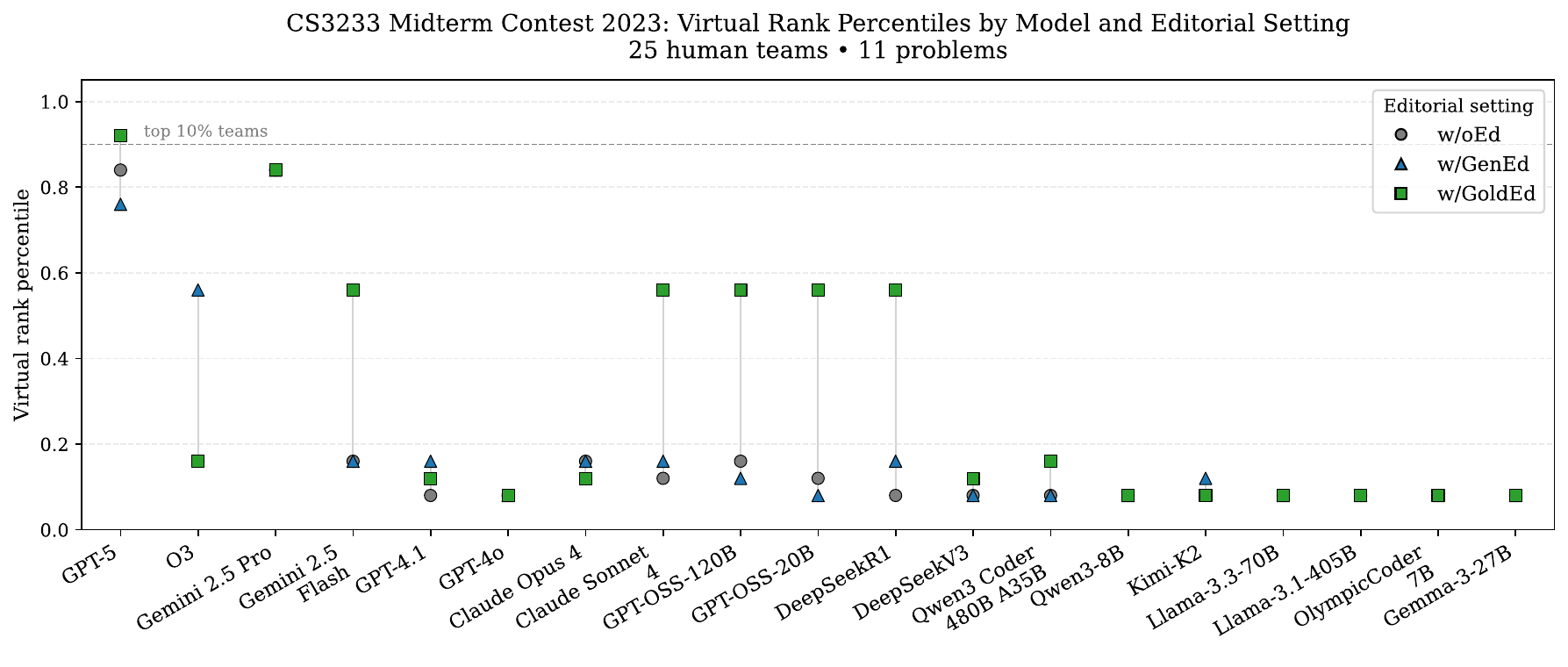}
  \caption{CS3233 Midterm 2023 (25 teams, 11 problems).}
  \label{fig:rank-cs3233-2023}
\end{subfigure}
\hfill
\begin{subfigure}[t]{0.49\textwidth}
  \centering
  \includegraphics[width=\linewidth]{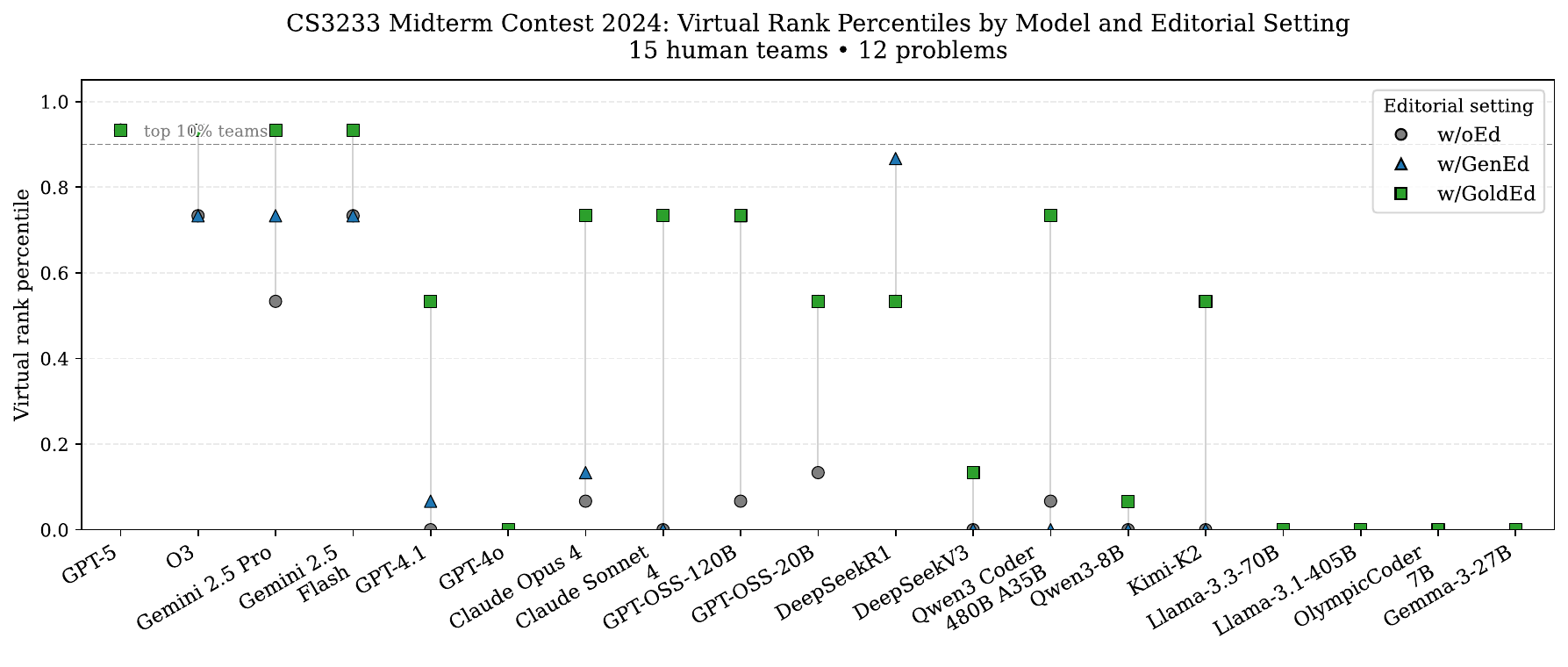}
  \caption{CS3233 Midterm 2024 (15 teams, 12 problems).}
  \label{fig:rank-cs3233-2024}
\end{subfigure}
\caption{Per-contest virtual rank percentiles by model and editorial setting (CS3233 midterms).}
\label{fig:rank-cs3233-2023-2024}
\end{figure*}

\begin{figure*}[t]
\centering
\begin{subfigure}[t]{0.49\textwidth}
  \centering
  \includegraphics[width=\linewidth]{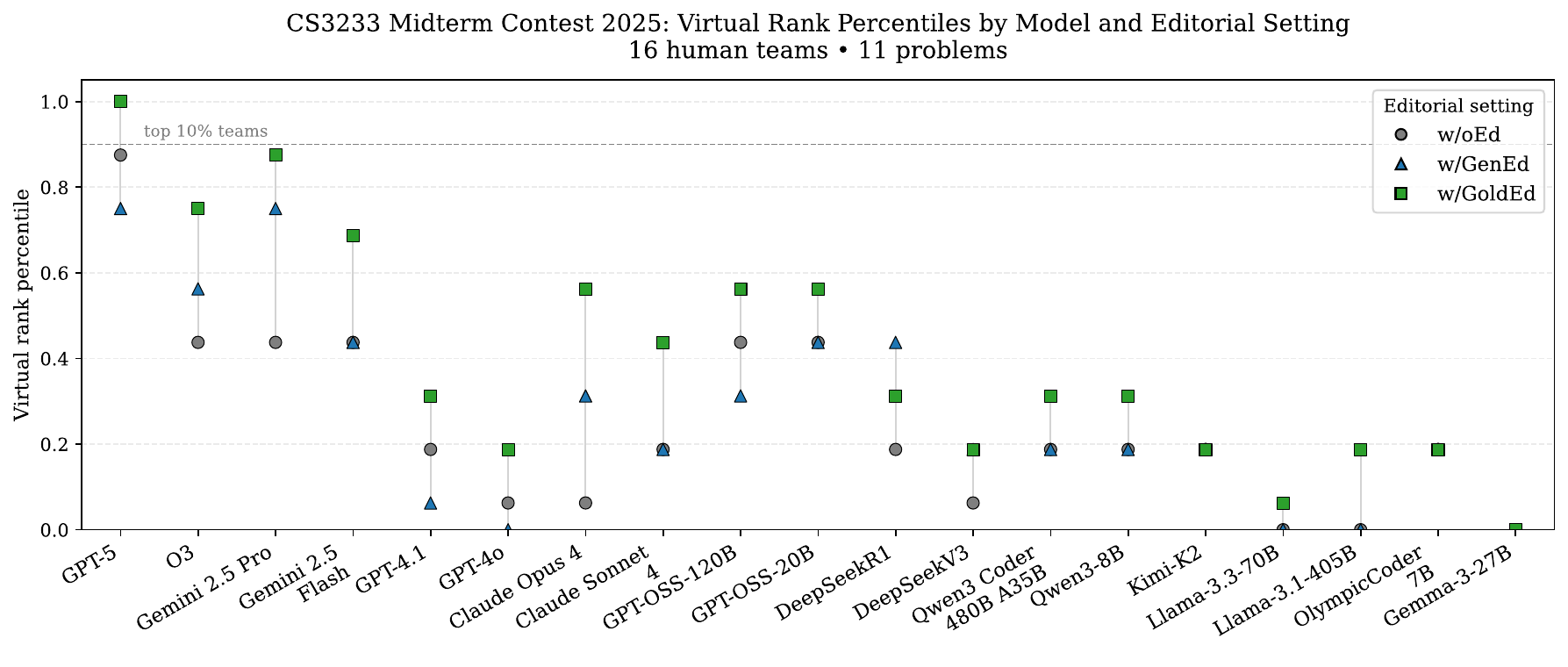}
  \caption{CS3233 Midterm 2025 (16 teams, 11 problems).}
  \label{fig:rank-cs3233-2025}
\end{subfigure}
\hfill
\begin{subfigure}[t]{0.49\textwidth}
  \centering
  \includegraphics[width=\linewidth]{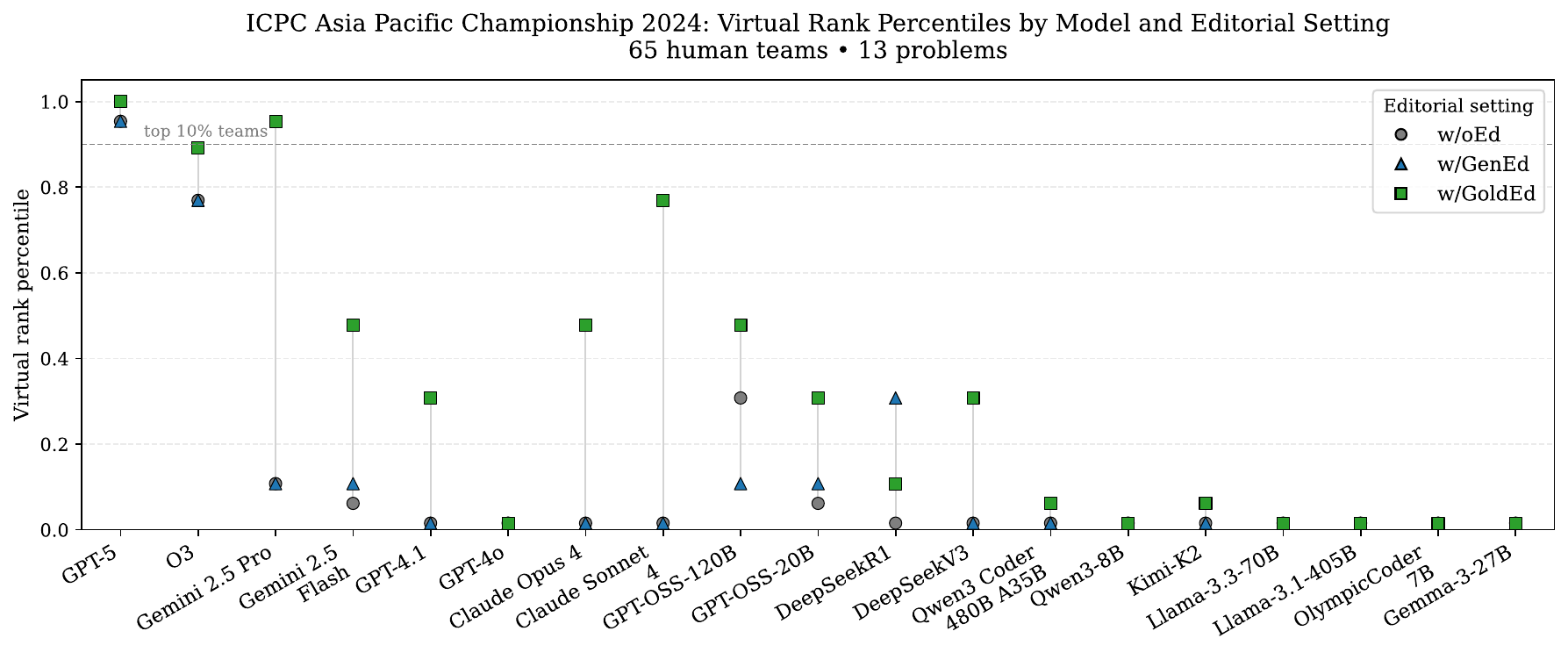}
  \caption{ICPC APAC 2024 (65 teams, 13 problems).}
  \label{fig:rank-apac-2024}
\end{subfigure}
\caption{Per-contest virtual rank percentiles (CS3233 and ICPC APAC).}
\label{fig:rank-cs3233-2025-apac-2024}
\end{figure*}

\begin{figure*}[t]
\centering
\begin{subfigure}[t]{0.49\textwidth}
  \centering
  \includegraphics[width=\linewidth]{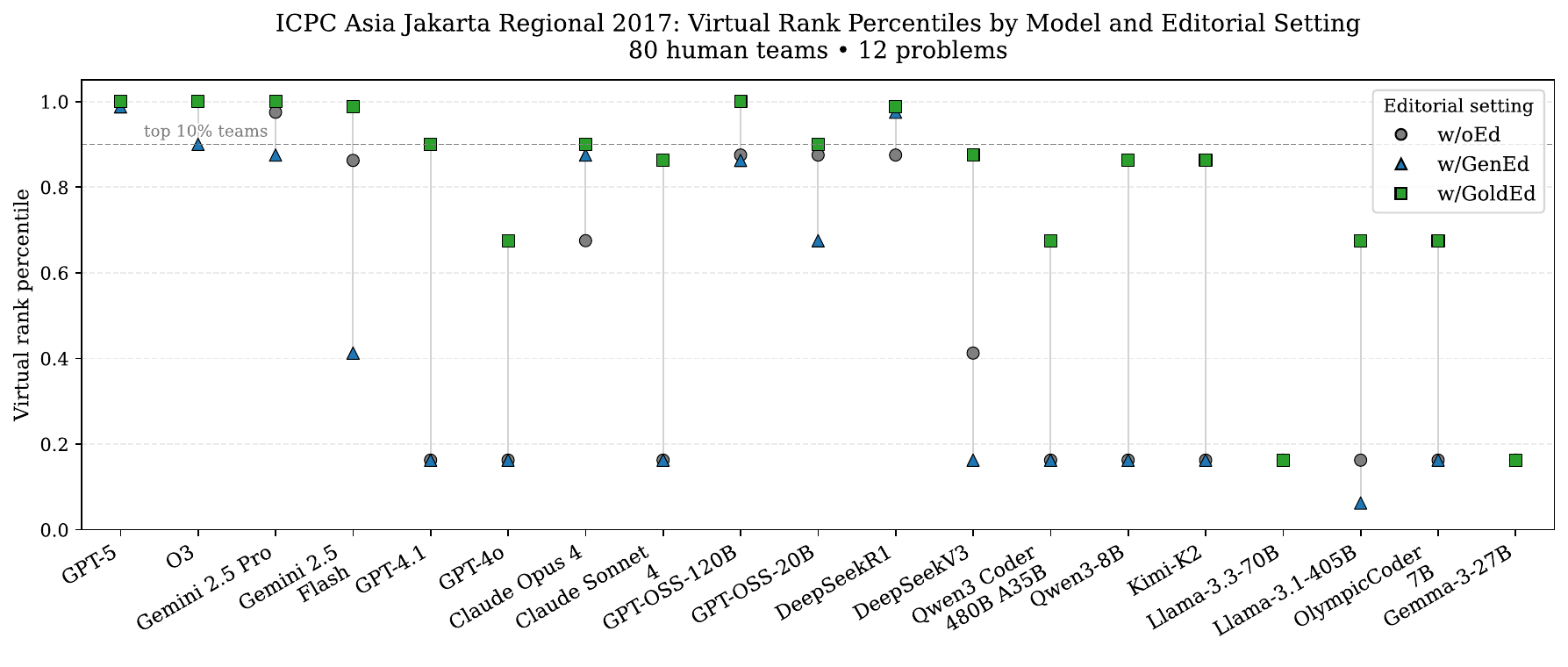}
  \caption{ICPC Jakarta 2017 (80 teams, 12 problems).}
  \label{fig:rank-jakarta-2017}
\end{subfigure}
\hfill
\begin{subfigure}[t]{0.49\textwidth}
  \centering
  \includegraphics[width=\linewidth]{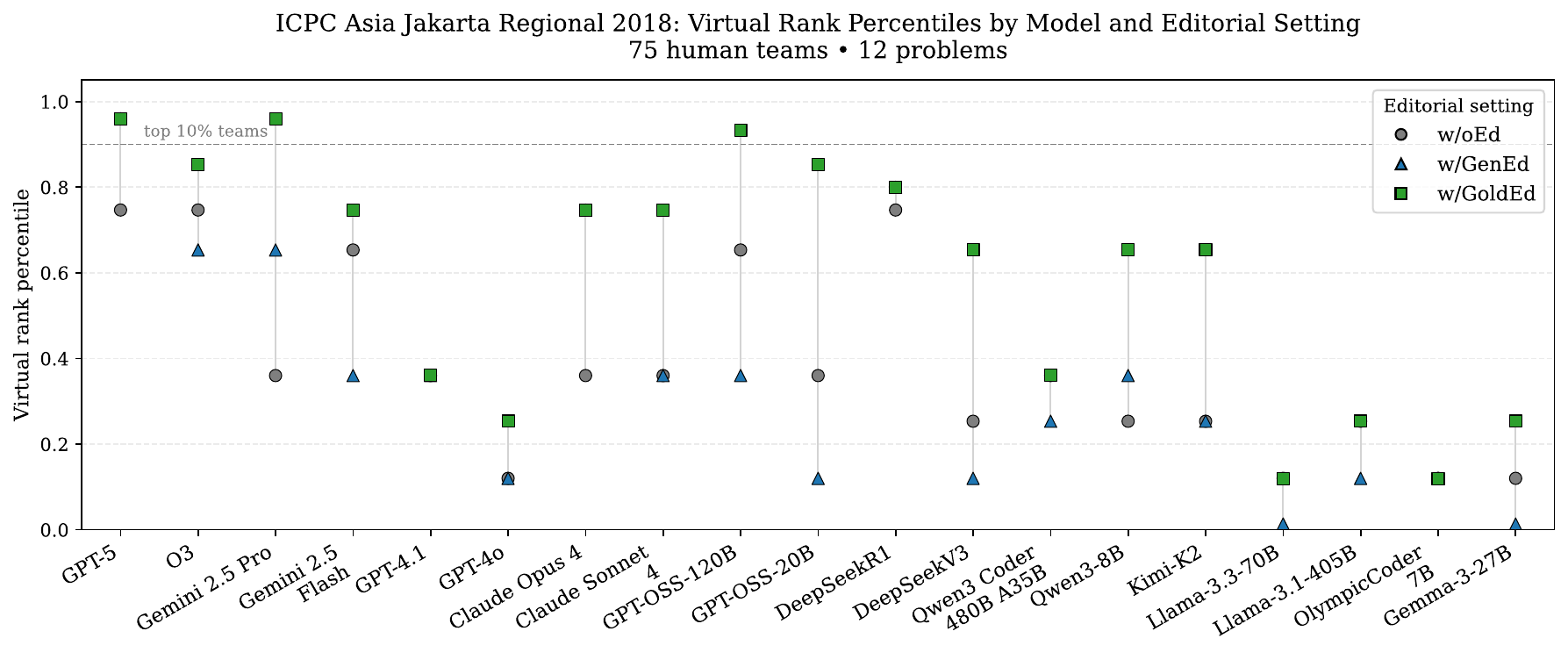}
  \caption{ICPC Jakarta 2018 (75 teams, 12 problems).}
  \label{fig:rank-jakarta-2018}
\end{subfigure}
\caption{Per-contest virtual rank percentiles (ICPC Jakarta regionals).}
\label{fig:rank-jakarta-2017-2018}
\end{figure*}

\begin{figure}[t]
\centering
\includegraphics[width=\linewidth]{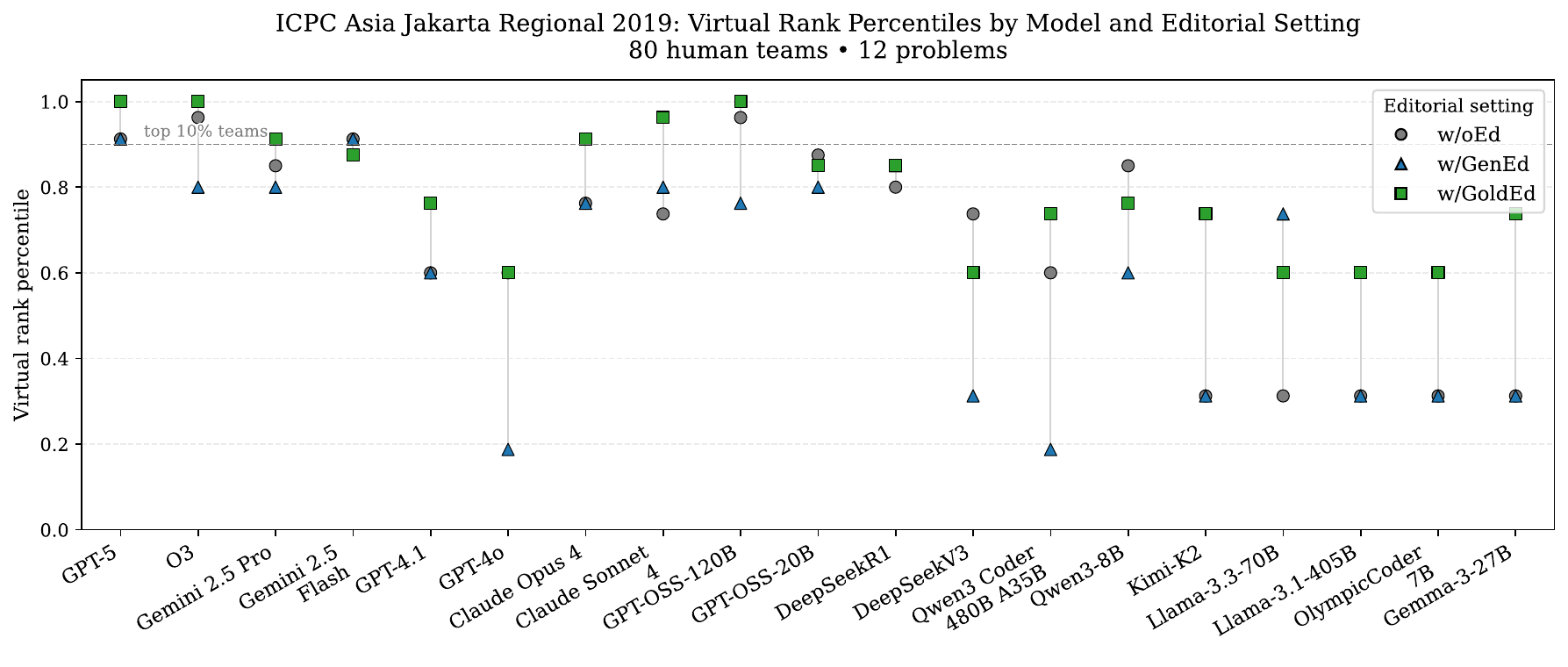}
\caption{ICPC Jakarta 2019: Virtual rank percentiles by model and editorial setting (80 teams, 12 problems).}
\label{fig:rank-jakarta-2019}
\end{figure}

Figures~\ref{fig:rank-cs3233-2023-2024}--\ref{fig:rank-jakarta-2019} report virtual rank percentiles separately for each contest.
Absolute rank percentiles vary across contests due to differences in the problem set, number of teams, so percentiles should be interpreted relative to each contest's field.
Despite this heterogeneity, the qualitative pattern is consistent: w/GoldEd yields the largest and most reliable upward rank shifts relative to w/oEd, w/GenEd produces smaller and higher-variance changes, and only a small subset of models attains high human-relative rank in any single contest.

\section{Detailed Failure Analysis by Error Type}
\label{appendix:failure-dist}

\begin{figure*}[t]
  \centering
  \includegraphics[width=\columnwidth]{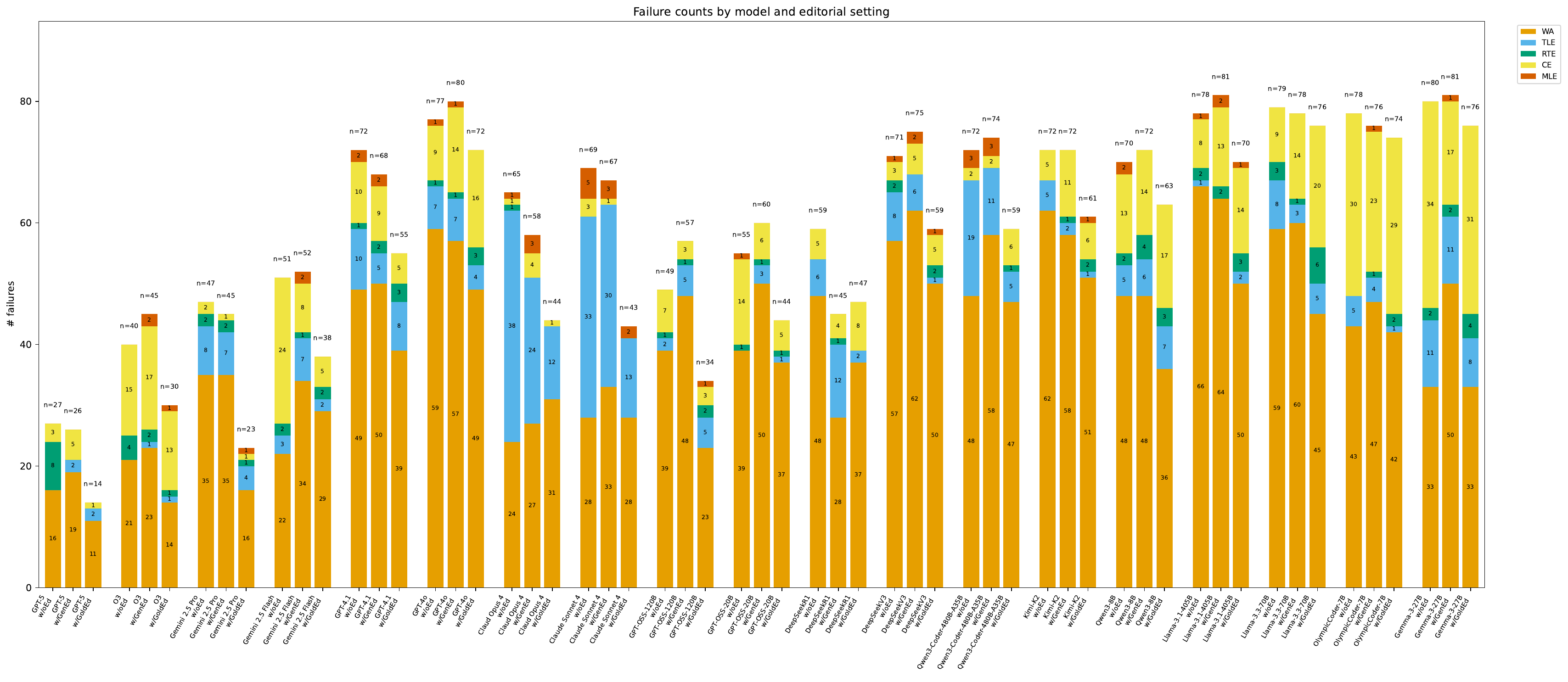}
  \caption{
  Failure counts by model and editorial setting. For each (model, setting) pair we show the number of failed submissions decomposed into Wrong Answer (WA), Time Limit Exceeded (TLE), Runtime Error (RTE), Compile Error (CE), and Memory Limit Exceeded (MLE). The label above each bar indicates the total number of failures $n$ in that condition.}
  \label{fig:failure-dist}
\end{figure*}

\paragraph{Overall failure mix.}
Across all models and settings, Wrong Answer (WA) is the single largest failure type in Figure~\ref{fig:failure-dist}. For almost every bar, the WA segment is the biggest component, which indicates that many failed submissions do compile and run but implement an incorrect or incomplete algorithm. Compile Error (CE) is also substantial: for several models, especially \textsc{O3}, \textsc{Gemini~2.5 Flash}, \textsc{Qwen3-8B}, and \textsc{OlympicCoder-7B}, CE is the second largest or even co-dominant mode. These CEs include both ordinary C++ syntax and type errors and the no-code behaviors described below. Time Limit Exceeded (TLE) and Runtime Error (RTE) account for a smaller share of failures, while Memory Limit Exceeded (MLE) is rare.

\paragraph{Effect of editorials on failure types.}
Changing the editorial setting affects both how often models fail and how they fail. Moving from \textbf{w/oEd} to \textbf{w/GenEd} typically shifts some mass from CE and RTE into WA: more runs produce code that compiles and runs, but the resulting programs still fail on hidden tests. In the \textbf{w/GoldEd} setting, total failure counts drop noticeably for most models and WA segments shrink, which shows that correct human editorials remove many misplanned solutions. However, CE remains non-trivial, especially for models that already struggle with compilation such as \textsc{O3} and \textsc{Gemini~2.5 Flash}. For these systems, even gold editorials do not reliably prevent syntax errors, truncated outputs, or off-spec completions.

\paragraph{Reasoning-limited versus implementation-limited models.}
The per-model stacks in Figure~\ref{fig:failure-dist} also reveal systematic differences between model families. Many open-weight coders, such as GPT-OSS-20B/120B, Qwen3-8B, Kimi-K2, Llama-3.1/3.3, and Gemma-3-27B, have failures where WA is the largest segment but CE is also a sizable fraction in all three settings. These models usually manage to reach compilation, yet a large share of runs still encode an incorrect algorithm or miss important edge cases, and a non-trivial number fail already at compile time. In contrast, the strongest closed models such as GPT-5, O3, and Gemini~2.5 Pro have relatively small CE segments and a mix of WA and TLE, which suggests that their main residual bottlenecks are logical errors and efficiency rather than basic syntax. \textsc{O3} and \textsc{Gemini~2.5 Flash} are notable outliers: their bars contain unusually large CE blocks in every setting, indicating that unstable or incomplete code generation is itself a major source of failure.

\paragraph{No-code and no-output pathologies.}
Among all $83$ problems, $20$ models, and $3$ editorial settings ($83 \times 19 \times 3 = 4{,}731$ model--problem--setting runs), we observe a small but qualitatively distinct set of failures where the model never really submits a candidate program. Using the raw logs, we identify $91$ completions that are labeled \textsc{Compile Error} but contain no compilable C++ at all. In these cases the model either explicitly declines to solve the problem (for example, ``The problem is very hard. A correct solution that works within the limits is not feasible to provide in this format.''), produces an editorial-style explanation that stops just before the ``Reference implementation'' section, or outputs a full solution in another language such as Python while the prompt requests C++. These no-code CEs are heavily concentrated in \textsc{O3} and \textsc{Gemini~2.5 Flash}, with smaller numbers for \textsc{GPT-OSS-20B}, \textsc{GPT-OSS-120B}, GPT-5, \textsc{Qwen3-8B}, and \textsc{OlympicCoder-7B}, and they occur disproportionately on the hardest contests (ICPC APAC 2024 and CS3233 midterms).

We also find $12$ runs that compile but terminate with a runtime error without producing any output on our judge, effectively giving no answer at execution time. These cases arise mainly for GPT-5, \textsc{O3}, \textsc{Gemini~2.5 Flash}, and \textsc{OlympicCoder-7B} on the most challenging problems. Qualitatively, they behave similarly to the no-code CEs: the model never produces a testable solution, as opposed to producing a plausible but wrong program. In the main figures we conservatively count such behaviors under the standard \textsc{CE} or \textsc{RTE} buckets, but they highlight that a small share of failures correspond to genuine non-attempts (or off-spec attempts) rather than buggy implementations. Models such as \textsc{Qwen3-8B} and \textsc{Qwen3-Coder-30B-A3B} occasionally exhibit related behavior by exhausting their token budgets on long ``thinking'' traces or hallucinated analyses without ever emitting code, which our pass@1 metric naturally penalizes.

\section{Editorial length statistics}
\label{appendix:editorial-length}

To quantify how model-generated editorials differ from human-written gold editorials in level of detail, 
we compare their word counts across all problems and models.

\begin{figure}
    \centering
    \includegraphics[width=0.9\columnwidth]{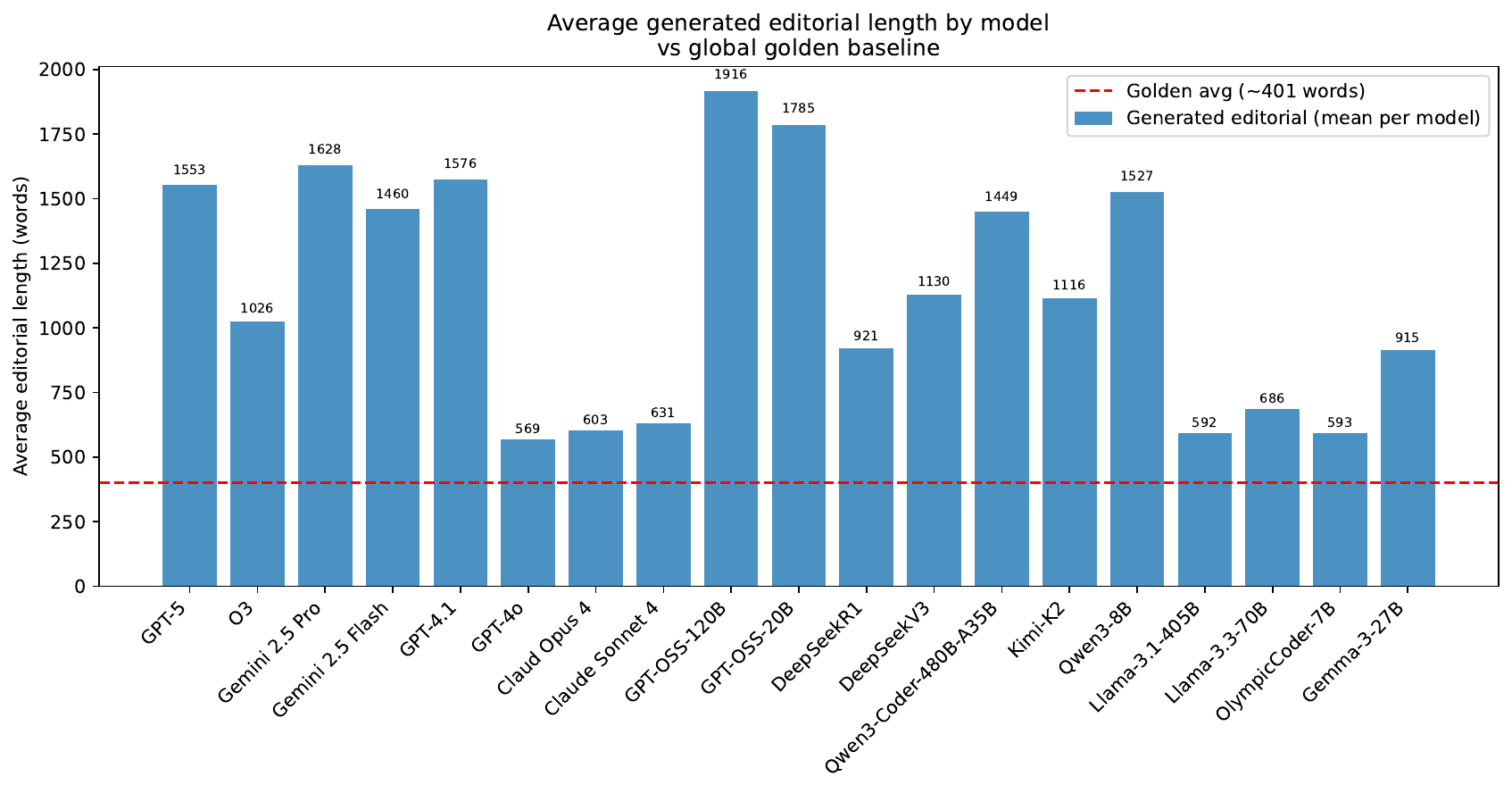}
    \caption{
    Word-count comparison between human-written gold editorials and model-generated editorials, 
    aggregated over all problems. For each model we plot the distribution of word counts for its 
    w/GenEd editorials and the corresponding gold editorials. Across all models, model-generated 
    editorials are consistently longer than the gold ones, indicating that they tend to provide 
    more verbose, step-by-step explanations.
    }
    \label{fig:editorial-length}
\end{figure}

Across all 19 models, model-generated editorials are systematically longer than the corresponding 
gold editorials. In other words, when models are asked to ``write an editorial'', they typically 
produce more expansive and didactic explanations than the contest editorials themselves, even when 
both describe essentially the same algorithmic plan. This quantitative pattern complements our 
qualitative observation that model editorials often trade concision for more step-by-step reasoning.




\newcommand{\xtypeExp}{Explicit}
\newcommand{\xtypeImp}{Implicit}
\newcommand{\xtypeNA}{--}

\section{CS3233 2025 Midterm Contest Full Annotations}
\label{app:cs3233-qual-annotations}

This appendix provides the complete human annotations for the CS3233 2025 Midterm qualitative case study.
Annotations follow the rubric in Appendix~\ref{appendix:annotation-rubric} and are applied to model-generated
editorials from the generated-editorial setting.
Table~\ref{tab:cs3233-pu-alg} summarizes problem understanding, including wrong or missing crucial details,
misinformation type, misleading severity, and annotator comments.
Tables~\ref{tab:cs3233-alg-tags} and its continuation report the algorithmic paradigms assigned to each editorial
and the corresponding freeform summaries for both the model and gold editorials.
Table~\ref{tab:cs3233-alg-cor} reports editorial-level algorithmic correctness, diagnostic failure categories,
and the final judge verdicts of the generated code, enabling direct comparison between editorial reasoning quality
and downstream execution outcomes.


\begin{table*}
\centering
\scriptsize
\setlength{\tabcolsep}{3pt}
\renewcommand{\arraystretch}{1.12}
\begin{tabularx}{\textwidth}{@{}l l c c c c c c p{5.5cm}@{}}

\toprule
Problem & Model &
PU-W & Misinformation Type &
PU-M & Misinformation Type &
PU-X & PU-D &
PU-Comments \\
\midrule

\rowcolor{LightGray}
\multicolumn{9}{@{}l}{\textsc{arts and computing students}} \\
DeepSeek R1 & & \puNo & \xtypeNA & \puNo & \xtypeNA & \sevMinor & 2 &
The understanding becomes wrong under ``Key Insights'', it assumes the student can be moved rather than shifted \\
GPT 5 & & \puNo & \xtypeNA & \puNo & \xtypeNA & \sevNone & 2 &
\xtypeNA \\

\rowcolor{LightGray}
\multicolumn{9}{@{}l}{\textsc{brilliance of wings}} \\
DeepSeek R1 & & \puNo & \xtypeNA & \puNo & \xtypeNA & \sevNone & 1 &
\xtypeNA \\
GPT 5 & & \puNo & \xtypeNA & \puNo & \xtypeNA & \sevNone & 1 &
\xtypeNA \\

\rowcolor{LightGray}
\multicolumn{9}{@{}l}{\textsc{chained maimai slides}} \\
DeepSeek R1 & & \puYes & \xtypeExp & \puYes & \xtypeExp & \sevMajor & 2 &
\xtypeNA \\
GPT 5 & & \puNo & \xtypeNA & \puNo & \xtypeNA & \sevNone & 2 &
\xtypeNA \\

\rowcolor{LightGray}
\multicolumn{9}{@{}l}{\textsc{dependency flood}} \\
DeepSeek R1 & & \puNo & \xtypeNA & \puNo & \xtypeNA & \sevMinor & 2 &
\xtypeNA \\
GPT 5 & & \puNo & \xtypeNA & \puNo & \xtypeNA & \sevNone & 2 &
\xtypeNA \\

\rowcolor{LightGray}
\multicolumn{9}{@{}l}{\textsc{easygoing workplace}} \\
DeepSeek R1 & & \puNo & \xtypeNA & \puNo & \xtypeNA & \sevNone & 0 &
missing the $B_i < i$, i.e.\ superior is always the smaller number \\
GPT 5 & & \puNo & \xtypeNA & \puNo & \xtypeNA & \sevNone & 0 &
missing the $B_i < i$ written in the I/O section, i.e.\ superior is always the smaller number \\

\rowcolor{LightGray}
\multicolumn{9}{@{}l}{\textsc{ficketts conjecture for polyominoes}} \\
DeepSeek R1 & & \puNo & \xtypeNA & \puNo & \xtypeNA & \sevMinor & 2 &
it says: ``max fraction of overlapping border cells between $A$ and $A_1$''. it should be: ``max fraction of border cells between $A$ and $A_1$, within overlapping cells'' \\
GPT 5 & & \puNo & \xtypeNA & \puNo & \xtypeNA & \sevNone & 2 &
\xtypeNA \\

\rowcolor{LightGray}
\multicolumn{9}{@{}l}{\textsc{georgette me georgette you}} \\
DeepSeek R1 & & \puNo & \xtypeNA & \puNo & \xtypeNA & \sevNone & 0 & \\
GPT 5 & & \puNo & \xtypeNA & \puNo & \xtypeNA & \sevNone & 0 &
\xtypeNA \\

\rowcolor{LightGray}
\multicolumn{9}{@{}l}{\textsc{hungry piplups}} \\
DeepSeek R1 & & \puNo & \xtypeNA & \puNo & \xtypeNA & \sevNone & 2 &
\xtypeNA \\
GPT 5 & & \puNo & \xtypeNA & \puNo & \xtypeNA & \sevNone & 2 &
\xtypeNA \\

\rowcolor{LightGray}
\multicolumn{9}{@{}l}{\textsc{imperfection}} \\
DeepSeek R1 & & \puNo & \xtypeNA & \puNo & \xtypeNA & \sevNone & 0 &
\xtypeNA \\
GPT 5 & & \puNo & \xtypeNA & \puNo & \xtypeNA & \sevNone & 0 &
\xtypeNA \\

\rowcolor{LightGray}
\multicolumn{9}{@{}l}{\textsc{jaunt through the garden}} \\
DeepSeek R1 & & \puNo & \xtypeNA & \puNo & \xtypeNA & \sevNone & 0 &
\xtypeNA \\
GPT 5 & & \puNo & \xtypeNA & \puNo & \xtypeNA & \sevNone & 0 &
\xtypeNA \\

\rowcolor{LightGray}
\multicolumn{9}{@{}l}{\textsc{keep the ordering}} \\
DeepSeek R1 & & \puNo & \xtypeNA & \puNo & \xtypeNA & \sevNone & 0 &
\xtypeNA \\
GPT 5 & & \puNo & \xtypeNA & \puNo & \xtypeNA & \sevNone & 0 &
\xtypeNA \\

\bottomrule
\end{tabularx}
\caption{
Problem understanding annotations for the CS3233 2025 Midterm case study.
PU W and PU M indicate the presence of wrong or missing crucial problem details, with misinformation type recorded
when applicable.
PU X records misleading severity and PU D is the annotator-rated understanding difficulty.
PU Comments provides brief annotator notes.
}

\label{tab:cs3233-pu-alg}
\end{table*}

\begin{table*}
\centering
\tiny
\setlength{\tabcolsep}{3pt}
\renewcommand{\arraystretch}{1.12}
\begin{tabularx}{\textwidth}{@{}l l p{1cm} p{0.75cm} p{3.5cm} p{0.75cm} p{1cm} p{3.5cm} p{2.2cm}@{}}
\toprule
Problem & Model &
ALG-TAG &
ALG-TAG-OTHER &
ALG-FREE &
Gold-ALG-TAG &
Gold-ALG-TAG-OTHER &
Gold-ALG-FREE &
ALG-Comments \\
\midrule

\rowcolor{LightGray}
\multicolumn{9}{@{}l}{\textsc{arts and computing students}} \\
DeepSeek R1 & &
Greedy & &
Count the A and C for general feasability. Scan left-to-right and right-to-left tracking the count difference. Constructing by rearranging the whole seating with AC pair first, followed by remaining As and empty seats. &
DP, Greedy & &
DP with possibility to either pair with next student or no pairing. Then check whehter the possible arrangement has enough seating. Alternatively, split the ACCA arrangement, the remaining can be picked greedily for any adjacent AC and CA. The possible arrangement has the minimum number of seatings needed. &
\xtypeNA \\
GPT 5 & &
Greedy & &
Count A, C and '.' for feasibility. Remove '.', greedily pair the different consecutive chars from left-to-right. Re-iterate to pair the unpaired chars: A with A, unfeasible for C. & & & &
The generated editorial does not consider cases where picking arbitrarily one of 2 forms (AC or CA) would end up in a longer arrangement \\

\rowcolor{LightGray}
\multicolumn{9}{@{}l}{\textsc{brilliance of wings}} \\
DeepSeek R1 & &
Other &
Set Manipulation (inclusion-exclusion) &
Total number of edges that are present in the first tree but not in second tree &
Other &
Graph Theory (Tree), Set Manipulation (inclusion-exclusion) &
Each step require edge deletion from tree 1 and edge insertion from tree 2. Hence, minimum step is total edge - shared edges. &
\xtypeNA \\
GPT 5 & &
Other &
Set Manipulation (inclusion-exclusion) &
Create hash set or balanced BST for tree-2, then iterate edges of tree-1 and check if it is in the set. Output (N-1) - edge\_in\_set. & & & &
\xtypeNA \\

\rowcolor{LightGray}
\multicolumn{9}{@{}l}{\textsc{chained maimai slides}} \\
DeepSeek R1 & &
DP, Two-Pointer / Sliding Window & &
\xtypeNA &
Other &
MCMF &
\xtypeNA &
\xtypeNA \\
GPT 5 & &
Other &
MCMF &
\xtypeNA & & &
\xtypeNA &
\xtypeNA \\

\rowcolor{LightGray}
\multicolumn{9}{@{}l}{\textsc{dependency flood}} \\
DeepSeek R1 & &
DFS/BFS (Graph Traversal), Other &
Graph Theory &
Graph traversal for: (1) constructing prefix and suffix count, (2) updating accepted query. The observation shows that there is an upper bound of this update, K, so not all the queries are update-operation. &
DFS/BFS (Graph Traversal), Other &
Graph Theory &
Graph traversal for: (1) constructing prefix and suffix count, (2) updating accepted query. The observation shows that there is an upper bound of this update, K, so not all the queries are update-operation. &
LLM-editorial says DP but only the memoization part, no relation in the dimension;
surprisingly, the generated editorial is easier to grasp \\
GPT 5 & &
DFS/BFS (Graph Traversal), Other &
Graph Theory &
Graph traversal for: (1) constructing prefix count from root to particular node, (2) updating accepted query. & & & &
\xtypeNA \\

\rowcolor{LightGray}
\multicolumn{9}{@{}l}{\textsc{easygoing workplace}} \\
DeepSeek R1 & &
DFS/BFS (Graph Traversal), Greedy & &
Two traversals where the first one computes the range of feasible K for each subtree and also each node's feasibility to slack off. Second traversal to assign so that only K workers work, the answer construction follows this decision. &
Binary Search, Math/Number Theory & &
If K is feasible, there should exist an x where (x, ..., N, x-1, ..., 1) is the answer. x could be found by binary-search or iterative construction. &
got WA by following the golden editorial. \\
GPT 5 & &
DP, DFS/BFS (Graph Traversal), Other & &
DP + Disjoint Set Union, using graph traversal. & & & &
\xtypeNA \\

\rowcolor{LightGray}
\multicolumn{9}{@{}l}{\textsc{ficketts conjecture for polyominoes}} \\
DeepSeek R1 & &
Other &
Brute Force &
Try all possible rotations and tranlations of A1, then count by traversing each cell one-by-one. &
Geometry & &
Pre-compute the boundary cells, then consider all rotations and translations, i.e., all possible A1. When counting the value, use this precomputed coordinates, ensuring O(m+n) per A1. &
\xtypeNA \\
GPT 5 & &
Other &
Geometry &
Build row and coloumn intervals, then count the border cells via erosion of 3x3 blocks. Translations are then considered exhaustively for computing the best ratio. & & & &
\xtypeNA \\

\rowcolor{LightGray}
\multicolumn{9}{@{}l}{\textsc{georgette me georgette you}} \\
DeepSeek R1 & &
DP & &
Same with golden, but the DP memoization is referred as involutions. &
DP & &
The permutation P equals inverse permutation Q when all elements satisfy P\_i = k and P\_k = i for all (i, k). By counting the P=Q permutations using DP, then the answer is all permutations minus P=Q permutations. &
LLM-editorial uses 1-dimensional DP, like prefix-sum, but never mention this paradigm is DP. \\
GPT 5 & &
DP & &
Precompute the factorial and the involutions of 1-cycle and 2-cycles. Answer each query as factorial minus involutions. & & & &
\xtypeNA \\

\bottomrule
\end{tabularx}
\caption{
Algorithmic paradigm annotations for the CS3233 2025 Midterm case study.
ALG TAG and ALG TAG OTHER list the paradigms attributed to each model-generated editorial, and ALG FREE provides a
freeform summary of the described approach.
Gold columns report the corresponding gold editorial tags and summary.
ALG Comments records brief annotator notes on the algorithm description.
}

\label{tab:cs3233-alg-tags}
\end{table*}

\begin{table*}
\ContinuedFloat
\centering
\tiny
\setlength{\tabcolsep}{3pt}
\renewcommand{\arraystretch}{1.12}
\begin{tabularx}{\textwidth}{@{}l l p{1cm} p{0.75cm} p{3.5cm} p{0.75cm} p{1cm} p{3.5cm} p{2.2cm}@{}}
\toprule
Problem & Model &
ALG-TAG &
ALG-TAG-OTHER &
ALG-FREE &
Gold-ALG-TAG &
Gold-ALG-TAG-OTHER &
Gold-ALG-FREE &
ALG-Comments \\
\midrule

\rowcolor{LightGray}
\multicolumn{9}{@{}l}{\textsc{hungry piplups}} \\
DeepSeek R1 & &
Greedy & &
Simulating the problem statement, with small tweaks without any effect to the overall solution complexity &
Segment Tree, Binary Search, Greedy & &
Construct a segment tree of N, where each node stores value count of A\_i, exploiting the limits of 1 <= A\_i <= 500. Thus, each query (update and merging), only cost $O(\max_{A_i}\log N)$. &
Model is the expected naive or BF version \\
GPT 5 & &
Binary Search, Greedy, Segment Tree & &
Segment-tree like approach by managing max\_{A\_i} (500) ordered sets, each query is done by O($\log N$) operation. The output is counted greedy-ly. & & & &
\xtypeNA \\

\rowcolor{LightGray}
\multicolumn{9}{@{}l}{\textsc{imperfection}} \\
DeepSeek R1 & &
Segment Tree, Bit Manipulation, Other &
Fenwick Tree (Binary Indexed Tree) &
Precompute fixed cost of deleting all elements greater than any given x, and maintain variable cost to efficiently delete elements smaller than x. The answer is the minimum of fixed + variable cost for all x. &
Segment Tree, Bit Manipulation, Other &
Binary Indexed Tree (Fenwick Tree) &
Finding two algorithms to: (1) compute the sum of costs to delete K cheapest elements, and (2) remove the maximum element from consideration. Both can be solved using BIT or Segment Tree data structures. &
feels like different algo but actually similar \\
GPT 5 & &
Segment Tree, Bit Manipulation, Other &
Binary Indexed Tree \& Fenwick Tree &
Group by value to precompute cost to delete for a fixed value, iterate the unique values in increasing order, saving the best answer. BIT/Fenwick tree is required to maintain frequency and cost. & & & &
\xtypeNA \\

\rowcolor{LightGray}
\multicolumn{9}{@{}l}{\textsc{jaunt through the garden}} \\
DeepSeek R1 & &
DP & &
1-dimensional DP &
DP & &
2-dimensional DP &
The generated editorial suddenly changed to 1-dimensional DP while previously already mentioning the DP state is 2 dimensions. The gold editorial mentions an O(N) solution rather than the O(N$^2$) DP; this formula seems to be what the generated editorial tries to achieve. \\
GPT 5 & &
Math/Number Theory, DP & &
DP for precomputation of combinatorics, base-2 powers, and the inversion. Iterate in linear time to compute the answer based on the formula. & & & &
\xtypeNA \\

\rowcolor{LightGray}
\multicolumn{9}{@{}l}{\textsc{keep the ordering}} \\
DeepSeek R1 & &
Other, Binary Search &
Precompute (brute-force/BFS) &
Construct valid N, compute and check N$^2$, save in precomputation. The answer is found by binary searching the index in this precomputed list. &
Other, Binary Search &
Precompute (brute-force/DFS) &
Construct valid N, compute and check N$^2$, save in precomputation. The answer is found by binary searching the index in this precomputed list. &
\xtypeNA \\
GPT 5 & &
Other, Binary Search &
Precompute (brute-force/DFS) &
Construct valid N, compute and check N$^2$, save in precomputation. Answer is binary search of finding index. & & & &
\xtypeNA \\
\bottomrule
\end{tabularx}

\caption{Algorithmic paradigm annotations for the CS3233 2025 Midterm case study (continued).}
\end{table*}

\begin{table*}
\centering
\scriptsize
\setlength{\tabcolsep}{3pt}
\renewcommand{\arraystretch}{1.12}
\begin{tabularx}{\textwidth}{@{}l l c c X c X c@{}}
\toprule
Problem & Model &
ALG COR &
Correct type &
Why incorrect &
Severity &
Annotator comments &
Final verdict \\
\midrule

\rowcolor{LightGray}
\multicolumn{8}{@{}l}{\textsc{arts and computing students}} \\
DeepSeek R1 & &
\algwrong & \xtypeNA &
Wrong algorithm &
Completely wrong &
Shift $\neq$ Move &
WA \\
GPT 5 & &
\algwrong & \xtypeNA &
Wrong algorithm &
Major edits needed &
Need to reconsider a valid case of: ``A.\ AC\ AA\ CC\ AA'' &
WA \\

\rowcolor{LightGray}
\multicolumn{8}{@{}l}{\textsc{brilliance of wings}} \\
DeepSeek R1 & &
\algok & Same as golden &
\xtypeNA & \xtypeNA &
\xtypeNA &
PASS \\
GPT 5 & &
\algok & Same as golden &
\xtypeNA & \xtypeNA &
\xtypeNA &
PASS \\

\rowcolor{LightGray}
\multicolumn{8}{@{}l}{\textsc{chained maimai slides}} \\
DeepSeek R1 & &
\algwrong & \xtypeNA &
Wrong algorithm &
Completely wrong &
Hallucinated &
RTE \\
GPT 5 & &
\algok & Same as golden &
\xtypeNA & \xtypeNA &
Seems to have same core idea despite differing in the complexity analysis. &
PASS \\

\rowcolor{LightGray}
\multicolumn{8}{@{}l}{\textsc{dependency flood}} \\
DeepSeek R1 & &
\algok & Same as golden &
\xtypeNA & \xtypeNA & \xtypeNA\\
GPT 5 & &
\algwrong & \xtypeNA &
Wrong algorithm &
Minor edits needed &
The model only keeps track of one count from root to particular node but not from particular node to leaves. &
WA \\

\rowcolor{LightGray}
\multicolumn{8}{@{}l}{\textsc{easygoing workplace}} \\
DeepSeek R1 & &
\algok & Different from golden &
\xtypeNA & \xtypeNA &\xtypeNA &
TLE \\
GPT 5 & &
\algok & Different from golden &
\xtypeNA & \xtypeNA &
The solution does not rely on the fact that subordinate always have higher index. &
PASS \\

\rowcolor{LightGray}
\multicolumn{8}{@{}l}{\textsc{ficketts conjecture for polyominoes}} \\
DeepSeek R1 & &
\algwrong & \xtypeNA &
Suboptimal (Likely TLE or MLE), but correct algorithm &
Major edits needed &
should be TLE as it considers O(R$\cdot$C), rather than O(R+C), i.e.\ there can only be 2 border-cells per row or per column &
TLE \\
GPT 5 & &
\algok & Different from golden &
\xtypeNA & \xtypeNA &
Most parts are similar; rather than straightforward counting, the model proposes a count through slope arrays. &
RTE \\

\rowcolor{LightGray}
\multicolumn{8}{@{}l}{\textsc{georgette me georgette you}} \\
DeepSeek R1 & &
\algok & Same as golden &
\xtypeNA & \xtypeNA & &
PASS \\
GPT 5 & &
\algok & Same as golden &
\xtypeNA & \xtypeNA &
\xtypeNA &
PASS \\

\rowcolor{LightGray}
\multicolumn{8}{@{}l}{\textsc{hungry piplups}} \\
DeepSeek R1 & &
\algwrong & \xtypeNA &
Suboptimal (Likely TLE or MLE), but correct algorithm &
Major edits needed &
Does not make use or ensure the range could be calculated only from the range of $A_i$ values, yielding O(N) instead of expected O($\log N$) &
TLE \\
GPT 5 & &
\algok & Same as golden &
\xtypeNA & \xtypeNA &
\xtypeNA &
PASS \\

\rowcolor{LightGray}
\multicolumn{8}{@{}l}{\textsc{imperfection}} \\
DeepSeek R1 & &
\algok & Same as golden &
\xtypeNA & \xtypeNA & &
NA \\
GPT 5 & &
\algok & Same as golden &
\xtypeNA & \xtypeNA &
\xtypeNA &
PASS \\

\rowcolor{LightGray}
\multicolumn{8}{@{}l}{\textsc{jaunt through the garden}} \\
DeepSeek R1 & &
\algwrong & \xtypeNA &
Wrong algorithm &
Minor edits needed &
The golden editorial mention there exist an O(N) solution rather than O(N$^2$) DP detailed in the editorial. This formula seems to be what the generated editorial tries to achieve. &
WA \\
GPT 5 & &
\algok & Different from golden &
\xtypeNA & \xtypeNA &
The gold editorial leaves the O(N) solution for exercise to the reader. &
PASS \\

\rowcolor{LightGray}
\multicolumn{8}{@{}l}{\textsc{keep the ordering}} \\
DeepSeek R1 & &
\algok & Same as golden &
\xtypeNA & \xtypeNA &
Implementation inefficiency as it involves conversion to string for every check &
TLE \\
GPT 5 & &
\algok & Same as golden &
\xtypeNA & \xtypeNA &
\xtypeNA &
PASS \\

\bottomrule
\end{tabularx}
\caption{
Algorithmic correctness annotations and execution outcomes for the CS3233 2025 Midterm case study.
ALG COR indicates whether the editorial-level algorithm is correct under contest constraints.
Correct type distinguishes editorials that match the gold approach from those that use a different but valid approach.
For incorrect editorials, we report the diagnosed failure mode and severity.
Final verdict is the judge outcome of the code generated after the editorial in the generated-editorial setting.
}
\label{tab:cs3233-alg-cor}
\end{table*}

\section{LLM-as-a-Judge}

\subsection{LLM-as-a-Judge Prompts}
\label{appendix:llm-judge-prompts}

We evaluate LLM-generated editorials using an LLM-based judge with a fixed system prompt and a templated user prompt.
Figures~\ref{fig:judge-prompt-1}--\ref{fig:judge-prompt-4} show the full prompt used for LLM-based editorial evaluation. We instantiate the template by inserting the problem statement, the gold editorial, and the model-generated editorial.

All editorial evaluations are performed using
\texttt{google/gemini-3-pro-preview} accessed through OpenRouter \footnote{\url{https://openrouter.ai}}, with a maximum generation budget of
\texttt{65{,}536} tokens and \texttt{high} reasoning effort enabled.

\begin{figure*}
\begin{tcolorbox}[ title={LLM-as-a-judge prompt (editorial evaluation)}]
\small
\setlength{\parskip}{0pt}
\setlength{\parindent}{0pt}
\setlist[itemize]{leftmargin=*, topsep=1pt, itemsep=1pt, parsep=0pt, partopsep=0pt}
\textbf{System message.}

You are an expert competitive-programming judge and annotation assistant.

Your job is to:

1) Read the problem statement, the gold editorial, and the LLM-generated editorial.

2) Compare the LLM-generated editorial against the problem statement and the gold editorial.

3) Assign labels and short explanations according to the rubric and JSON schema provided in the user message.

STRICT FORMAT RULES (MUST OBEY):

- You MUST output exactly ONE JSON object.

- The JSON MUST be syntactically valid and parseable by json.loads in Python.

- The JSON MUST match the schema in the user message (all keys present).

- Use only the specified categorical values (case-sensitive).

- Use null where a field does not apply.

- Do NOT include any extra commentary, markdown, backticks, or text outside the JSON.

\medskip
\textbf{User message.}

\textbf{Goal}

Your task is to evaluate editorials, not code.
An editorial is a written explanation of the algorithm and the reasoning behind to solve a competitive-programming problem.

For each competitive-programming problem, you will assess TWO write-ups:

1) The gold editorial provided by the problem setter or tester (reference solution).

2) An LLM-generated editorial (the one you are evaluating).

You will assign ratings/labels in three categories:

1) Problem Understanding (PU)

2) Algorithm Description (ALG)

3) Algorithm Correctness (ALG-COR)

Your primary evaluation target is the LLM-generated editorial.
The gold editorial is used only as a reference for the intended solution.

---------------------------

INPUTS

---------------------------

Problem statement:

\texttt{<<PROBLEM\_STATEMENT>>}

Gold editorial:

\texttt{<<GOLD\_EDITORIAL>>}

LLM-generated editorial (the one you are evaluating):

\texttt{<<LLM\_EDITORIAL>>}

===========================

1. PROBLEM UNDERSTANDING (PU)

===========================

Purpose: Decide whether the LLM editorial correctly understands the problem statement.

You are checking only the LLM-generated editorial here.

\begin{itemize}[leftmargin=*]

\item \textbf{PU-W -- Wrong crucial detail}

\begin{itemize}[leftmargin=1.5em]
  \item \textbf{Question:} Does the LLM editorial assert something that changes the problem’s meaning?
  \item \texttt{"value"}: \texttt{"Yes"} or \texttt{"No"}
  \item If \texttt{"value"} = \texttt{"Yes"}, then:
    \begin{itemize}[leftmargin=1.5em]
      \item \texttt{"type"}:
        \begin{itemize}[leftmargin=1.5em]
          \item \texttt{"explicit"} if the wrong detail is directly stated
          \item \texttt{"implicit"} if it is strongly implied but not literally stated
        \end{itemize}
      \item \texttt{"notes"}: Short explanation describing the wrong detail and why it is crucial
    \end{itemize}
  \item If \texttt{"value"} = \texttt{"No"}:
    \begin{itemize}[leftmargin=1.5em]
      \item \texttt{"type"}: \texttt{null}
      \item \texttt{"notes"}: Short confirmation that there are no crucial wrong details
    \end{itemize}
\end{itemize}

\item \textbf{PU-M -- Missing crucial detail}

\begin{itemize}[leftmargin=1.5em]
  \item \textbf{Question:} Does the LLM editorial omit a constraint or subtlety that affects correctness?
  \item \texttt{"value"}: \texttt{"Yes"} or \texttt{"No"}
  \item If \texttt{"value"} = \texttt{"Yes"}, then:
    \begin{itemize}[leftmargin=1.5em]
      \item \texttt{"type"}:
        \begin{itemize}[leftmargin=1.5em]
          \item \texttt{"explicit"} if the missing info is explicitly present in the statement
          \item \texttt{"implicit"} if it is only implied but important
        \end{itemize}
      \item \texttt{"notes"}: Short explanation of the missing detail and why it matters
    \end{itemize}
  \item If \texttt{"value"} = \texttt{"No"}:
    \begin{itemize}[leftmargin=1.5em]
      \item \texttt{"type"}: \texttt{null}
      \item \texttt{"notes"}: Short confirmation that no crucial detail is missing
    \end{itemize}
\end{itemize}
\end{itemize}
\end{tcolorbox}
\caption{LLM-as-a-judge prompt for editorial evaluation. (part 1)}
\label{fig:judge-prompt-1}
\end{figure*}

\begin{figure*}
\begin{tcolorbox}
\small
\setlength{\parskip}{0pt}
\setlength{\parindent}{0pt}
\setlist[itemize]{leftmargin=*, topsep=1pt, itemsep=1pt, parsep=0pt, partopsep=0pt}
\begin{itemize}
\item \textbf{PU-X -- Irrelevant / misleading detail}

\begin{itemize}[leftmargin=1.5em]
  \item \textbf{Question:} Does the LLM editorial add extra statements that do not change correctness but muddy understanding?
  \item \texttt{"value"}: one of \texttt{"None"}, \texttt{"Minor"}, \texttt{"Major"}
  \item \texttt{"notes"}: Short explanation describing the extra or misleading material (if any)
\end{itemize}

\medskip
\item \textbf{PU-D -- Problem understanding difficulty}

\begin{itemize}[leftmargin=1.5em]
  \item \textbf{Question:} How difficult is the \emph{original problem statement} to understand (not the editorial)?
  \item \texttt{"value"}: integer in \([0,5]\)
    \begin{itemize}[leftmargin=1.5em]
      \item 0 = very clear — no difficulty
      \item 5 = extremely difficult
    \end{itemize}
  \item \texttt{"rationale"}: Short explanation of why this difficulty rating was chosen
\end{itemize}

\end{itemize}

===========================

2. ALGORITHM DESCRIPTION (ALG)

===========================

Purpose: Describe the high-level idea/algorithm in both editorials.

You must label the LLM editorial and the gold editorial separately.

For the LLM-generated editorial:

\begin{itemize}[leftmargin=*]

\item \textbf{ALG-TAG}

\begin{itemize}[leftmargin=1.5em]
  \item \textbf{Purpose:} High-level paradigms used in the LLM-generated editorial
  \item \texttt{"ALG-TAG"}: array of one or more strings
  \item Each element must be chosen from the following fixed list (case-sensitive):
    \begin{itemize}[leftmargin=1.5em]
      \item \texttt{"DP"}
      \item \texttt{"Greedy"}
      \item \texttt{"DFS/BFS"}
      \item \texttt{"Dijkstra"}
      \item \texttt{"Segment Tree"}
      \item \texttt{"Binary Lifting"}
      \item \texttt{"FFT"}
      \item \texttt{"Flow"}
      \item \texttt{"Geometry"}
      \item \texttt{"Math/Number Theory"}
      \item \texttt{"Other"}
    \end{itemize}
  \item If and only if \texttt{"Other"} is included, you MUST also provide extra tags in \texttt{"ALG-TAG-OTHER"}
\end{itemize}

\item \textbf{ALG-TAG-OTHER}

\begin{itemize}[leftmargin=1.5em]
  \item \texttt{"ALG-TAG-OTHER"}: array of strings (possibly empty)
  \item If \texttt{"Other"} is in \texttt{"ALG-TAG"}, then:
    \begin{itemize}[leftmargin=1.5em]
      \item Include one or more extra algorithm tags, for example:
        \begin{itemize}[leftmargin=1.5em]
          \item \texttt{"Meet-in-the-middle"}
          \item \texttt{"Bitmask brute force"}
          \item \texttt{"2-SAT via implication graph"}
          \item \texttt{"Game theory / Grundy numbers"}
        \end{itemize}
    \end{itemize}
  \item If \texttt{"Other"} is \textbf{NOT} in \texttt{"ALG-TAG"}, then:
    \begin{itemize}[leftmargin=1.5em]
      \item Set this field to \texttt{[]}
    \end{itemize}
\end{itemize}

\item \textbf{ALG-FREE}

\begin{itemize}[leftmargin=1.5em]
  \item \texttt{"ALG-FREE"}: one or two concise sentences (approximately 40 words total)
  \item Purpose: summarise the core idea of the LLM-generated editorial
\end{itemize}

\item \textbf{Gold editorial (reference solution)}

\begin{itemize}[leftmargin=1.5em]

  \item \textbf{Golden-ALG-TAG}

  \begin{itemize}[leftmargin=1.5em]
    \item High-level paradigms used in the gold editorial
    \item Same fixed tag list as \texttt{"ALG-TAG"}:
      \begin{itemize}[leftmargin=1.5em]
        \item \texttt{"DP"}
        \item \texttt{"Greedy"}
        \item \texttt{"DFS/BFS"}
        \item \texttt{"Dijkstra"}
        \item \texttt{"Segment Tree"}
        \item \texttt{"Binary Lifting"}
        \item \texttt{"FFT"}
        \item \texttt{"Flow"}
        \item \texttt{"Geometry"}
        \item \texttt{"Math/Number Theory"}
        \item \texttt{"Other"}
      \end{itemize}
    \item If and only if \texttt{"Other"} is included, you MUST also provide extra tags in \texttt{"Golden-ALG-TAG-OTHER"}
  \end{itemize}
\end{itemize}
\end{itemize}
\end{tcolorbox}
\caption{LLM-as-a-judge prompt for editorial evaluation. (part 2)}
\label{fig:judge-prompt-2}
\end{figure*}

\begin{figure*}
\begin{tcolorbox}
\small
\setlength{\parskip}{0pt}
\setlength{\parindent}{0pt}
\setlist[itemize]{leftmargin=*, topsep=1pt, itemsep=0pt, parsep=0pt, partopsep=0pt}

 \begin{itemize}
  \medskip
  \item \textbf{Golden-ALG-TAG-OTHER}

  \begin{itemize}[leftmargin=1.5em]
    \item \texttt{"Golden-ALG-TAG-OTHER"}: array of strings (possibly empty)
    \item If \texttt{"Other"} is in \texttt{"Golden-ALG-TAG"}, include one or more extra algorithm tags
    \item Otherwise, set this field to \texttt{[]}
  \end{itemize}

  \medskip
  \item \textbf{Golden-ALG-FREE}

  \begin{itemize}[leftmargin=1.5em]
    \item \texttt{"Golden-ALG-FREE"}: one or two concise sentences (approximately 40 words total)
    \item Purpose: summarise the core idea of the gold editorial
  \end{itemize}

\end{itemize}


===========================

3. ALGORITHM CORRECTNESS (ALG-COR)

===========================
Purpose: Evaluate whether the LLM editorial’s algorithm, as described, solves the problem correctly and efficiently under the given constraints.

You are judging the LLM-generated editorial here.

\begin{itemize}[leftmargin=*]

\item \textbf{ALG-COR.overall (overall correctness)}

\begin{itemize}[leftmargin=1.5em]
  \item \texttt{"overall"}: \texttt{"Correct"} or \texttt{"Incorrect"}
  \item Interpret \texttt{"Correct"} as: the described algorithm solves all valid inputs within time/memory limits.
  \item Interpret \texttt{"Incorrect"} as: there exists some valid input where the algorithm fails (wrong answer or infeasible resource usage).
\end{itemize}

\medskip
\item \textbf{Case 1: If \texttt{"overall"} = \texttt{"Correct"}}

\begin{itemize}[leftmargin=1.5em]
  \item \textbf{if\_correct.correct\_type}
    \begin{itemize}[leftmargin=1.5em]
      \item \texttt{"correct\_type"}: \texttt{"Same as golden"} or \texttt{"Different from golden"}
      \item \texttt{"notes"}: short explanation of this choice (e.g., why it matches or differs from the gold editorial).
      \item Use \texttt{null} for \texttt{"correct\_type"} and \texttt{"notes"} only if \texttt{"overall"} = \texttt{"Incorrect"}.
    \end{itemize}

  \item \texttt{"Same as golden"}:
    \begin{itemize}[leftmargin=1.5em]
      \item The LLM editorial matches the official approach (same core algorithmic idea).
    \end{itemize}

  \item \texttt{"Different from golden"}:
    \begin{itemize}[leftmargin=1.5em]
      \item The LLM editorial uses a different but still valid algorithm.
    \end{itemize}
\end{itemize}

\medskip
\item \textbf{Case 2: If \texttt{"overall"} = \texttt{"Incorrect"}}

\begin{itemize}[leftmargin=1.5em]

  \item \textbf{if\_incorrect.why\_incorrect}
    \begin{itemize}[leftmargin=1.5em]
      \item \texttt{"why\_incorrect"}: EXACTLY one of:
        \begin{itemize}[leftmargin=1.5em]
          \item \texttt{"Wrong algorithm"}
          \item \texttt{"Correct algorithm but incorrect approach"}
          \item \texttt{"Suboptimal but correct algorithm"}
          \item \texttt{"Suboptimal and wrong algorithm"}
          \item \texttt{null} (only if \texttt{"overall"} = \texttt{"Correct"})
        \end{itemize}
    \end{itemize}

  \item \textbf{Meaning}
    \begin{itemize}[leftmargin=1.5em]
      \item \texttt{"Wrong algorithm"}: The high-level algorithmic idea itself cannot solve the full problem.
      \item \texttt{"Correct algorithm but incorrect approach"}: The idea is sound, but there is a clear implementation or reasoning bug.
      \item \texttt{"Suboptimal but correct algorithm"}: The algorithm would produce correct answers but is too slow or memory-heavy for the stated constraints (likely TLE/MLE).
      \item \texttt{"Suboptimal and wrong algorithm"}: The algorithm is both incorrect on some inputs and too slow / resource-heavy.
    \end{itemize}

  \item \textbf{if\_incorrect.severity}
    \begin{itemize}[leftmargin=1.5em]
      \item \texttt{"severity"}: one of
        \begin{itemize}[leftmargin=1.5em]
          \item \texttt{"Completely wrong"}
          \item \texttt{"Major edits needed"}
          \item \texttt{"Minor edits needed"}
          \item \texttt{null} (only if \texttt{"overall"} = \texttt{"Correct"})
        \end{itemize}
    \end{itemize}

  \item \textbf{Meaning}
    \begin{itemize}[leftmargin=1.5em]
      \item \texttt{"Completely wrong"}: No substantial part of the solution is reusable.
      \item \texttt{"Major edits needed"}: Core structure is somewhat aligned but requires substantial changes.
      \item \texttt{"Minor edits needed"}: A small fix (e.g., off-by-one, missing boundary case) would make it correct.
    \end{itemize}

  \item \textbf{if\_incorrect.diagnosis}
    \begin{itemize}[leftmargin=1.5em]
      \item \texttt{"diagnosis"}: short natural-language explanation of why the algorithm fails
      (e.g., specific counterexample pattern, asymptotic complexity issue).
    \end{itemize}

\end{itemize}

\end{itemize}

\begin{itemize}
  \item If \texttt{"overall"} = \texttt{"Correct"}, then:
    \begin{itemize}
      \item Set \texttt{if\_incorrect.why\_incorrect},
            \texttt{if\_incorrect.severity}, and
            \texttt{if\_incorrect.diagnosis} to \texttt{null}.
    \end{itemize}

  \item If \texttt{"overall"} = \texttt{"Incorrect"}, then:
    \begin{itemize}
      \item Set \texttt{if\_correct.correct\_type} and
            \texttt{if\_correct.notes} to \texttt{null}.
    \end{itemize}
\end{itemize}

You MUST now apply these guidelines to the given problem statement, gold editorial, and LLM editorial and produce labels accordingly.

\end{tcolorbox}
\caption{LLM-as-a-judge prompt for editorial evaluation. (part 3)}
\label{fig:judge-prompt-3}
\end{figure*}

\begin{figure*}
\begin{tcolorbox}
\small
\setlength{\parskip}{0pt}
\setlength{\parindent}{0pt}
\setlist[itemize]{leftmargin=*, topsep=1pt, itemsep=0pt, parsep=0pt, partopsep=0pt}

---------------------------

REQUIRED JSON OUTPUT SCHEMA

---------------------------

Return ONLY a single valid JSON object with EXACTLY this structure:

\begingroup
\ttfamily
\begin{verbatim}
{
  "PU": {
    "PU-W": { "value": "Yes" or "No", "type": "explicit" or "implicit" or null, "notes": "..." },
    "PU-M": { "value": "Yes" or "No", "type": "explicit" or "implicit" or null, "notes": "..." },
    "PU-X": { "value": "None" or "Minor" or "Major", "notes": "..." },
    "PU-D": { "value": 0 or 1 or 2 or 3 or 4 or 5, "rationale": "..." }
  },
  "ALG": {
    "ALG-TAG": ["..."],
    "ALG-TAG-OTHER": ["..."],
    "ALG-FREE": "...",
    "Golden-ALG-TAG": ["..."],
    "Golden-ALG-TAG-OTHER": ["..."],
    "Golden-ALG-FREE": "..."
  },
  "ALG-COR": {
    "overall": "Correct" or "Incorrect",
    "if_correct": {
      "correct_type": "Same as golden" or "Different from golden" or null,
      "notes": "..." or null
    },
    "if_incorrect": {
      "why_incorrect": "Wrong algorithm" or "Correct algorithm but incorrect approach" or
      "Suboptimal but correct algorithm" or "Suboptimal and wrong algorithm" or null,
      "severity": "Completely wrong" or "Major edits needed" or "Minor edits needed" or null,
      "diagnosis": "..." or null
    }
  }
}
\end{verbatim}
\endgroup

IMPORTANT:
- Every key shown above MUST be present.
- Use null exactly where a field does not apply, as explained.
- Do NOT output any text outside the JSON object.

\end{tcolorbox}
\caption{LLM-as-a-judge prompt for editorial evaluation. (part 4)}
\label{fig:judge-prompt-4}
\end{figure*}

\subsection{Additional agreement results}
\label{app:additional-agreement}

For completeness, we additionally evaluate agreement between the expert annotator and the Gemini~3~Pro judge on auxiliary rubric fields not used as primary signals in our analysis, including the \textsc{PU} dimensions.

Table~\ref{tab:llm-judge-aux} reports raw agreement and Cohen’s $\kappa$ for these auxiliary fields.
While several \textsc{PU} fields exhibit high raw agreement, their chance-corrected agreement is low or near zero.
This behavior reflects a combination of strong class imbalance and limited sample size, under which Cohen’s $\kappa$ is known to be unstable.
Accordingly, we do not rely on these fields in our main analysis and report them here for completeness.

\begin{table}[b]
\centering
\small
\caption{
Agreement between the expert annotator and the Gemini 3 Pro judge on auxiliary rubric fields.
}
\label{tab:llm-judge-aux}
\setlength{\tabcolsep}{4pt}
\begin{tabular}{lccc}
\toprule
\textbf{Rubric field} & \textbf{$n$} & \textbf{Agr.} & \textbf{$\kappa$} \\
\midrule
PU-W (Yes vs.\ No) & 22 & 0.909 & $-0.048$ \\
PU-M (Yes vs.\ No) & 22 & 0.955 & 0.000 \\
PU-X (None/Minor/Major) & 22 & 0.818 & 0.000 \\
PU-D (0--5) & 22 & 0.318 & $-0.006$ \\
\bottomrule
\end{tabular}
\end{table}

\subsection{Additional judge-based diagnostics}
\label{app:llm-judge-extra-diagnostics}

The main paper uses the judge primarily for editorial-level correctness (\textsc{ALG-COR}) and its relationship to downstream outcomes.
Here we report additional judge fields that help contextualize \emph{why} editorials fail, and how often failures stem from problem understanding versus algorithmic reasoning.

\begin{figure*}[t]
\vspace{-6pt}
\centering
\begin{subfigure}[t]{0.49\textwidth}
  \centering
  \includegraphics[width=\linewidth,trim=0 5 0 6,clip]{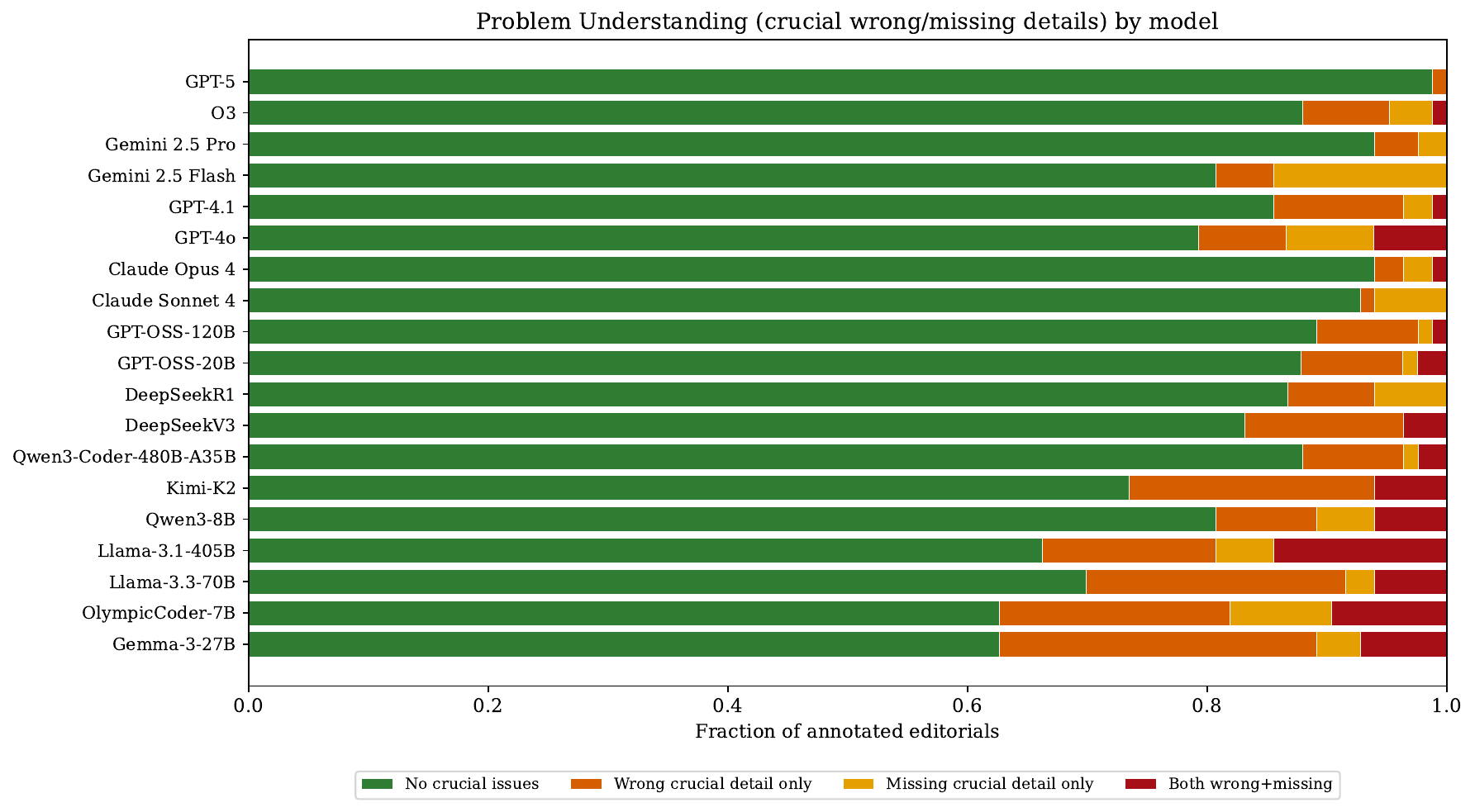}
  \label{fig:judge-pu-wm}
\end{subfigure}
\hfill
\begin{subfigure}[t]{0.49\textwidth}
  \centering
  \includegraphics[width=\linewidth,trim=0 5 0 6,clip]{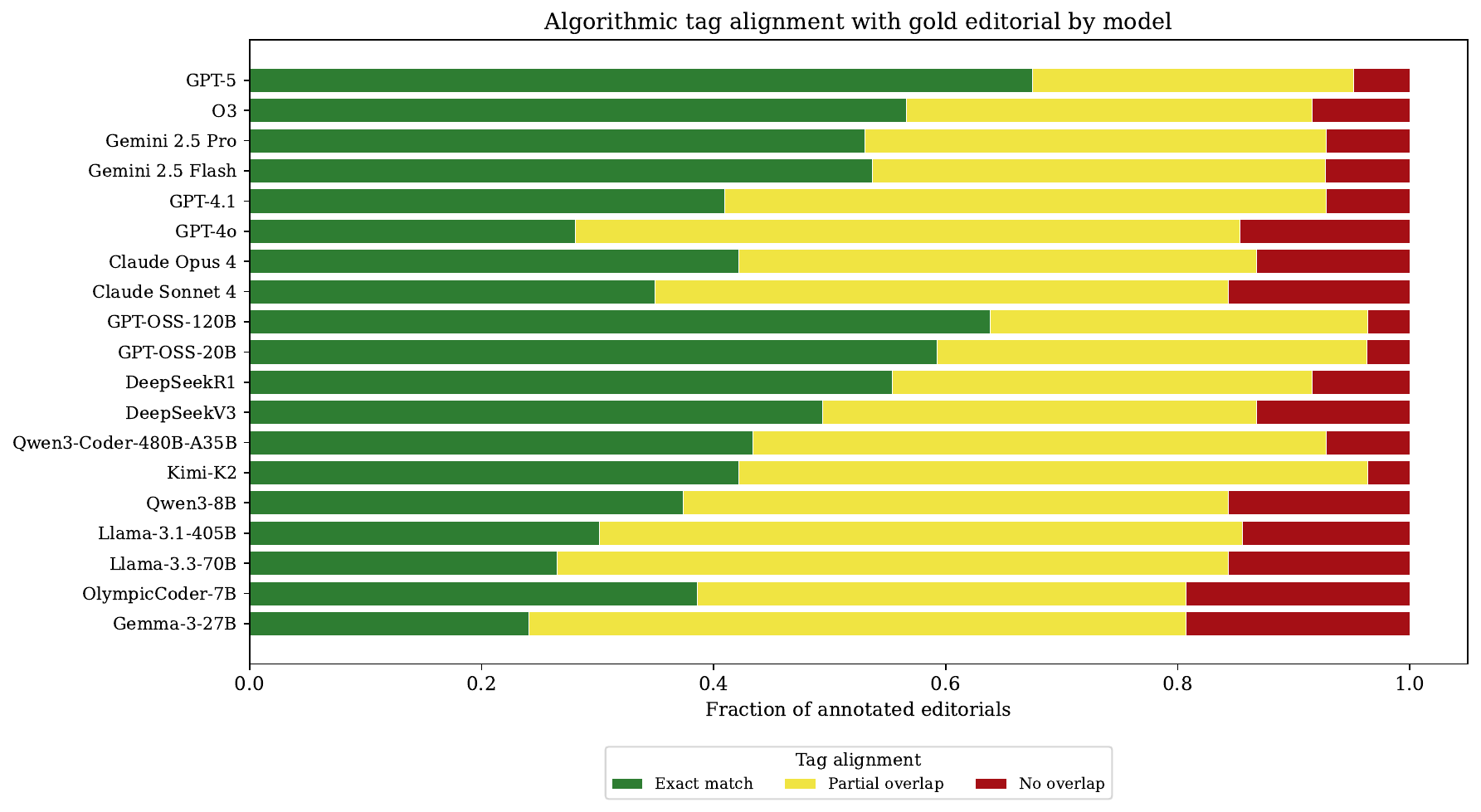}
  \label{fig:judge-tag-align}
\end{subfigure}
\vspace{6pt}
\begin{subfigure}[t]{0.49\textwidth}
  \centering
  \includegraphics[width=\linewidth,trim=0 5 0 6,clip]{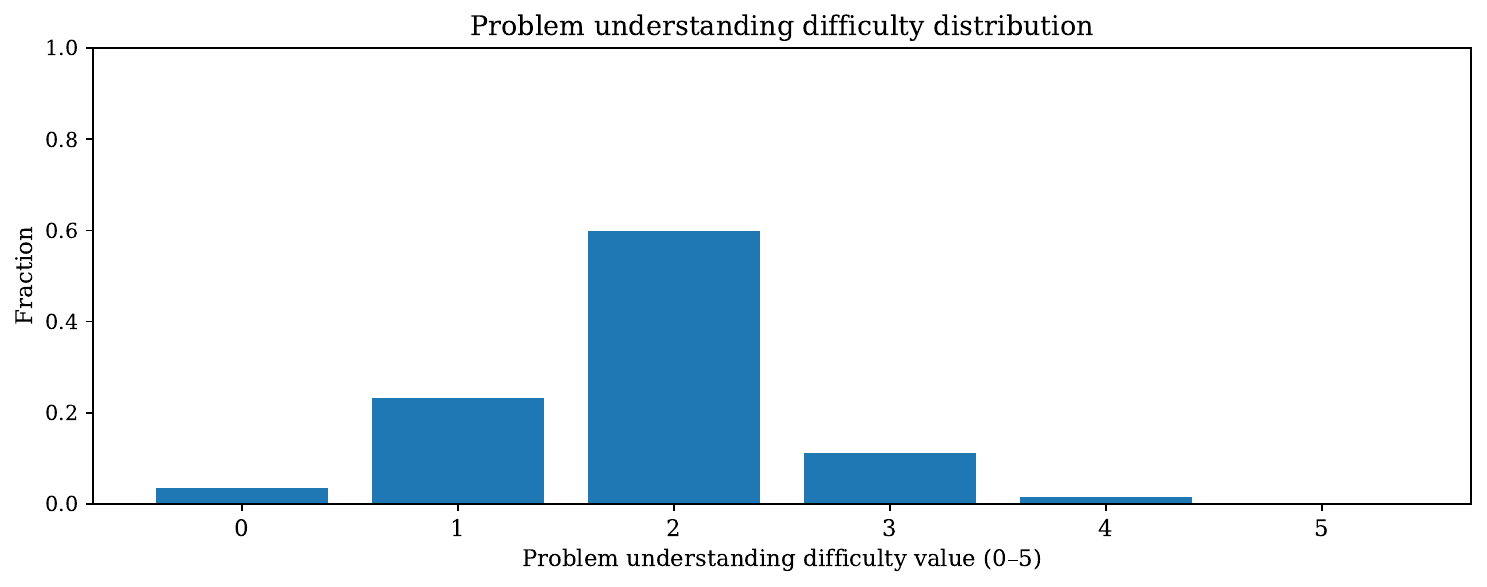}
  \label{fig:judge-pu-d}
\end{subfigure}
\hfill

\vspace{-6pt}
\caption{
Additional LLM-judge diagnostics for \textbf{w/GenEd} editorials.
Top-left: prevalence of crucial problem-understanding issues (wrong/missing details).
Bottom: distribution of judge-rated understanding difficulty (\textsc{PU-D}) across problems.
Top-right: algorithm-tag alignment between generated and gold editorials (exact match / partial overlap / no overlap).
}
\label{fig:llm-judge-extra-panels}
\vspace{-10pt}
\end{figure*}

\paragraph{Problem-understanding errors are uncommon compared to algorithmic errors.}
Figure~\ref{fig:llm-judge-extra-panels} (top-left) shows that most generated editorials do not exhibit \emph{crucial} wrong or missing problem details, indicating that gross misinterpretation is relatively rare.
However, the fraction of such errors increases for open-weight models.
In competitive programming, correct problem understanding is a strict prerequisite rather than an optional skill; even a small error rate at this stage is severe, as it renders all downstream reasoning and implementation invalid.
Thus, while less frequent than algorithmic failures, problem-understanding errors remain a non-trivial and important limitation.

\paragraph{Most problems are rated as medium to understand.}
Figure~\ref{fig:llm-judge-extra-panels} (bottom-left) shows the distribution of \textsc{PU-D} scores, with most mass concentrated at intermediate values.
This supports our interpretation that the dominant difficulty in this benchmark is deriving a correct and efficient algorithm, not parsing the statement.

\paragraph{Tag alignment is a coarse but imperfect proxy for correctness.}
Figure~\ref{fig:llm-judge-extra-panels} (bottom-right) reports the overlap between generated and gold algorithm tags.
While exact matches are common for stronger models, partial overlap is also frequent and ``no overlap'' remains non-trivial for weaker models.
Consistent with Appendix~\ref{app:corr-by-model}, tag overlap is a weaker predictor of downstream success than \textsc{ALG-COR}.

\subsection{Full verdict breakdown by six-way \textsc{ALG-COR}}
\label{app:llm-judge-6way-verdict}

\begin{figure}[t]
\vspace{-6pt}
\centering
\includegraphics[width=\linewidth,trim=0 6 0 6,clip]{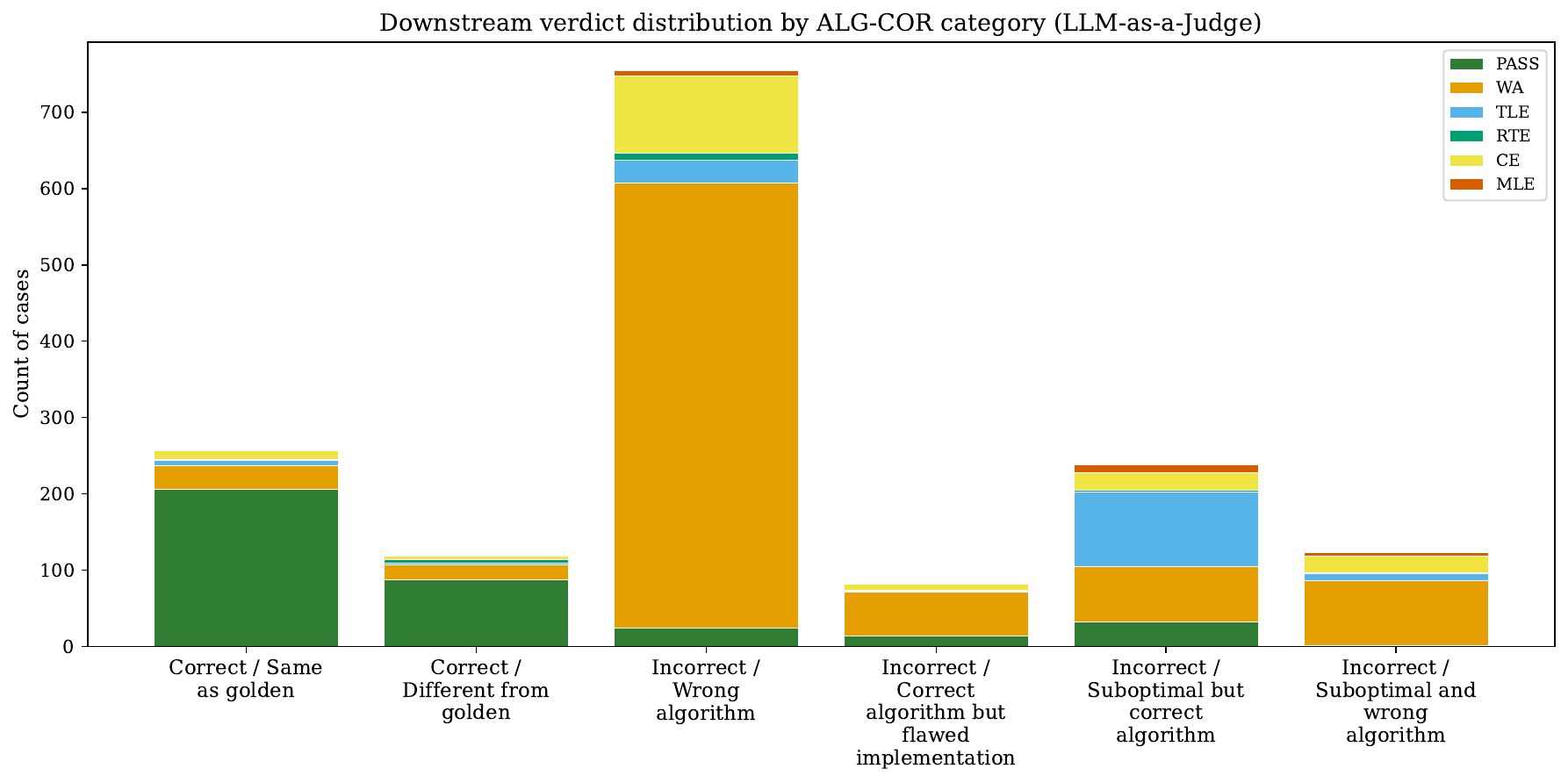}
\vspace{-6pt}
\caption{
Downstream verdict counts (PASS/WA/TLE/RTE/CE/MLE) for \textbf{w/GenEd} code, stratified by the full six-way \textsc{ALG-COR} label.
This complements Figure~\ref{fig:llm-judge-vs-verdict} by showing absolute sample sizes per category.
}
\label{fig:llm-judge-6way-verdict-fig}
\vspace{-10pt}
\end{figure}

In Figure~\ref{fig:llm-judge-vs-verdict} we collapse \textsc{ALG-COR} into three groups for readability.
Figure~\ref{fig:llm-judge-6way-verdict-fig} provides the full six-way breakdown, which makes the mapping from fine-grained judge diagnoses to runtime outcomes explicit.

\subsection{Correlates of downstream success under generated editorials}
\label{app:corr-by-model}

To complement the qualitative validation of the LLM judge (Table~\ref{tab:llm-judge-agreement}). For each model, across the 83 problems, we correlate each LLM-judge rubric field for the
generated editorial with the binary indicator $\mathbb{1}[T(C)=\textsc{PASS}]$ for the resulting code.
Binary fields use the $\phi$ coefficient; ordinal/numeric fields use Spearman's $\rho$.
Stars denote two-sided significance: $^{***}p<0.001$, $^{**}p<0.01$, $^{*}p<0.05$.
``--'' indicates the field had no variance for that model (correlation undefined).

Overall, \texttt{ALG-COR\_overall} is the strongest and most consistent correlate of passing for
\emph{every} model (all significant), supporting our use of judge-labeled correctness as a scalable
proxy for reasoning quality. Tag alignment (\texttt{tag\_equal}, \texttt{tag\_jaccard}) is a weaker but
often significant signal, while \texttt{PU-D\_value} (annotator-rated understanding difficulty) is
frequently negatively associated with success. Length features (\texttt{editorial\_words},
\texttt{code\_lines}) are negatively correlated for several models, which we interpret primarily as a
difficulty confound rather than a causal effect of verbosity.

\begin{table*}
\centering
\scriptsize
\setlength{\tabcolsep}{2.2pt}
\renewcommand{\arraystretch}{1.15}
\caption{
Per-model correlations between LLM-judge editorial diagnostics and downstream success under \textbf{w/GenEd}.
Top row reports each model's \textbf{w/GenEd pass@1} for context; subsequent rows report correlation with
$\mathbb{1}[T(C)=\textsc{PASS}]$.
Binary fields use $\phi$; ordinal/numeric fields use Spearman's $\rho$.
Stars denote two-sided significance: $^{***}p<0.001$, $^{**}p<0.01$, $^{*}p<0.05$.
``--'' indicates undefined correlation due to zero variance.
}
\label{tab:corr-by-model}
\begin{adjustbox}{max width=\textwidth}
\begin{tabular}{lccccccccccccccccccc}
\toprule
 & \makecell{Claude\\Sonnet 4} & \makecell{O3\\(medium)} & \makecell{Qwen3\\8B} & \makecell{Qwen3-Coder\\480B-A35B} &
\makecell{Claude\\Opus 4} & \makecell{DeepSeek\\V3} & \makecell{DeepSeek\\R1} &
\makecell{Gemini\\2.5 Flash} & \makecell{Gemini\\2.5 Pro} & \makecell{Gemma-3\\27B} &
\makecell{GPT\\4.1} & \makecell{GPT\\4o} & \makecell{GPT\\5} &
\makecell{Llama\\3.3-70B} & \makecell{Llama\\3.1-405B} & \makecell{Kimi\\K2} &
\makecell{Olympic\\Coder-7B} & \makecell{GPT-OSS\\120B} & \makecell{GPT-OSS\\20B} \\
\midrule
\textbf{Pass@1} &
0.193 & 0.458 & 0.130 & 0.110 & 0.309 & 0.100 & 0.475 & 0.384 & 0.450 & 0.024 & 0.175 & 0.037 & 0.679 & 0.060 & 0.024 & 0.133 & 0.086 & 0.313 & 0.295 \\
\midrule
\texttt{ALG-COR\_overall} &
0.644*** & 0.640*** & 0.719*** & 0.738*** & 0.581*** & 0.614*** & 0.659*** & 0.708*** & 0.772*** & 0.703*** & 0.671*** & 0.444*** & 0.654*** & 0.765*** & 0.391*** & 0.790*** & 0.469*** & 0.790*** & 0.711*** \\
\texttt{PU\_X\_ord} &
0.133 & 0.171 & 0.145 & 0.053 & -0.101 & 0.180 & -- & -0.149 & 0.016 & -0.038 & 0.040 & 0.075 & -- & -0.079 & -0.066 & 0.010 & -0.108 & 0.063 & 0.051 \\
\texttt{PU-W\_value} &
-0.054 & -0.279* & -0.141 & -0.123 & -0.131 & -0.140 & -0.176 & -0.066 & -0.179 & -0.114 & -0.174 & -0.077 & -0.163 & -0.157 & -0.100 & -0.235* & -0.200 & -0.221* & -0.203 \\
\texttt{PU-D\_value} &
-0.315** & -0.221* & -0.514*** & -0.235* & -0.379*** & -0.250* & -0.102 & -0.194 & -0.187 & -0.319** & -0.165 & -0.298** & -0.146 & -0.228* & -0.189 & -0.186 & -0.282* & -0.180 & -0.257* \\
\texttt{tag\_equal} &
0.347** & 0.121 & 0.417*** & 0.240* & 0.298** & 0.167 & 0.330** & 0.258* & 0.234* & 0.101 & 0.403*** & 0.168 & 0.150 & 0.077 & -0.103 & 0.314** & 0.210 & 0.184 & 0.326** \\
\texttt{tag\_jaccard} &
0.322** & 0.195 & 0.333** & 0.240* & 0.225* & 0.069 & 0.282* & 0.297* & 0.225* & -0.000 & 0.365*** & 0.168 & 0.078 & -0.013 & -0.147 & 0.303** & 0.173 & 0.209 & 0.323** \\
\texttt{editorial\_words} &
-0.086 & 0.002 & -0.086 & 0.034 & -0.309** & -0.033 & -0.197 & -0.139 & -0.146 & -0.242* & -0.303** & 0.005 & -0.338** & -0.117 & -0.277* & -0.335** & -0.061 & -0.243* & -0.139 \\
\texttt{code\_lines} &
-0.213 & 0.074 & -0.064 & -0.195 & -0.283* & -0.285* & -0.348** & -0.178 & -0.309** & -0.114 & -0.186 & -0.223* & -0.183 & -0.230* & -0.227* & -0.152 & -0.263* & -0.201 & -0.217 \\
\bottomrule
\end{tabular}
\end{adjustbox}
\end{table*}

\section{Extended Related Work}
\label{app:extended-related}

\subsection{Competitive Programming Benchmarks}

\paragraph{End-to-end code generation and test-suite quality.}
A large fraction of LLM-for-code evaluation uses an end-to-end setup where models translate problem statements into code
and are scored by unit tests. APPS \cite{hendrycksapps2021} and HumanEval \cite{chen2021codex} are canonical benchmarks in
this paradigm. Because unit tests can be incomplete, multiple works aim to strengthen correctness evaluation by augmenting
tests; EvalPlus \cite{liu2023is} expands HumanEval/MBPP test suites and shows that stronger tests can change absolute
scores and even model rankings.

\paragraph{Contest-style benchmarks and competitive programming protocols.}
Contest-oriented evaluation draws problems from competitive platforms and uses contest-style judging. AlphaCode introduced
competition-level generation and released the CodeContests dataset \cite{Li_2022}. LiveCodeBench \cite{jain2024livecodebench}
improves temporal realism by continuously collecting fresh problems and explicitly discussing contamination. USACO
\cite{shi2024languagemodelssolveolympiad} and LLM-ProS \cite{hossain2025llmprosanalyzinglargelanguage} focus on higher
difficulty distributions (Olympiad/ICPC-style), and CodeElo \cite{quan2025codeelo} evaluates via Codeforces-like submissions
and maps results to human-comparable Elo ratings. Since evaluation quality depends critically on test coverage, CodeContests+
constructs higher-quality/verified test cases for the CodeContests problem set \cite{wang-etal-2025-codecontests}.

\paragraph{Decomposed evaluation and process benchmarks.}
Most benchmarks still primarily score the final program, making it difficult to localize failures to reasoning vs.
implementation. Coding Triangle explicitly evaluates multiple dimensions of programming capability---editorial analysis,
code implementation, and test-case generation \cite{zhang2025codingtriangledoeslarge}. ELABORATION benchmarks the broader human--LLM
competitive programming process, introducing a taxonomy of human feedback and a protocol/dataset for staged interaction
\cite{yang-etal-2025-elaboration}. Our work shares the goal of decomposing the pipeline but focuses on a fully automated
editorial-centric setting: editorials act as explicit intermediate plans that can be independently evaluated and transferred
across models.

\subsection{Code LLMs, Reasoning LLMs, and Multi-stage Pipelines}

AlphaCode \cite{Li_2022} demonstrated that large-scale sampling and selection can yield strong contest performance, while
code-specialized models (e.g., CodeGeeX \cite{zheng2024codegeexpretrainedmodelcode}, StarCoder \cite{li2023starcodersourceyou},
Code Llama \cite{rozière2024codellamaopenfoundation}) improved implementation quality on standard code-generation benchmarks.
Reasoning-oriented models such as OpenAI's O-series \cite{openai2025competitiveprogramminglargereasoning} and DeepSeek-R1
\cite{deepseekai2025deepseekr1incentivizingreasoningcapability} further highlight the importance of multi-step reasoning for
harder tasks.

Multi-stage methods aim to improve reliability by introducing explicit intermediate steps or feedback. MapCoder structures
generation into retrieval, planning, coding, and debugging agents \cite{islam-etal-2024-mapcoder}. CodeT selects among
multiple candidate programs using generated tests \cite{chen2023codet}. Our approach is complementary: we treat editorials as
first-class intermediate artifacts, directly measuring plan quality and implementation fidelity, and enabling modular
writer--coder composition through editorial transfer.


\section{Test-Time Feedback on Generated Editorials (Exploratory)}
\label{app:feedback-ablation}

All results so far use a strictly single-shot setup, where a model generates one editorial $E$ and one program $C$. We introduce a small ablation that adds limited test-time feedback \emph{only in the w/GenEd setting}, allowing the model to revise its outputs using coarse signals.

Starting from
\(
E^{(0)} = f_{\mathrm{ed}}(P), \qquad
C^{(0)} = f_{\mathrm{code}}(P, E^{(0)}),
\)
we allow a bounded number of revisions to the editorial and/or the code. No test inputs or reference outputs are revealed; feedback is limited to judge verdicts or editorial-level self-assessment. We evaluate this setting on four representative open-weight models. Editorial refinement uses the same rubric as Appendix~\ref{appendix:annotation-rubric}. Since editorial-level algorithmic correctness strongly predicts downstream outcomes, it provides a natural signal for self-refinement.

\paragraph{Formalization.}
At iteration $t$, the model maintains a pair $(E^{(t)}, C^{(t)})$. Let $T(C)$ denote the judge verdict for a program, and $A(E)$ the self-assessment of an editorial. Updates differ by feedback variant.

\paragraph{Code feedback only.}
The editorial is fixed, and only the code is revised using verdict feedback:
\(
E^{(t+1)} = E^{(0)},
C^{(t+1)} = f_{\mathrm{code}}\!\bigl(P, E^{(0)}, C^{(t)}, T(C^{(t)})\bigr).
\)

\paragraph{Editorial refinement only.}
The editorial is iteratively refined, and code is generated once at the end:
\(
E^{(t+1)} = f_{\mathrm{ed}}\!\bigl(P, E^{(t)}, A(E^{(t)})\bigr),
C = f_{\mathrm{code}}\!\bigl(P, E^{(T_E)}\bigr).
\)

\paragraph{Editorial refinement + code feedback.}
We first refine the editorial, then revise the code:
\(
E^{(t+1)} = f_{\mathrm{ed}}\!\bigl(P, E^{(t)}, A(E^{(t)})\bigr),
\)
\(
C^{(0)} = f_{\mathrm{code}}\!\bigl(P, E^{(T_E)}\bigr),
\)
\(
C^{(k+1)} = f_{\mathrm{code}}\!\bigl(P, E^{(T_E)}, C^{(k)}, T(C^{(k)})\bigr).
\)

\noindent We use small fixed budgets, with $T_E$, $T_C \le 5$.

\begin{figure}[t]
    \centering
    \includegraphics[width=\linewidth]{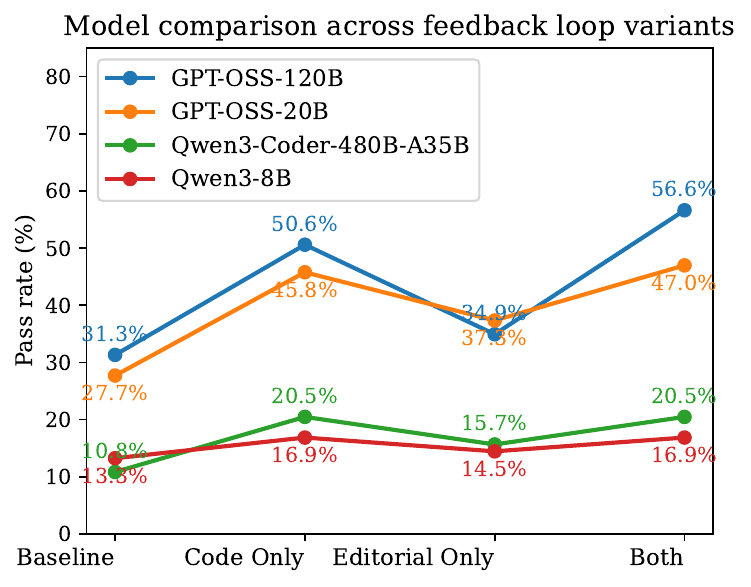}
    \caption{Pass@1 comparison across test-time feedback variants for four open-weight models.
    Baseline corresponds to w/GenEd without feedback.}
    \label{fig:feedback-ablation}
\end{figure}

Figure~\ref{fig:feedback-ablation} reports pass@1 across variants. Code-only feedback yields the largest gains over w/GenEd, showing that coarse verdicts are effective for repairing implementations. Editorial refinement alone provides smaller but generally positive improvements. Combining both yields additional gains for GPT-OSS models but little benefit over code-only feedback for Qwen variants.

\paragraph{Findings.}
Limited test-time feedback can substantially improve performance. Most gains come from iterative code repair, while editorial refinement helps correct flawed plans and steer revisions toward viable solution strategies.

\section{Gold vs.\ Model-Generated Editorials Examples}
\label{app:examples}

Figures~\ref{fig:app-dependency-flood-1}--~\ref{fig:app-dependency-flood-3} provides a detailed comparison between gold and model-generated editorials for a representative problem (\textsc{dependency flood}). Both editorials describe the same algorithm and satisfy the same correctness invariant.
However, the model editorial is judged easier to understand because it explicitly linearizes
the reasoning process: it first reframes the problem as longest-path maintenance,
then isolates the accept/reject condition, and finally explains how local updates propagate
and why they terminate within the given constraints.

Figures~\ref{fig:app-hungry-piplups-1}--~\ref{fig:app-hungry-piplups-3} provides a detailed comparison between gold and model-generated editorials for a representative problem (\textsc{Hungry Piplups}).
Both the gold and model editorials describe the same algorithmic approach and are labeled \textbf{ALG-COR = Correct (Same as golden)}. The model editorial is easier to operationalize due to its explicit left-to-right backlog interpretation and step-by-step reasoning, whereas the gold editorial assumes familiarity with standard competitive-programming abstractions.

\begin{figure*}
\begin{tcolorbox}[ title={\textsc{dependency flood}: Full text with highlighted diagnostic evidence}]
\small

\textbf{Problem statement.}

A university offers $N$ courses, numbered from $1$ to $N$ in increasing order of difficulty.

To ensure that students acquire the necessary knowledge before taking advanced courses, the university may impose
\textbf{course dependencies}. Each course dependency is represented by a pair of integers
$(u, v)\ (1 \le u < v \le N)$, meaning that students must complete course $u$ before enrolling in course $v$.

A set of course dependencies is called \textbf{acceptable} if all $N$ courses can be completed within $K$ semesters,
assuming that a student can take any number of courses (possibly zero) in a semester.

Formally, a set of course dependencies is acceptable if and only if there exists a sequence of $N$ numbers
$a_1, a_2, \ldots, a_N\ (1 \le a_i \le K)$ such that
\hl{$a_u < a_v$ for every $(u, v)$} in the set.

Let $S$ be the set of course dependencies. Initially, $S$ consists of $M$ course dependencies
$(A_1, B_1), (A_2, B_2), \ldots, (A_M, B_M)$, and
\hl{\textbf{it is guaranteed that $S$ is acceptable.}}

You need to process $Q$ queries sequentially. The $i$-th query $(1 \le i \le Q)$ is as follows:

\begin{itemize}
  \item Given two integers $C_i, D_i\ (1 \le C_i < D_i \le N)$, determine whether adding a new dependency
  $(C_i, D_i)$ to $S$ maintains its acceptability. If adding this dependency makes $S$ unacceptable, output
  \textbf{reject} and leave $S$ unchanged. Otherwise, output \textbf{accept} and add $(C_i, D_i)$ to $S$.
\end{itemize}


\textbf{Input.} The first line of input contains three integers $N,\ M$ and $K\ (2 \le N \le 2 \times 10^5,\ 0 \le M \le 2 \times 10^5,\ 1 \le K \le 100)$, representing the number of courses, the number of initial course dependencies, and the number of semesters, respectively.

\small
The $i$-th line $(1 \le i \le M)$ of the next $M$ lines contain two integers $A_i$ and $B_i\ (1 \le A_i < B_i \le N)$, denoting the $i$-th initial course dependency.

The following line contains an integer $Q\ (1 \le Q \le 2 \times 10^5)$, the number of queries.

The $i$-th line $(1 \le i \le Q)$ of the next $Q$ lines contain two integers $C_i$ and $D_i\ (1 \le C_i < D_i \le N)$, representing the $i$-th query.

\medskip

\textbf{Output.} Output $Q$ lines. The $i$-th line $(1 \le i \le Q)$ of the output should contain the result for the $i$-th query: Print \textbf{accept} if adding the dependency $(C_i, D_i)$ keeps $S$ acceptable, or \textbf{reject} if it makes $S$ unacceptable.
\medskip

\textbf{Sample Input \#1}
\begin{verbatim}
4 1 2
1 2
3
2 3
3 4
1 3
\end{verbatim}

\textbf{Sample Output \#1}
\begin{verbatim}
reject
accept
reject
\end{verbatim}

\textbf{Sample Input \#2}
\begin{verbatim}
6 4 3
2 5
1 3
1 4
4 6
8
2 6
3 4
4 5
1 3
2 4
3 6
2 3
5 6
\end{verbatim}

\textbf{Sample Output \#2}
\begin{verbatim}
accept
reject
accept
accept
accept
accept
accept
reject
\end{verbatim}
\end{tcolorbox}
\caption{Full problem, gold editorial, and model-generated editorial for \textsc{Dependency Flood}, with highlighted diagnostic evidence. (part 1)}
\label{fig:app-dependency-flood-1}
\end{figure*}

\begin{figure*}
\begin{tcolorbox}
\small
\textbf{Gold editorial.}

For simplicity, we will use graph theory terminology instead of the definitions given in the problem statement.
\hl{Note that acceptability is equivalent to ensuring that there is no path of length $K$.}

Let $f_i$ ($1 \leq i \leq N$) be the maximum path length ending at vertex $i$, and let $b_i$
($1 \leq i \leq N$) be the maximum path length starting from vertex $i$.
The values of $f_i$ satisfy the following recurrence relation, where $E$ denotes the set of edges:

\[
f_i =
\left\{
\begin{array}{ll}
1 + \max_{(j, i) \in E} f_j & \exists (j, i) \in E \\
0 & \text{otherwise}
\end{array}
\right.
\]

Thus, the initial values of $f_i$ can be computed in $\mathcal{O}(N + M)$ time by processing vertices in ascending order.
Similarly, the initial values of $b_i$ can be computed in descending order.

If we can efficiently maintain the values of $f_i$ and $b_i$, checking whether a new edge can be added becomes
straightforward using the following lemma:

\medskip
\textbf{Lemma.}
\hl{\emph{A new edge $(u, v)$ can be added if and only if $f_u + b_v + 1 < K$.}}

\textbf{Proof.} Exercise.

\small
The key challenge is efficiently updating $f_i$ and $b_i$ when a new edge $(u, v)$ is added.
We describe the update process for $f_i$ (the update for $b_i$ follows analogously).

Given the way $f_i$ is initially computed, we can derive an update algorithm using a simple DFS, as shown in the following pseudocode:


\small
\begin{verbatim}
Function update(i, val):
    f[i] = val
    for each edge (i, j):
        if f[j] < f[i] + 1:
            update(j, f[i] + 1)
Function add_new_edge(u, v):
    if f[v] < f[u] + 1:
        update(v, f[u] + 1)
\end{verbatim}

This recursive algorithm appears to require $\mathcal{O}(N + M)$ time per edge addition.
However, we observe that the function update is called at most $(K - 1)N$ times in total.
This is because:
\begin{itemize}
  \item Every call of the function increases some $f_i$ by at least one.
  \item \hl{The sum of all $f_i$ values never exceeds $(K - 1) \times N$.}
  \item The values of $f_i$ never decrease.
\end{itemize}

Hence we can solve this problem in the overall time complexity of
\colorbox{yellow!35}{$\mathcal{O}(Q + K(N + M))$}.

\medskip\hrule\medskip

\textbf{Model editorial (DeepSeek-R1)}

\medskip
\textbf{Problem Analysis.} The problem involves managing a set of course dependencies in a directed acyclic graph (DAG)
where nodes represent courses and edges represent dependencies (an edge from \(u\) to \(v\)
means course \(u\) must be completed before \(v\)). The goal is to process queries that add new dependencies while ensuring that the entire set of courses can be scheduled within \(K\) semesters. 
\colorbox{yellow!35}{The scheduling constraint requires that the longest chain of dependencies (a path of consecutive dependencies) does }

\colorbox{yellow!35}{not exceed \(K\) courses, as each course in the chain must be in a distinct semester.}

\medskip
\textbf{Key Insight.} The core insight is that the acceptability of the dependency set is determined by the length of the longest chain (in terms of the number of nodes). A chain of length \(L\) requires at least \(L\) semesters. Thus, the dependency set is acceptable only if the longest chain in the DAG has length at most \(K\). 

For each query to add a new edge \((C, D)\), the critical check is whether the new chain formed by concatenating the longest chain ending at \(C\) and the longest chain starting at \(D\) (via the new edge) exceeds \(K\) nodes. Specifically, \colorbox{yellow!35}{if the sum of the lengths of these two chains is at least \(K+1\), adding the edge creates a chain that is too long,}
\colorbox{yellow!35}{making the set unacceptable.}

\end{tcolorbox}
\caption{Full problem, gold editorial, and model-generated editorial for \textsc{Dependency Flood}, with highlighted diagnostic evidence. (part 2)}
\label{fig:app-dependency-flood-2}
\end{figure*}

\begin{figure*}
\begin{tcolorbox}
\small
\textbf{Algorithm Selection.}

\begin{itemize}
  \item \textbf{Precomputation}:
    \begin{itemize}
      \item \textbf{Initial Graph Setup}: Store the initial dependencies in adjacency lists for outgoing and incoming edges.
      \item \textbf{Dynamic Programming Arrays}:
        \begin{itemize}
          \item \(\texttt{dp\_in}[u]\) represents the length of the longest chain ending at node \(u\).
          \item \(\texttt{dp\_out}[u]\) represents the length of the longest chain starting at node \(u\).
        \end{itemize}
      \item Compute \texttt{dp\_in} by processing nodes in increasing order (since edges go from lower to higher indices, this is a topological order). For each node \(u\), \texttt{dp\_in[u]} is the maximum of 1 and \(\texttt{dp\_in[v]} + 1\) for all incoming edges \((v, u)\).
      \item Compute \texttt{dp\_out} by processing nodes in decreasing order. For each node \(u\), \texttt{dp\_out[u]} is the maximum of 1 and \(\texttt{dp\_out[w]} + 1\) for all outgoing edges \((u, w)\).
    \end{itemize}
    \item \textbf{Query Processing}:
    \begin{itemize}
      \item For each query \((C, D)\):
      \begin{itemize}
          \item If the edge \((C, D)\) already exists, output ``accept'' (no change).
          \item Otherwise, check if \colorbox{yellow!35}{\(\texttt{dp\_in[C]} + \texttt{dp\_out[D]} \ge K+1\)}. If true, output ``reject'' (adding the edge would create a chain longer than \(K\)).
          \item If false, output ``accept'' and add the edge to the graph. Then update:
          \begin{itemize}
              \item \textbf{Update \texttt{dp\_in} for \(D\)}: Set \(\texttt{dp\_in[D]} = \max(\texttt{dp\_in[D]}, \texttt{dp\_in[C]} + 1)\) if this new value is \(\le K\). Propagate this update to all descendants of \(D\) using BFS: for each descendant \(w\), if \(\texttt{dp\_in[u]} + 1\) (where \(u\) is the current node) is greater than \(\texttt{dp\_in[w]}\) and \(\le K\), update \(\texttt{dp\_in[w]}\) and continue propagating.
              \item \textbf{Update \texttt{dp\_out} for \(C\)}: Set \(\texttt{dp\_out[C]} = \max(\texttt{dp\_out[C]}, \texttt{dp\_out[D]} + 1)\) if this new value is \(\le K\). Propagate this update to all ancestors of \(C\) using BFS: for each ancestor \(w\), if \(\texttt{dp\_out[u]} + 1\) is greater than \(\texttt{dp\_out[w]}\) and \(\le K\), update \(\texttt{dp\_out[w]}\) and continue propagating.
        \end{itemize}

        \end{itemize}

    \end{itemize}

\end{itemize}

\small
\begin{itemize}
  \item \textbf{Complexity Analysis}:
    \begin{itemize}
      \item \textbf{Precomputation}:
        \begin{itemize}
          \item Building adjacency lists: \(O(N + M)\).
          \item Computing \texttt{dp\_in} and \texttt{dp\_out}: \(O(N + M)\) each.
        \end{itemize}
      \item \textbf{Query Processing}:
        \begin{itemize}
          \item Each query check: \(O(1)\).
          \item Edge addition and propagation: \colorbox{yellow!35}{Each node can be updated at most \(K\) times (since chain lengths are}
          \colorbox{yellow!35}{bounded by \(K\)).} Each update processes the node's outgoing or incoming edges. The total work over all queries is \(O(K \cdot (M + Q))\), where \(M + Q\) is the total number of edges after all queries.
        \end{itemize}
      \item \textbf{Overall Complexity}: \(O(N + M + Q + K \cdot (M + Q))\). Given \(K \le 100\) and \(N, M, Q \le 2 \times 10^{5}\), this is efficient.
    \end{itemize}
\end{itemize}

\textbf{Explanation}

\begin{itemize}
  \item \textbf{Initial Setup}: The initial graph is built from input, storing edges in adjacency lists and a set for quick lookup.
  \item \textbf{Precomputation}: The \texttt{dp\_in} array is computed by processing nodes from 1 to \(N\), updating each node's value based on incoming edges. The \texttt{dp\_out} array is computed by processing nodes from \(N\) down to 1, updating each node's value based on outgoing edges.
  \item \textbf{Query Handling}: For each query:
    \begin{itemize}
      \item If the edge exists, ``accept'' is outputted immediately.
      \item Otherwise, the sum \(\texttt{dp\_in[C]} + \texttt{dp\_out[D]}\) is checked. If it exceeds \(K\), the edge is rejected.
      \item If accepted, the edge is added to the graph, and BFS propagates updates to \texttt{dp\_in} for \(D\)'s descendants and \texttt{dp\_out} for \(C\)'s ancestors, ensuring all affected nodes reflect the new longest chains within the \(K\) constraint.
    \end{itemize}
  \item \textbf{Efficiency}: The BFS propagation is bounded by \(K\), ensuring each node is updated at most \(K\) times, making the solution efficient for large inputs. The overall approach efficiently maintains the DAG properties and checks the scheduling constraint dynamically.
\end{itemize}

\end{tcolorbox}

\caption{Full problem, gold editorial, and model-generated editorial for \textsc{Dependency Flood}, with highlighted diagnostic evidence. (part 3)}
\label{fig:app-dependency-flood-3}

\end{figure*}



\begin{figure*}
\begin{tcolorbox}[ title={\textsc{hungry\_piplups}: Full problem, editorials, and highlighted diagnosis}]
\small

\textbf{Problem Statement.}

Near Snowpoint City, food has become scarce, and the local Piplups have been fighting over fishing rights on the many icebergs in the area. In Snowpoint Bay, there are an infinite number of icebergs arranged in a line. Every day, exactly one Piplup journeys from the mainland to one of these icebergs, making it their home and fishing ground.

Piplups are very territorial. If a piplup swims to an iceberg $X$ that has already been claimed, he will be chased out by the piplup already on the island and will attempt to swim to iceberg $X + 1$ instead. \hl{This continues until he reaches an empty iceberg.}

Professor Piplup, who has eaten enough gummis to maximize his IQ, has been observing and recording these Piplup migration patterns. Over $N$ days, he has recorded, for the $i^{\text{th}}$ day, exactly one piplup left for iceberg $A_i$. Some of his graduate students managed to steal some of his records to plan for their migration trip and have the following queries for you.

For each query the student only have records from days $L_i$ to $R_i$, they assume that no Piplups left on days outside this range. As a group of $X_i$ Piplups, they will start leaving one by one, all heading for iceberg $1$. Not wanting to drift too far downstream, they wish to determine the \hl{furthest iceberg} (i.e., the largest numbered iceberg) that any of them will end up at with the information they have. Note that despite being friends, \hl{\textbf{there must still be at most one piplup per iceberg}.}

There are also updates as the professor occasionally remembers that he recorded the $P_i^{\text{th}}$ Piplup going to a different iceberg, $X_i$, prompting an update to his records.

The students are unsure of the correct answer and seek your help in answering their queries. Please assist the students in determining the answer to their migration dilemma!

\medskip\hrule\medskip

\textbf{Task Summary.}

Notably, the task is as such. Define an operation $\oplus$ as
\[
S \oplus x = S \cup \{\text{Smallest } y : y \notin S \text{ and } y \ge x \}.
\]

Given $N$ numbers $\{A_1, A_2, \ldots, A_N\}$. Handle $M$ queries:
\begin{itemize}
  \item \textbf{Update:} Change $A_{P_i} = X_i$.
  \item \textbf{Query:} Find the $X_i^{\text{th}}$ smallest number not in
  $(\ldots ((\emptyset \oplus A_{L_i}) \oplus \ldots) \oplus A_{R_i})$.
\end{itemize}

\textbf{Input.}

The first line of input contains two integers $N, M$, as mentioned above.

The second line of input contains $N$ integers
$A_1, A_2, \ldots, A_N\ (0 \le A_i \le N - 1)$.

The third line of input contains $M$ queries of two possible types:
\begin{itemize}
  \item Type 0: \texttt{0 Pi Xi}
  \item Type 1: \texttt{1 Li Ri Xi}
\end{itemize}

For type 0 query, the value at $P_i$ is changed to $X_i$ (i.e.\ $A_{P_i} = X_i$).

For type 1 query, you are asked to output the position of the $X_i^{\text{th}}$ piplup only considering piplups from day $L_i$ to $R_i$ inclusive.

\medskip
\textbf{Limits.}
\begin{itemize}
  \item $1 \le N \le 10^6$
  \item $1 \le M \le 5 \times 10^3$
  \item \hl{$1 \le X_i, A_i \le 5 \times 10^2$}
  \item $1 \le L_i \le R_i \le N$
  \item $1 \le P_i \le N$
\end{itemize}

\medskip
\textbf{Output.}

For each type 1 query, output a single integer representing the position of the $X_i^{\text{th}}$ piplup only considering piplups from day $L_i$ to $R_i$ inclusive.

\end{tcolorbox}

\caption{Full problem, gold editorial, and model-generated editorial for \textsc{Hungry Piplups}, with highlighted diagnostic evidence. (part 1)}
\label{fig:app-hungry-piplups-1}

\end{figure*}
\begin{figure*}
\begin{tcolorbox}

\small
\medskip
\textbf{Sample Explanation.}

For the first type 1 query, the piplups are at icebergs 2, 3, 4, thus the $2^{\text{nd}}$ piplup will be at iceberg 5.

For the second type 1 query, the piplups are at icebergs 1, 2, 3, thus the $2^{\text{nd}}$ piplup will be at iceberg 5.

For the second type 1 query, the piplups are at icebergs 1, 2, 3, 4, 5, thus the $10^{\text{th}}$ piplup will be at iceberg 15.

\medskip
\textbf{Sample Input \#1}
\begin{verbatim}
5 4
1 3 2 4 5
1 2 4 2
0 2 1
1 1 3 2
1 1 5 10
\end{verbatim}

\textbf{Sample Output \#1}
\begin{verbatim}
5
5
15
\end{verbatim}

\medskip\hrule\medskip

\textbf{Gold editorial (full).}

Let $S_{L,R}$ be the set of locations of piplups, if we only consider the piplups from days $[L, R]$.
Let’s first try to generate this set for all queries quickly for each query.

\medskip
\textbf{Observation 1:}
Firstly, note that the operation (piplups going to iceburgs) is \hl{associative}.
I.e.\ instead of processing the piplups one by one from left to right, we can first process
the left half, $S_{L,M}$ and the right half, $S_{M+1,R}$ and merge them together.
The merging can be done simply in $\mathcal{O}(R - L)$ via a merge sort like operation.
This motivates a \hl{segment tree like solution}.

This can be done by constructing a normal segment tree on the range of $[1,N]$ and in each node
storing $S_{s,e}$. Note that this takes $\mathcal{O}(N \log N)$ to construct.
Unfortunately, for a $[L, R]$ query, it still takes $\mathcal{O}(N)$ to generate $S_{L,R}$
as the size of the sets may be up to $\mathcal{O}(N)$ and merging will take $\mathcal{O}(N)$.

\medskip
\textbf{Observation 2:}
Note that for a set $S_{L,R}$, it consists of \hl{consecutive segments of iceburgs}
where the starting iceburg of the segment is $\le \max A_i$.
Assume otherwise, it means that there exists a segment that start at $X > \max A_i$,
which means that iceburg $X - 1 \ge \max A_i$ is empty.
In which case, consider the piplup at $X$ it should have started at a position
$\le \max A_i$, and thus stop at $X - 1$ instead of $X$ which is a contradiction.

\medskip
Thus, instead of storing all the positions, we can reduce the size of the set to be
\hl{$\mathcal{O}(\max A_i)$} and simply store the segments $(\texttt{StartIdx}, \texttt{Size})$.
The merging operation should still be similar but take $\mathcal{O}(\max A_i)$.

\medskip
Finally to answer the queries we can simply generate $S_{L,R}$ in
$\mathcal{O}(\max A_i \times \log N)$
then iterate through the set until we encounter $X_i$ empty iceburgs,
which can be done in $\mathcal{O}(\max A_i)$.
For updates, we can do a normal segment tree update and regenerate the node sets in
$\mathcal{O}(\max A_i \times \log N)$ as well.

\medskip
Final complexity:
\[
\mathcal{O}(N \log N + M \log N \times \max A_i).
\]

\medskip
\textbf{Note:}
Alternative solutions that simply involves a min+index segment tree can also pass
but may require some constant time optimisations.

\medskip\hrule\medskip

\textbf{Model editorial (GPT-5, full; verbatim content but LaTeX-formatted).}

\textbf{Restatement.}

We have an infinite line of icebergs labeled $1,2,3,\ldots$. Each recorded day $i$ contributes one Piplup trying to claim iceberg $A_i$, and if it is already occupied he moves right to the first free iceberg $\ge A_i$. Over a range of days $[L,R]$, this produces a set $S$ of occupied icebergs (one distinct iceberg per day). A group of $X$ new Piplups then leaves for iceberg $1$ (one at a time with the same ``go to first free $\ge$ start'' rule), and we want the furthest iceberg any of them reaches. Equivalently, we want the $X$-th smallest positive integer not in $S$.

Additionally, records can be updated: a single $A[P]$ changes to a new value.

\medskip
\textbf{Constraints to exploit.}
\begin{itemize}
  \item $N$ up to $10^6$ (large array).
  \item $M$ up to $5 \times 10^3$ (few total operations).
  \item All values $A_i$ and $X$ in queries are small: \hl{$\le 500$}. This is the key.
\end{itemize}
\textbf{Key ideas.}

\textbf{(1) The final set depends only on the multiset of starting points (order does not matter).}
The insertion rule ``place each item at the smallest free $\ge$ its starting point'' admits an equivalent left-to-right scan:

\end{tcolorbox}

\caption{Full problem, gold editorial, and model-generated editorial for \textsc{Hungry Piplups}, with highlighted diagnostic evidence. (part 2)}
\label{fig:app-hungry-piplups-2}

\end{figure*}
\begin{figure*}
\begin{tcolorbox}

\small

Maintain a backlog $b$ of Piplups that have started but not yet been placed. Let $c[p]$ be the number of Piplups starting at position $p$.
For $p = 1,2,3,\ldots$:
\begin{itemize}
  \item Add arrivals: $b \leftarrow b + c[p]$.
  \item If $b > 0$, occupy $p$ and set $b \leftarrow b - 1$; otherwise, $p$ is empty.
\end{itemize}
This reproduces the greedy placement: position $p$ is the earliest free location for any still-unplaced Piplup whose start is $\le p$.


\textbf{(2) Bounded region (only first $500$ positions can have arrivals).}
Let $\texttt{MAXA} = 500$. Then $c[p]=0$ for all $p > \texttt{MAXA}$. After processing $p=\texttt{MAXA}$, the backlog drains deterministically:
positions $\texttt{MAXA}+1,\ldots,\texttt{MAXA}+b$ are all occupied, and everything after that is empty.

\textbf{(3) Query answer from holes + tail.}
Let:
\begin{itemize}
  \item $\texttt{occPrefix}$ = number of occupied positions among $1..\texttt{MAXA}$,
  \item $\texttt{holes} = \texttt{MAXA} - \texttt{occPrefix}$,
  \item $\texttt{tail}$ = backlog $b$ after processing $p=\texttt{MAXA}$.
\end{itemize}
Then:
\begin{itemize}
  \item If $X \le \texttt{holes}$, the answer is the position of the $X$-th empty among $1..\texttt{MAXA}$.
  \item Otherwise, the answer is $\texttt{MAXA} + \texttt{tail} + (X - \texttt{holes})$.
\end{itemize}

\textbf{(4) Data structure for dynamic range counts.}
Maintain an ordered set of positions for each value $v \in [1,\texttt{MAXA}]$ to count occurrences in any range $[L,R]$ efficiently:
\[
c[v] = \#\{i \in [L,R] : A[i]=v\}.
\]
Support point updates by moving index $P$ between sets.

\medskip
\textbf{Algorithm in detail.}

\textbf{Preprocessing.}
For each $v \in [1,\texttt{MAXA}]$, build an ordered set $S[v]$ containing all indices $i$ such that $A[i]=v$.

\textbf{Update operation \texttt{0 P X}.}
Let $\texttt{old} = A[P]$. Remove $P$ from $S[\texttt{old}]$, insert $P$ into $S[X]$, and set $A[P]\leftarrow X$.

\textbf{Query operation \texttt{1 L R X}.}
First compute $c[v]$ for all $v=1..\texttt{MAXA}$ using two order-statistic queries per $v$ (count of indices in $[L,R]$).
Then run the backlog scan for $p=1..\texttt{MAXA}$:
\begin{itemize}
  \item $b \leftarrow b + c[p]$.
  \item If $b>0$, occupy $p$ and decrement $b$; else, $p$ is a hole (empty).
\end{itemize}
If the $X$-th hole occurs within $1..\texttt{MAXA}$, output its position immediately. Otherwise, after the scan, output
$\texttt{MAXA} + b + (X - \texttt{holes})$.

\medskip
\textbf{Why it works.}
The backlog scan is equivalent to the greedy ``first free $\ge$ start'' placement: at each position $p$, after adding arrivals, $b>0$ exactly when some unplaced Piplup has start $\le p$, so the earliest free position must be $p$ and should be occupied. After $p>\texttt{MAXA}$, there are no new arrivals, so the remaining backlog occupies a contiguous tail.

\medskip
\textbf{Complexity and limits.}
Per query:
\[
\mathcal{O}(\texttt{MAXA}\log N) \text{ to compute all } c[v] \quad + \quad \mathcal{O}(\texttt{MAXA}) \text{ for the scan}.
\]
Per update: $\mathcal{O}(\log N)$.
With $\texttt{MAXA}=500$ and $M \le 5000$, this is well within limits.

\medskip
\textbf{Notes and edge cases.}
This approach relies on the constraint $\max A_i \le 500$. If $A_i$ can be $0$, either start the scan at $p=0$ or shift indices by $+1$.


\textbf{Worked example (sample).}
\begin{itemize}
  \item Initial $A=[1,3,2,4,5]$, query $(L,R,X)=(2,4,2)$: occupied at $2,3,4$, so the $2$-nd missing is $5$.
  \item Update $P=2$ to $1$: $A=[1,1,2,4,5]$. Query $(1,3,2)$: occupied at $1,2,3$, so the $2$-nd missing is $5$.
  \item Query $(1,5,10)$: occupied at $1,2,3,4,5$, so the $10$-th missing is $15$.
\end{itemize}

\medskip
\textbf{Conclusion.}
Aggregate counts by starting positions, simulate occupancy with the backlog scan on the bounded region $1..\texttt{MAXA}$, and answer the $X$-th missing via holes + tail. With ordered sets per value, both updates and range queries are efficient.


\end{tcolorbox}

\caption{Full problem, gold editorial, and model-generated editorial for \textsc{Hungry Piplups}, with highlighted diagnostic evidence. (part 3)}
\label{fig:app-hungry-piplups-3}

\end{figure*}

\end{document}